\documentclass[letterpaper,journal]{IEEEtran}

\usepackage{amsmath,amsfonts}
\usepackage{algorithmic}
\usepackage{algorithm}
\usepackage{array}
\usepackage[caption=false,font=normalsize,labelfont=sf,textfont=sf]{subfig}
\usepackage{textcomp}
\usepackage{stfloats}
\usepackage{placeins}
\usepackage{url}
\usepackage{verbatim}
\usepackage{graphicx}
\usepackage{cite}
\usepackage{booktabs}
\usepackage{multirow}
\usepackage{multicol}
\usepackage{fix-cm}
\usepackage{iftex}
\usepackage{etoolbox}
\usepackage[T1]{fontenc}
\usepackage{tgadventor}
\newcommand{\tablefont}{\fontfamily{qag}\selectfont}
\newcommand{\roadmapfont}{\fontencoding{T1}\fontfamily{ppl}\selectfont}
\renewcommand{\tablefont}{\roadmapfont}
\AtBeginEnvironment{tabular}{\tablefont}
\AtBeginEnvironment{table*}{\small\setlength{\tabcolsep}{2pt}\renewcommand{\arraystretch}{1.28}}
\usepackage[table]{xcolor}
\usepackage{tikz}
\usepackage{forest}
\useforestlibrary{edges}
\usepackage[hidelinks]{hyperref}
\usetikzlibrary{arrows.meta,positioning,fit,backgrounds,calc,matrix}

\definecolor{Lzero}{RGB}{150,155,165}
\definecolor{Lone}{RGB}{77,182,172}
\definecolor{Ltwo}{RGB}{74,144,226}
\definecolor{Lthree}{RGB}{125,91,190}
\definecolor{Lfour}{RGB}{238,147,55}

\newcolumntype{T}[1]{>{\raggedright\arraybackslash}p{#1}}

\makeatletter
\newcommand{\roadmaplink}[2]{\hyperref[#1]{#2}}
\providecommand{\roadmapname}[1]{%
  \ifcsname roadmapname@#1\endcsname
    \csname roadmapname@#1\endcsname
  \else #1\fi
}
\expandafter\def\csname roadmapname@a221209611\endcsname{Promptist}
\expandafter\def\csname roadmapname@a230409337\endcsname{Promptify}
\expandafter\def\csname roadmapname@a230515328\endcsname{VPGen}
\expandafter\def\csname roadmapname@a230917102\endcsname{MGIE}
\expandafter\def\csname roadmapname@a231010640\endcsname{LLM Blueprint}
\expandafter\def\csname roadmapname@a231206739\endcsname{SmartEdit}
\expandafter\def\csname roadmapname@a240210882\endcsname{POSI}
\expandafter\def\csname roadmapname@a240212741\endcsname{MuLan}
\expandafter\def\csname roadmapname@a240304997\endcsname{DiffChat}
\expandafter\def\csname roadmapname@a240313248\endcsname{Mora}
\expandafter\def\csname roadmapname@a240317804\endcsname{OPT2I}
\expandafter\def\csname roadmapname@a240809787\endcsname{Anim-Director}
\expandafter\def\csname roadmapname@a240810453\endcsname{Kubrick}
\expandafter\def\csname roadmapname@a240811788\endcsname{DreamFactory}
\expandafter\def\csname roadmapname@a241010076\endcsname{VideoAgent}
\expandafter\def\csname roadmapname@a241104925\endcsname{StoryAgent}
\expandafter\def\csname roadmapname@a241106558\endcsname{RAG}
\expandafter\def\csname roadmapname@a241108127\endcsname{TIPO}
\expandafter\def\csname roadmapname@a241202259\endcsname{VideoGen-of-Thought}
\expandafter\def\csname roadmapname@a241204440\endcsname{GenMAC}
\expandafter\def\csname roadmapname@a241210419\endcsname{PASTA}
\expandafter\def\csname roadmapname@a250112909\endcsname{FilmAgent}
\expandafter\def\csname roadmapname@a250200848\endcsname{RealRAG}
\expandafter\def\csname roadmapname@a250203207\endcsname{MotionAgent}
\expandafter\def\csname roadmapname@a250209411\endcsname{ImageRAG}
\expandafter\def\csname roadmapname@a250301298\endcsname{FoX}
\expandafter\def\csname roadmapname@a250305242\endcsname{MM-StoryAgent}
\expandafter\def\csname roadmapname@a250307314\endcsname{MovieAgent}
\expandafter\def\csname roadmapname@a250310719\endcsname{LVAS-Agent}
\expandafter\def\csname roadmapname@a250400010\endcsname{LayerCraft}
\expandafter\def\csname roadmapname@a250407046\endcsname{Unified Agentic Framework}
\expandafter\def\csname roadmapname@a250414868\endcsname{Twin-Co}
\expandafter\def\csname roadmapname@a250502648\endcsname{MCCD}
\expandafter\def\csname roadmapname@a250521956\endcsname{Cross-modal RAG}
\expandafter\def\csname roadmapname@a250601370\endcsname{PointT2I}
\expandafter\def\csname roadmapname@a250604158\endcsname{IEAP}
\expandafter\def\csname roadmapname@a250616853\endcsname{RATTPO}
\expandafter\def\csname roadmapname@a250623138\endcsname{VisualPrompter}
\expandafter\def\csname roadmapname@a250719939\endcsname{LLMControl}
\expandafter\def\csname roadmapname@a250722076\endcsname{TIR}
\expandafter\def\csname roadmapname@a250805606\endcsname{Uni-cot}
\expandafter\def\csname roadmapname@a250808487\endcsname{MAViS}
\expandafter\def\csname roadmapname@a250816644\endcsname{CountLoop}
\expandafter\def\csname roadmapname@a250818781\endcsname{AniME}
\expandafter\def\csname roadmapname@a250910704\endcsname{Maestro}
\expandafter\def\csname roadmapname@a250910761\endcsname{EditDuet}
\expandafter\def\csname roadmapname@a250912446\endcsname{PromptSculptor}
\expandafter\def\csname roadmapname@a250913642\endcsname{LLM-I}
\expandafter\def\csname roadmapname@a251010633\endcsname{Collaborative Text-to-Image Generation}
\expandafter\def\csname roadmapname@a250922761\endcsname{MILR}
\expandafter\def\csname roadmapname@a251015831\endcsname{VISTA}
\expandafter\def\csname roadmapname@a251022431\endcsname{Hollywood Town}
\expandafter\def\csname roadmapname@a251108521\endcsname{UniVA}
\expandafter\def\csname roadmapname@a251111483\endcsname{ImAgent}
\expandafter\def\csname roadmapname@a251111780\endcsname{Image-POSER}
\expandafter\def\csname roadmapname@a251114100\endcsname{RIVER}
\expandafter\def\csname roadmapname@a251119458\endcsname{Personalized Reward Modeling}
\expandafter\def\csname roadmapname@a251204221\endcsname{MoReGen}
\expandafter\def\csname roadmapname@a251205112\endcsname{DraCo}
\expandafter\def\csname roadmapname@a251209081\endcsname{AgentComp}
\expandafter\def\csname roadmapname@a251212196\endcsname{AutoMV}
\expandafter\def\csname roadmapname@a251222536\endcsname{CoAgent}
\expandafter\def\csname roadmapname@a251223568\endcsname{ThinkGen}
\expandafter\def\csname roadmapname@a260102046\endcsname{Agentic Retoucher}
\expandafter\def\csname roadmapname@a260118543\endcsname{GenAgent}
\expandafter\def\csname roadmapname@a260201756\endcsname{Mind-Brush}
\expandafter\def\csname roadmapname@a260202051\endcsname{SIDiffAgent}
\expandafter\def\csname roadmapname@a260202437\endcsname{UniReason 1.0}
\expandafter\def\csname roadmapname@a260206166\endcsname{M3}
\expandafter\def\csname roadmapname@a260209084\endcsname{Agent Banana}
\expandafter\def\csname roadmapname@a260211790\endcsname{LASEV}
\expandafter\def\csname roadmapname@a260212279\endcsname{UniT}
\expandafter\def\csname roadmapname@a260222809\endcsname{PhotoAgent}
\expandafter\def\csname roadmapname@a260302697\endcsname{ShareVerse}
\expandafter\def\csname roadmapname@a260302816\endcsname{BrandFusion}
\expandafter\def\csname roadmapname@a260311421\endcsname{ShotVerse}
\expandafter\def\csname roadmapname@a260312155\endcsname{GlyphBanana}
\expandafter\def\csname roadmapname@a260312829\endcsname{coDrawAgents}
\expandafter\def\csname roadmapname@a260328088\endcsname{GEMS}
\expandafter\def\csname roadmapname@a260328767\endcsname{Gen-Searcher}
\expandafter\def\csname roadmapname@a260329620\endcsname{Unify-Agent}
\expandafter\def\csname roadmapname@a260329664\endcsname{CutClaw}
\expandafter\def\csname roadmapname@a260404746\endcsname{Think in Strokes, Not Pixels}
\expandafter\def\csname roadmapname@a260409195\endcsname{Camera Artist}
\expandafter\def\csname roadmapname@a260410456\endcsname{CineAgents}
\expandafter\def\csname roadmapname@a260413491\endcsname{FiRe}
\expandafter\def\csname roadmapname@a260416541\endcsname{BOOKAGENT}
\expandafter\def\csname roadmapname@a260424842\endcsname{Co-Director}
\expandafter\def\csname roadmapname@a260425636\endcsname{RvR}
\expandafter\def\csname roadmapname@a260504040\endcsname{UniReasoner}
\expandafter\def\csname roadmapname@a260507457\endcsname{EditRefiner}
\expandafter\def\csname roadmapname@a260512495\endcsname{AlphaGRPO}
\expandafter\def\csname roadmapname@a260514709\endcsname{Breaking Dual Bottlenecks}
\expandafter\def\csname roadmapname@a260516961\endcsname{LAC}
\expandafter\def\csname roadmapname@a260517969\endcsname{Generation Navigator}
\expandafter\def\csname roadmapname@a260518748\endcsname{Aurora}
\expandafter\def\csname roadmapname@a260526525\endcsname{ReCA}
\expandafter\def\csname roadmapname@a260527374\endcsname{ICG}
\expandafter\def\csname roadmapname@a260530248\endcsname{GenClaw}
\expandafter\def\csname roadmapname@a260600204\endcsname{APE}
\expandafter\def\csname roadmapname@a260603243\endcsname{MemoGen}
\expandafter\def\csname roadmapname@a260605031\endcsname{MetaPoint}
\expandafter\def\csname roadmapname@a260607636\endcsname{Crayotter}
\expandafter\def\csname roadmapname@a260607649\endcsname{ViMax}
\expandafter\def\csname roadmapname@a260608091\endcsname{VideoWeaver}
\expandafter\def\csname roadmapname@a260626907\endcsname{Qwen-Image-Agent}
\expandafter\def\csname roadmapname@a260727380\endcsname{VideoCoCo}
\expandafter\def\csname roadmapname@aeslides\endcsname{AESlides}
\expandafter\def\csname roadmapname@agentic3d\endcsname{Agentic 3D Scene Generation}
\expandafter\def\csname roadmapname@autoslides\endcsname{AutoSlides}
\expandafter\def\csname roadmapname@avaencoder\endcsname{AVA-Encoder}
\expandafter\def\csname roadmapname@compagent\endcsname{CompAgent}
\expandafter\def\csname roadmapname@controlnet\endcsname{ControlNet}
\expandafter\def\csname roadmapname@designcoder\endcsname{DesignCoder}
\expandafter\def\csname roadmapname@diffusionagent\endcsname{Diffusion Agent}
\expandafter\def\csname roadmapname@dpok\endcsname{DPOK}
\expandafter\def\csname roadmapname@dreambooth\endcsname{DreamBooth}
\expandafter\def\csname roadmapname@dreamfusion\endcsname{DreamFusion}
\expandafter\def\csname roadmapname@dreamgaussian\endcsname{DreamGaussian}
\expandafter\def\csname roadmapname@evalcrafter\endcsname{EvalCrafter}
\expandafter\def\csname roadmapname@frontalk\endcsname{AceCoder}
\expandafter\def\csname roadmapname@frontcoder\endcsname{FrontCoder}
\expandafter\def\csname roadmapname@genartist\endcsname{GenArtist}
\expandafter\def\csname roadmapname@geneval\endcsname{GenEval}
\expandafter\def\csname roadmapname@genevolve\endcsname{GenEvolve}
\expandafter\def\csname roadmapname@genpilot\endcsname{GenPilot}
\expandafter\def\csname roadmapname@got\endcsname{GoT}
\expandafter\def\csname roadmapname@image_cot\endcsname{Image-CoT}
\expandafter\def\csname roadmapname@imagereward\endcsname{ImageReward}
\expandafter\def\csname roadmapname@instruct_pix2pix\endcsname{InstructPix2Pix}
\expandafter\def\csname roadmapname@ip_adapter\endcsname{IP-Adapter}
\expandafter\def\csname roadmapname@lave\endcsname{LAVE}
\expandafter\def\csname roadmapname@layoutgpt\endcsname{LayoutGPT}
\expandafter\def\csname roadmapname@ldm\endcsname{High-resolution image synthesis}
\expandafter\def\csname roadmapname@llm_grounded\endcsname{LMD}
\expandafter\def\csname roadmapname@magic3d\endcsname{Magic3D}
\expandafter\def\csname roadmapname@newton\endcsname{NEWTON}
\expandafter\def\csname roadmapname@octot2i\endcsname{OctoT2I}
\expandafter\def\csname roadmapname@pickapic\endcsname{Pick-a-Pic}
\expandafter\def\csname roadmapname@pptagent\endcsname{PPTAgent}
\expandafter\def\csname roadmapname@pptarena\endcsname{PPTArena}
\expandafter\def\csname roadmapname@re_imagen\endcsname{Re-Imagen}
\expandafter\def\csname roadmapname@rpg\endcsname{RPG}
\expandafter\def\csname roadmapname@scenecraft\endcsname{SceneCraft}
\expandafter\def\csname roadmapname@scenethesis\endcsname{Scenethesis}
\expandafter\def\csname roadmapname@scope\endcsname{SCOPE}
\expandafter\def\csname roadmapname@screencoder\endcsname{ScreenCoder}
\expandafter\def\csname roadmapname@searchgen\endcsname{SearchGen}
\expandafter\def\csname roadmapname@self_correcting\endcsname{SLD}
\expandafter\def\csname roadmapname@slideagent\endcsname{SlideAgent}
\expandafter\def\csname roadmapname@slidetailor\endcsname{SlideTailor}
\expandafter\def\csname roadmapname@slidetranslation\endcsname{Slide Translation}
\expandafter\def\csname roadmapname@spiral\endcsname{SPIRAL}
\expandafter\def\csname roadmapname@src250215972\endcsname{MosAIG}
\expandafter\def\csname roadmapname@src250317671\endcsname{ComfyGPT}
\expandafter\def\csname roadmapname@src250405306\endcsname{CREA}
\expandafter\def\csname roadmapname@src250420054\endcsname{Marmot}
\expandafter\def\csname roadmapname@src250500703\endcsname{T2I-R1}
\expandafter\def\csname roadmapname@src250515779\endcsname{IA-T2I}
\expandafter\def\csname roadmapname@src250517908\endcsname{ComfyMind}
\expandafter\def\csname roadmapname@src250521660\endcsname{PreGenie}
\expandafter\def\csname roadmapname@src250524875\endcsname{ReasonGen-R1}
\expandafter\def\csname roadmapname@src250602015\endcsname{OSPO}
\expandafter\def\csname roadmapname@src250604676\endcsname{Gen-n-Val}
\expandafter\def\csname roadmapname@src250605010\endcsname{ComfyUI-Copilot}
\expandafter\def\csname roadmapname@src250609790\endcsname{ComfyUI-R1}
\expandafter\def\csname roadmapname@src250705259\endcsname{X-Planner}
\expandafter\def\csname roadmapname@src250718634\endcsname{Captain Cinema}
\expandafter\def\csname roadmapname@src250806916\endcsname{Talk2Image}
\expandafter\def\csname roadmapname@src250810494\endcsname{MAGUS}
\expandafter\def\csname roadmapname@src250817188\endcsname{PosterGen}
\expandafter\def\csname roadmapname@src250817435\endcsname{RefineEdit-Agent}
\expandafter\def\csname roadmapname@src250906945\endcsname{IRG}
\expandafter\def\csname roadmapname@src251022521\endcsname{ORIG}
\expandafter\def\csname roadmapname@src251027452\endcsname{VisPainter}
\expandafter\def\csname roadmapname@src251121087\endcsname{MIRA}
\expandafter\def\csname roadmapname@src251123002\endcsname{JarvisEvo}
\expandafter\def\csname roadmapname@src260103250\endcsname{MultiMedia-Agent}
\expandafter\def\csname roadmapname@src260103741\endcsname{I2E}
\expandafter\def\csname roadmapname@src260104060\endcsname{ComfySearch}
\expandafter\def\csname roadmapname@src260104390\endcsname{SciFig}
\expandafter\def\csname roadmapname@src260104589\endcsname{MiLDEdit}
\expandafter\def\csname roadmapname@src260104794\endcsname{APEX}
\expandafter\def\csname roadmapname@src260105016\endcsname{From Idea to Co-Creation}
\expandafter\def\csname roadmapname@src260109150\endcsname{World Craft}
\expandafter\def\csname roadmapname@src260110332\endcsname{Think-Then-Generate}
\expandafter\def\csname roadmapname@src260111109\endcsname{VIGA}
\expandafter\def\csname roadmapname@src260114602\endcsname{3D Space as a Scratchpad}
\expandafter\def\csname roadmapname@src260117737\endcsname{The Script is All You Need}
\expandafter\def\csname roadmapname@src260122571\endcsname{PerfGuard}
\expandafter\def\csname roadmapname@src260123265\endcsname{PaperBanana}
\expandafter\def\csname roadmapname@src260203828\endcsname{AutoFigure}
\expandafter\def\csname roadmapname@src260203866\endcsname{PaperX}
\expandafter\def\csname roadmapname@src260208368\endcsname{T2VTree}
\expandafter\def\csname roadmapname@src260209153\endcsname{SceneSmith}
\expandafter\def\csname roadmapname@src260210116\endcsname{SAGE}
\expandafter\def\csname roadmapname@src260213318\endcsname{DECKBench}
\expandafter\def\csname roadmapname@src260214968\endcsname{PhyScensis}
\expandafter\def\csname roadmapname@src260217558\endcsname{RetouchIQ}
\expandafter\def\csname roadmapname@src260219542\endcsname{Vinedresser3D}
\expandafter\def\csname roadmapname@src260220664\endcsname{AnimeAgent}
\expandafter\def\csname roadmapname@src260222839\endcsname{DeepPresenter}
\expandafter\def\csname roadmapname@src260300483\endcsname{RAISE}
\expandafter\def\csname roadmapname@src260302681\endcsname{VisionCreator}
\expandafter\def\csname roadmapname@src260303646\endcsname{InfinityStory}
\expandafter\def\csname roadmapname@src260306032\endcsname{StruVis}
\expandafter\def\csname roadmapname@src260307106\endcsname{AutoUE}
\expandafter\def\csname roadmapname@src260307148\endcsname{Mukherjee et al. (2026)}
\expandafter\def\csname roadmapname@src260308059\endcsname{ImageEdit-R1}
\expandafter\def\csname roadmapname@src260311048\endcsname{COMIC}
\expandafter\def\csname roadmapname@src260311554\endcsname{MANSION}
\expandafter\def\csname roadmapname@src260312238\endcsname{SceneAssistant}
\expandafter\def\csname roadmapname@src260312310\endcsname{VQQA}
\expandafter\def\csname roadmapname@src260312597\endcsname{Feynman}
\expandafter\def\csname roadmapname@src260314724\endcsname{GameUIAgent}
\expandafter\def\csname roadmapname@src260314790\endcsname{Mind-of-Director}
\expandafter\def\csname roadmapname@src260316839\endcsname{Learning to Present}
\expandafter\def\csname roadmapname@src260316967\endcsname{MSRAMIE}
\expandafter\def\csname roadmapname@src260318627\endcsname{AFS-Search}
\expandafter\def\csname roadmapname@src260319708\endcsname{WorldAgents}
\expandafter\def\csname roadmapname@src260320644\endcsname{ScaleEdit-12M}
\expandafter\def\csname roadmapname@src260327817\endcsname{CAIAMAR}
\expandafter\def\csname roadmapname@src260329590\endcsname{FigAgent}
\expandafter\def\csname roadmapname@src260329602\endcsname{IMAGAgent}
\expandafter\def\csname roadmapname@src260403156\endcsname{CAMEO}
\expandafter\def\csname roadmapname@src260404875\endcsname{DIRECT}
\expandafter\def\csname roadmapname@src260405076\endcsname{GLANCE}
\expandafter\def\csname roadmapname@src260405489\endcsname{SCMAPR}
\expandafter\def\csname roadmapname@src260407721\endcsname{Sima 1.0}
\expandafter\def\csname roadmapname@src260407966\endcsname{LiVER}
\expandafter\def\csname roadmapname@src260409568\endcsname{EvoDiagram}
\expandafter\def\csname roadmapname@src260410383\endcsname{Authoring for Living Worlds}
\expandafter\def\csname roadmapname@src260413452\endcsname{CANVAS}
\expandafter\def\csname roadmapname@src260415917\endcsname{ATR}
\expandafter\def\csname roadmapname@src260416958\endcsname{Luo et al. (2026)}
\expandafter\def\csname roadmapname@src260420730\endcsname{Render-in-the-Loop}
\expandafter\def\csname roadmapname@src260423579\endcsname{CineAGI}
\expandafter\def\csname roadmapname@src260423580\endcsname{PhysCodeBench}
\expandafter\def\csname roadmapname@src260425318\endcsname{Cutscene Agent}
\expandafter\def\csname roadmapname@src260501477\endcsname{Action Agent}
\expandafter\def\csname roadmapname@src260509423\endcsname{SimWorld Studio}
\expandafter\def\csname roadmapname@src260511363\endcsname{PresentAgent-2}
\expandafter\def\csname roadmapname@src260515181\endcsname{From Plans to Pixels}
\expandafter\def\csname roadmapname@src260515187\endcsname{Articraft}
\expandafter\def\csname roadmapname@src260516748\endcsname{Genflow Ad Studio}
\expandafter\def\csname roadmapname@src260517423\endcsname{Soap2Soap}
\expandafter\def\csname roadmapname@src260519587\endcsname{SceneCode}
\expandafter\def\csname roadmapname@src260522144\endcsname{One Sentence, One Drama}
\expandafter\def\csname roadmapname@src260522448\endcsname{S2ED}
\expandafter\def\csname roadmapname@src260523527\endcsname{LiveFigure}
\expandafter\def\csname roadmapname@src260524453\endcsname{Code2UML}
\expandafter\def\csname roadmapname@src260527705\endcsname{AgenticVBench}
\expandafter\def\csname roadmapname@src260528056\endcsname{CogPortrait}
\expandafter\def\csname roadmapname@src260528173\endcsname{MangaFlow}
\expandafter\def\csname roadmapname@src260530090\endcsname{DirectorBench}
\expandafter\def\csname roadmapname@src260530611\endcsname{Crafter}
\expandafter\def\csname roadmapname@src260601057\endcsname{3DCodeBench}
\expandafter\def\csname roadmapname@src260602320\endcsname{TVIR}
\expandafter\def\csname roadmapname@src260602915\endcsname{Any2Poster}
\expandafter\def\csname roadmapname@src260603168\endcsname{JAVEDIT}
\expandafter\def\csname roadmapname@src260606002\endcsname{Global-Local Monte Carlo Tree Search}
\expandafter\def\csname roadmapname@src260608016\endcsname{IEA}
\expandafter\def\csname roadmapname@src260608402\endcsname{SceneConductor}
\expandafter\def\csname roadmapname@src260609738\endcsname{HDSL}
\expandafter\def\csname roadmapname@src260613368\endcsname{IterCAD}
\expandafter\def\csname roadmapname@src260613679\endcsname{InterleaveThinker}
\expandafter\def\csname roadmapname@src260613861\endcsname{TBS}
\expandafter\def\csname roadmapname@src260614168\endcsname{MUSE}
\expandafter\def\csname roadmapname@src260616103\endcsname{SceneCraft}
\expandafter\def\csname roadmapname@src260616184\endcsname{CoTriSyGen}
\expandafter\def\csname roadmapname@src260617162\endcsname{MemSlides}
\expandafter\def\csname roadmapname@src260617536\endcsname{OmniDrive}
\expandafter\def\csname roadmapname@src260618591\endcsname{CHIEF}
\expandafter\def\csname roadmapname@src260619073\endcsname{SCPE}
\expandafter\def\csname roadmapname@src260620764\endcsname{WMGen-v1}
\expandafter\def\csname roadmapname@src260623221\endcsname{RS-Gen}
\expandafter\def\csname roadmapname@src260623327\endcsname{VideoAgent}
\expandafter\def\csname roadmapname@src260627376\endcsname{Ask, Solve, Generate}
\expandafter\def\csname roadmapname@src260628016\endcsname{TempAct}
\expandafter\def\csname roadmapname@src260628971\endcsname{SEAR}
\expandafter\def\csname roadmapname@src260629395\endcsname{NaLA}
\expandafter\def\csname roadmapname@src260630296\endcsname{ManimAgent}
\expandafter\def\csname roadmapname@src260631537\endcsname{DataEvolver}
\expandafter\def\csname roadmapname@src260700920\endcsname{GMO-E2DIT}
\expandafter\def\csname roadmapname@src260701102\endcsname{SAGE}
\expandafter\def\csname roadmapname@src260701588\endcsname{OrchestrXR}
\expandafter\def\csname roadmapname@src260701709\endcsname{COMFYCLAW}
\expandafter\def\csname roadmapname@src260701766\endcsname{SimWorlds}
\expandafter\def\csname roadmapname@src260701883\endcsname{PairCoder++}
\expandafter\def\csname roadmapname@src260702590\endcsname{OmniPresent}
\expandafter\def\csname roadmapname@src260703731\endcsname{CoGen3D}
\expandafter\def\csname roadmapname@src260705465\endcsname{CanvasAgent}
\expandafter\def\csname roadmapname@src260708497\endcsname{Cognitive-structured Multimodal Agent}
\expandafter\def\csname roadmapname@src260709839\endcsname{Malik et al. (2026)}
\expandafter\def\csname roadmapname@src260711594\endcsname{MAGIC}
\expandafter\def\csname roadmapname@src260712042\endcsname{SymbOmni}
\expandafter\def\csname roadmapname@src260713125\endcsname{Boogu-Image-0.1}
\expandafter\def\csname roadmapname@src260715845\endcsname{Li et al. (2026)}
\expandafter\def\csname roadmapname@src260716352\endcsname{CLARE}
\expandafter\def\csname roadmapname@src260716355\endcsname{PhysAgent}
\expandafter\def\csname roadmapname@src260719038\endcsname{FilmWorld}
\expandafter\def\csname roadmapname@src260719947\endcsname{ETPDesigner}
\expandafter\def\csname roadmapname@src260720866\endcsname{Agentic Designer}
\expandafter\def\csname roadmapname@src260720889\endcsname{Lumera}
\expandafter\def\csname roadmapname@src260721522\endcsname{GS-Agent}
\expandafter\def\csname roadmapname@src260722241\endcsname{AgentHOI}
\expandafter\def\csname roadmapname@src260723491\endcsname{PlanCraft}
\expandafter\def\csname roadmapname@src260723588\endcsname{JarvisHub}
\expandafter\def\csname roadmapname@src260723920\endcsname{EmoScope}
\expandafter\def\csname roadmapname@src260724353\endcsname{PRISM}
\expandafter\def\csname roadmapname@src260726910\endcsname{CinemaTraj}
\expandafter\def\csname roadmapname@src260728073\endcsname{GVR-Coder}
\expandafter\def\csname roadmapname@src260800548\endcsname{DrawAI}
\expandafter\def\csname roadmapname@src260800629\endcsname{ParticleGen}
\expandafter\def\csname roadmapname@src260800891\endcsname{CADIR}
\expandafter\def\csname roadmapname@src260802218\endcsname{PosterMELD}
\expandafter\def\csname roadmapname@src260802694\endcsname{Crayotter}
\expandafter\def\csname roadmapname@src260803298\endcsname{SeaSlides}
\expandafter\def\csname roadmapname@src260804071\endcsname{MCTS-Report}
\expandafter\def\csname roadmapname@src260804622\endcsname{DAC-Pose}
\expandafter\def\csname roadmapname@src260804964\endcsname{WorldCycle}
\expandafter\def\csname roadmapname@src260805248\endcsname{WorldClaw}
\expandafter\def\csname roadmapname@src260805485\endcsname{VideoArgus}
\expandafter\def\csname roadmapname@src260806075\endcsname{Hu et al. (2026)}
\expandafter\def\csname roadmapname@src260806161\endcsname{iARCS}
\expandafter\def\csname roadmapname@src260806751\endcsname{Atelier}
\expandafter\def\csname roadmapname@src260810408\endcsname{VisEditBench}
\expandafter\def\csname roadmapname@src260811635\endcsname{VisPuzzle}
\expandafter\def\csname roadmapname@src260812290\endcsname{Agentic Self-Improvement}
\expandafter\def\csname roadmapname@src260812314\endcsname{StateFlow}
\expandafter\def\csname roadmapname@src260813560\endcsname{AutoDesign}
\expandafter\def\csname roadmapname@src260816721\endcsname{GenRouter}
\expandafter\def\csname roadmapname@t2i_adapter\endcsname{T2I-Adapter}
\expandafter\def\csname roadmapname@t2i_compbench\endcsname{T2I-CompBench}
\expandafter\def\csname roadmapname@t2i_copilot\endcsname{T2I-Copilot}
\expandafter\def\csname roadmapname@talkslides\endcsname{TalkSlides}
\expandafter\def\csname roadmapname@toolartist\endcsname{ToolArtist}
\expandafter\def\csname roadmapname@ui2coden\endcsname{UI2Code$^{\mathrm{N}}$}
\expandafter\def\csname roadmapname@vbench\endcsname{VBench}
\expandafter\def\csname roadmapname@video_diffusion\endcsname{Video diffusion models}
\expandafter\def\csname roadmapname@visionguided\endcsname{CITL}
\expandafter\def\csname roadmapname@visrefiner\endcsname{VisRefiner}
\expandafter\def\csname roadmapname@visual_chatgpt\endcsname{Visual ChatGPT}
\expandafter\def\csname roadmapname@webvia\endcsname{WebVIA}
\expandafter\def\csname roadmapname@whattoeditnext\endcsname{What to Edit Next}
\expandafter\def\csname roadmapname@world_to_image\endcsname{World-To-Image}

\newcommand{\roadmaprefs}[1]{\begingroup\fontsize{6.7}{7.85}\selectfont\def\roadmapsep{}\@for\roadmapkey:=#1\do{\roadmapsep\csname cite\endcsname{\roadmapkey}\def\roadmapsep{\kern-.38em\allowbreak}}\endgroup}
\newcommand{\roadmapmethodlist}[2]{\parbox[c]{\linewidth}{\raggedright\sloppy\textit{e.g.}~\def\roadmapsep{}\@for\roadmapkey:=#1\do{\roadmapsep{\color{black}\roadmapname{\roadmapkey}}~{\color{roadmapred}\csname cite\endcsname{\roadmapkey}}\def\roadmapsep{;\space}}#2\par}}
\newcommand{\roadmapmethods}[1]{\roadmapmethodlist{#1}{}}
\newcommand{\roadmapmethodsEtc}[1]{\roadmapmethodlist{#1}{;~\textit{etc.}}}
\makeatother

\begin{document}

\title{Agentic Visual Generation: From Generative Models to Agentic Control}

\author{Yinming Huang\textsuperscript{*}, Shuyuan Tu\textsuperscript{*}, Xi Yan\textsuperscript{*}, Jiahao Zhan, Zihan Yang, Zhen Xing, Hui Zhang, Tiehua Zhang, \emph{Member, IEEE}, Yu-Gang Jiang, \emph{Fellow, IEEE}, and Zuxuan Wu\textsuperscript{\textdagger}, \emph{Member, IEEE}%
\thanks{Y. Huang, S. Tu, X. Yan, Z. Yang, H. Zhang, Z. Wu, and Y-G. Jiang are with Fudan University. Emails: \{ymhuang26, sytu23, 24307140086, yangzh26, hui\_zhang23\}@m.fudan.edu.cn, \{zxwu, ygj\}@fudan.edu.cn

Y. Huang and Z. Wu are also with Shanghai Innovative Institute. 

J. Zhan is with CUHK, MMLab. Email: 1155271220@link.cuhk.edu.hk. 

Z. Xing is with Wan Team, Alibaba Tongyi Lab. Email: zxing20@fudan.edu.cn. 

T. Zhang is with the School of Computer Science and Technology, Tongji University. Emails: tiehuaz@tongji.edu.cn. 

$^{*}$These authors contributed equally. $^{\dagger}$Corresponding author.}}

\markboth{IEEE Transactions/Journal Draft}%
{Author \MakeLowercase{\textit{et al.}}: Agentic Visual Generation}

\makeatletter
\IEEEaftertitletext{%
\begin{minipage}{\textwidth}
\centering
\includegraphics[width=\textwidth]{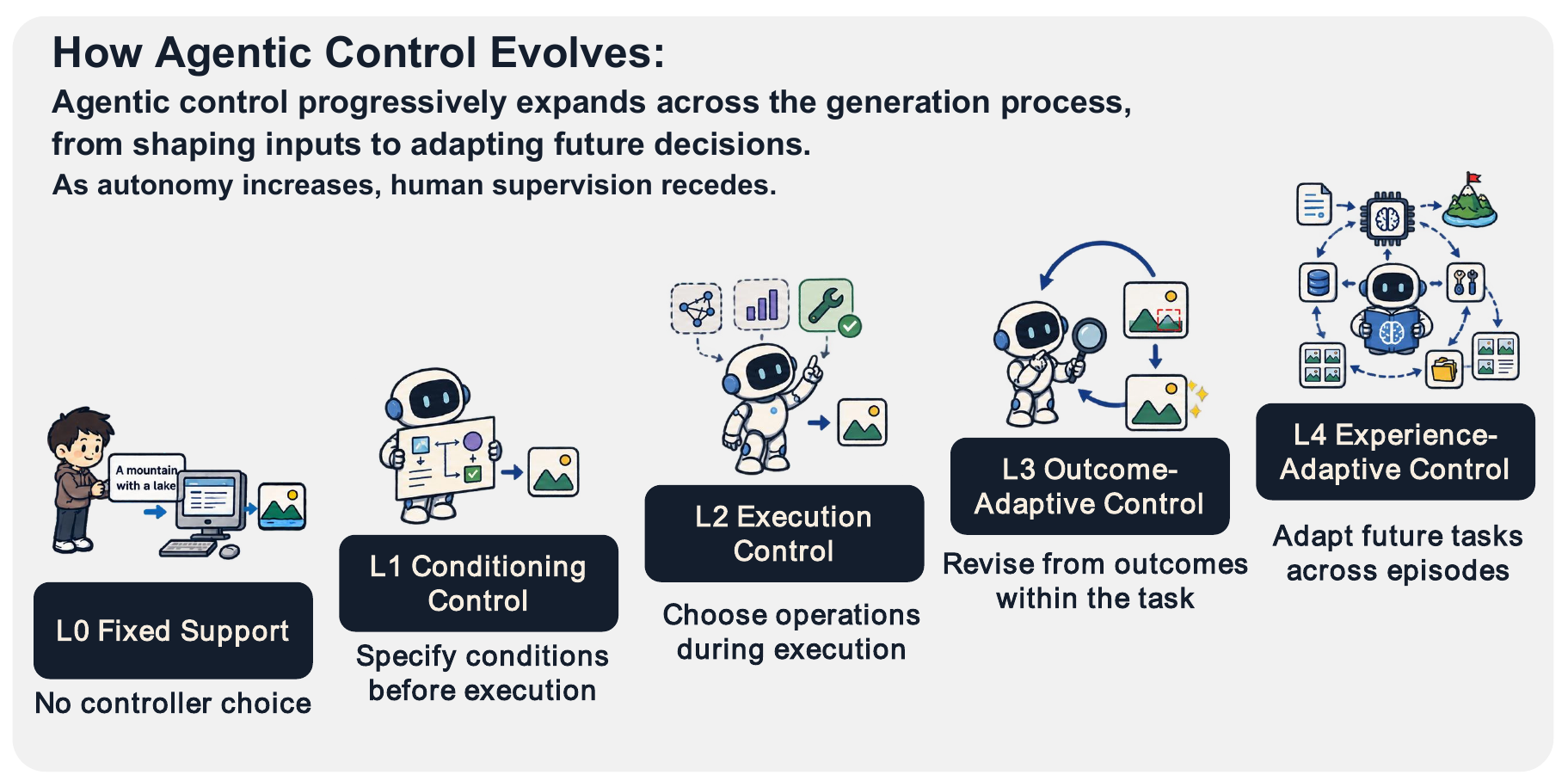}
\vspace{-0.8\baselineskip}
\refstepcounter{figure}
\@makecaption{\fnum@figure}{Visual overview of controller decision-making scope. L0 Fixed Support supplies predetermined generative components. L1 Conditioning Control constructs a specification for a predetermined executor. L2 Execution Control selects and invokes visual operations. L3 Outcome-Adaptive Control uses observed outcomes to revise the current task. L4 Experience-Adaptive Control retains completed-task experience that changes a later independent task.}
\label{fig:taxonomy-tree}
\end{minipage}
\vspace{1.0\baselineskip}}
\makeatother

\maketitle

\begin{abstract}
Visual generation is evolving from generative models used through a single invocation into agentic control processes that can plan, select tools, inspect intermediate synthesized outputs, revise failures, and reuse prior experience. In most existing systems, the controller is an LLM or VLM, while visual generation models serve as tools or executors. However, existing work lacks a consistent criterion for determining when a generation system becomes agentic. Planning depth, tool use, multi-role collaboration, and reinforcement learning are often treated as evidence of agenticity, even though none of them necessarily determines which generation decisions the controller can make. We organize the field according to what the controller can directly control in the generation process. At L1 Conditioning Control, the controller prepares the input to a predetermined generator but does not control which visual operation is executed. At L2 Execution Control, it selects and invokes actual generation, editing, rendering, or other content-modifying operations. At L3 Outcome-Adaptive Control, it observes an intermediate outcome and uses that observation to change a subsequent operation within the current task. At L4 Experience-Adaptive Control, it retains experience from completed tasks and uses that experience to change decisions on future tasks. L0 Fixed Support separately denotes generators, editors, evaluators, reward models, benchmarks, and fixed pipelines without a deployed controller that makes generation-level decisions. These levels describe a progressively broader decision-making scope rather than model size, system complexity, output quality, tool or role count, or training method. Applying this framework across image, video, editing, 3D, world, slide, and user-interface generation reveals how controller capabilities have evolved and how their mechanisms are distributed across levels. We further develop a level-conditioned evaluation framework that isolates the value of broadening the controller's decision-making scope by matching generators, tools, budgets, and evaluators across levels. Finally, we identify the key transitions from generative models to executable control, reliable outcome-driven revision, reusable cross-task experience, and a future generator-as-controller regime. Project resources are available at \href{https://github.com/YinmingHuang/Awesome-agentic-visual-generation-model}{the project repository}.
\end{abstract}


\section{Introduction}
\IEEEPARstart{V}{isual} generation, including image, video, structured visual content, and 3D generation, has long been a challenging research problem that supports creative applications, communication, software development, and simulation. Deep generative models have substantially improved visual quality and language alignment. DALL-E~2~\cite{dalle2}, Imagen~\cite{imagen}, and Parti~\cite{parti} established strong text-conditioned image synthesis at scale. Latent Diffusion~\cite{ldm}, SDXL~\cite{sdxl}, and DALL-E~3~\cite{dalle3} subsequently improved efficient training, high-resolution generation, and instruction following. Video Diffusion Models~\cite{video_diffusion} and Make-A-Video~\cite{make_a_video} transferred these advances to temporal synthesis, while Imagen Video~\cite{imagen_video}, Video LDM~\cite{video_ldm}, and Lumiere~\cite{lumiere} further developed high-resolution, latent-space, and space-time generation designs~\cite{video_diffusion_survey}. These advances improve what a visual executor can produce from a supplied condition~\cite{visual_generation_new_era}. In our hierarchy, these fixed generators, editors, retrievers, evaluators, and predetermined pipelines constitute \textbf{L0 Fixed Support}. L0 marks the inclusion boundary because these components provide generation and evidence capabilities but do not contain a deployed controller that decides how the generation process should proceed.

The transition from L0 Fixed Support to \textbf{L1 Conditioning Control} addresses a limitation of a fixed executor: the user's request may not directly provide a usable spatial specification or the external knowledge needed for generation. L1 methods solve this problem by constructing the condition consumed by one predetermined executor. LLM-grounded Diffusion (LMD)~\cite{llm_grounded} converts a complex request into object descriptions and bounding boxes that guide a frozen diffusion model. LayoutGPT~\cite{layoutgpt} similarly uses in-context reasoning to produce explicit 2D or 3D layouts before rendering. World-To-Image~\cite{world_to_image} retrieves definitions and reference images when the generator lacks knowledge of a requested entity, then incorporates that evidence into the generator-facing condition. These methods therefore resolve ambiguity before generation. Their remaining limitation is that the controller still cannot decide which visual operation should execute.

The transition from L1 Conditioning Control to \textbf{L2 Execution Control} addresses this operation-selection limitation. An L2 controller can choose and invoke a generator, editor, program, or workflow rather than only prepare the input to a fixed executor. Visual ChatGPT~\cite{visual_chatgpt} turns visual foundation models into callable tools and uses a language-model controller to select and invoke the operation required by the request. ComfyUI-Copilot~\cite{src250605010} constructs an executable node graph whose components and data flow determine the generation route. ViMax~\cite{a260607649} extends executable control to coordinated video operations such as script preparation, shot planning, character styling, and clip generation. These systems solve the problem of selecting and sequencing capabilities, but their selected route can remain open loop because a generated result does not necessarily change the next action.

The transition from L2 Execution Control to \textbf{L3 Outcome-Adaptive Control} addresses failures that become visible only after execution. Benchmarks such as T2I-CompBench~\cite{t2i_compbench} and GenEval~\cite{geneval} reveal compositional and object-binding errors, while VBench~\cite{vbench} and EvalCrafter~\cite{evalcrafter} expose temporal and perceptual defects in generated videos. These evaluators diagnose the limitation, but they do not solve it by themselves. L3 systems close the loop by mapping an observed result to a later generation action. SLD~\cite{self_correcting} converts a diagnosed mismatch into a new sampling decision, while GenPilot~\cite{genpilot} uses visual feedback to choose a subsequent refinement. L3 therefore extends control beyond execution, although the resulting state and repair experience can remain confined to the current task.

The transition from L3 Outcome-Adaptive Control to \textbf{L4 Experience-Adaptive Control} addresses this episode boundary. An L4 controller retains information from a completed task and uses it to change decisions on a later independent task. OctoT2I~\cite{octot2i} updates persistent capability profiles that inform future routing decisions. GenEvolve~\cite{genevolve} distills successful and failed generation trajectories into reusable procedures, while COMFYCLAW~\cite{src260701709} promotes verified workflow-construction procedures into a reusable skill library. Together, L0 through L4 are the five labels in our hierarchy. The progression has a specific causal meaning: each transition extends the latest point at which evidence can change a future generation decision, from constructing a condition at L1, to selecting an operation at L2, revising the current task at L3, and adapting future tasks at L4. A future generator-as-controller is therefore not introduced as an additional level. It is classified by the same maximum-causal-reach test.

The literature nevertheless remains fragmented across tasks and uses inconsistent criteria for identifying agenticity. A unified account must distinguish controller decision-making scope from generator progress, implementation topology, and learning procedure.

\textbf{Scope, classification, and evaluation.} Against this background, we develop a controller-centered framework that places fixed supporting components in L0 Fixed Support, which marks the inclusion boundary, and classifies L1--L4 systems by the latest future generation decision their controller can change. This single test applies across modular systems, multi-role workflows, and unified generative policies. It also determines the matched evaluation design: each comparison adds one class of controller decisions while holding the generator, tools, budget, and evaluator fixed.

\textbf{Inclusion boundary.} This definition also determines the inclusion boundary of the framework. Every system within our principal scope must contain a primary visual generator or editor controlled by a generation-level decision process. The system may collaborate with other agents or call external tools. A standalone supporting module or a fixed pipeline is not treated as an agentic visual generation system by itself.

\textbf{Contributions.} The contributions are as follows:
\begin{itemize}
    \item We define agenticity by the maximum temporal and causal reach of the decisions a controller can make. This criterion yields a reproducible hierarchy from L0 Fixed Support through L4 Experience-Adaptive Control. L0 marks the inclusion boundary, while a category and subcategory taxonomy organizes systems by the object controlled and its technical realization.
    \item We release a structured corpus with level, task, mechanism, feedback, memory, resource, and provenance fields. Its temporal and cross-sectional statistics reveal the rapid shift toward within-trajectory feedback while persistent cross-task experience remains uncommon.
    \item We organize representative systems as a design space of controlled variables and feedback paths, emphasizing the causal decisions that distinguish methods rather than enumerating paper titles.
    \item We propose level-conditioned evaluation that isolates the value of specification, execution, outcome feedback, and reusable experience under matched generators, tools, budgets, and evaluators.
\end{itemize}

\textbf{Organization.} Sections II--III define the generation--control distinction and the L0--L4 taxonomy. Sections IV--VIII analyze L0--L4, Section IX covers training, Section X develops matched evaluation, and Sections XI--XII discuss open transitions and conclude the paper.

\section{From Visual Generation to Generation-Level Control}
\label{sec:prelim}

Visual generation describes what an executor can create or edit. Agentic visual generation additionally describes how a controller selects conditions, invokes operations, and changes later actions around that executor. This section establishes that distinction, defines the controller and generator roles, and introduces the task and mechanism axes used in the paper.

\begin{figure*}[!t]
\centering
\includegraphics[width=\textwidth]{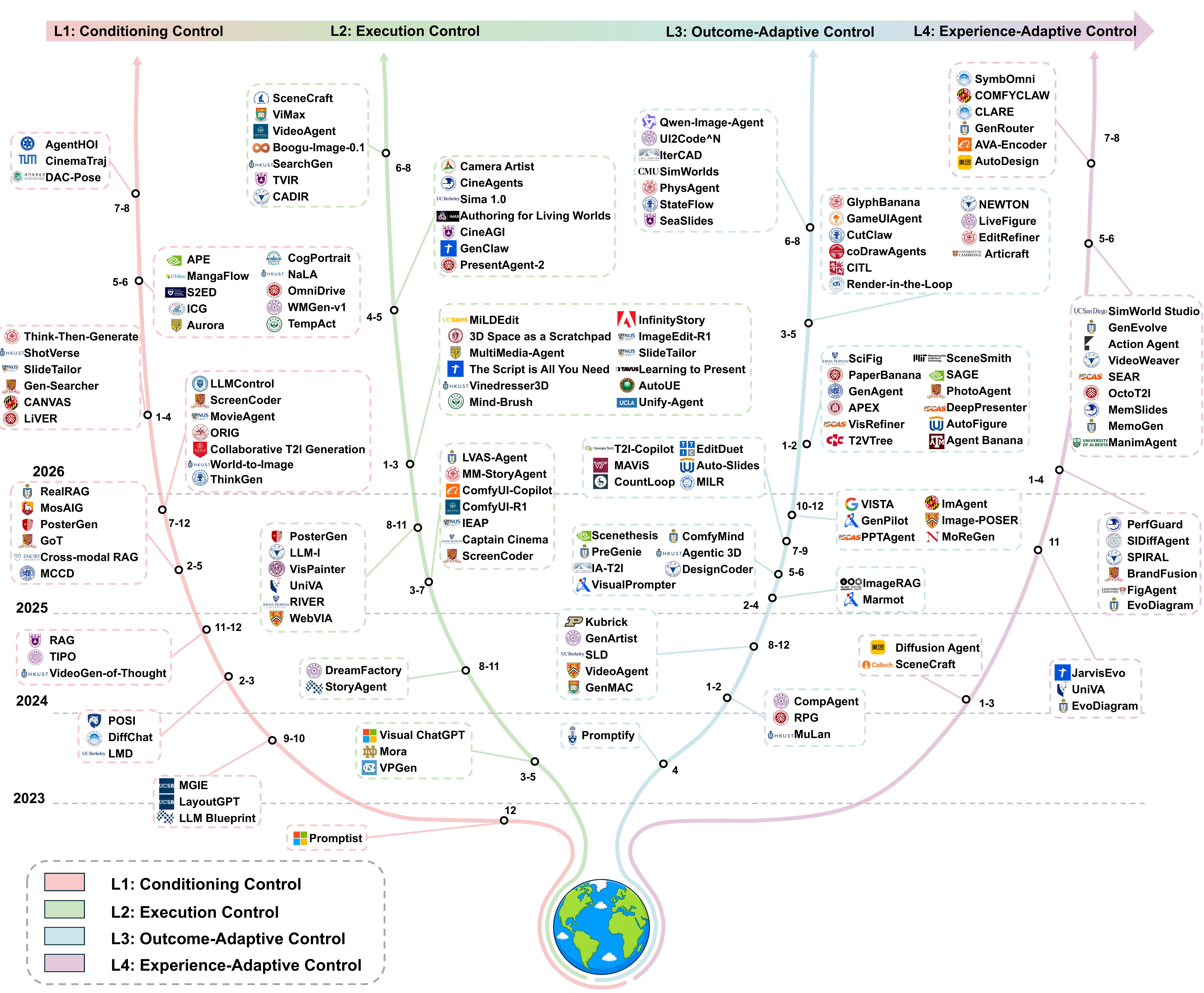}
\caption{Popular-paper roadmap across the L1--L4 controller levels, organized by first public release and the latest decision capability exposed by each representative system, from condition construction and operation selection to outcome-driven revision and reusable cross-task experience; the roadmap is a qualitative overview rather than a quantitative ranking.}
\label{fig:popular-paper-roadmap}
\end{figure*}

\begin{figure*}[!t]
\centering
\includegraphics[width=\textwidth]{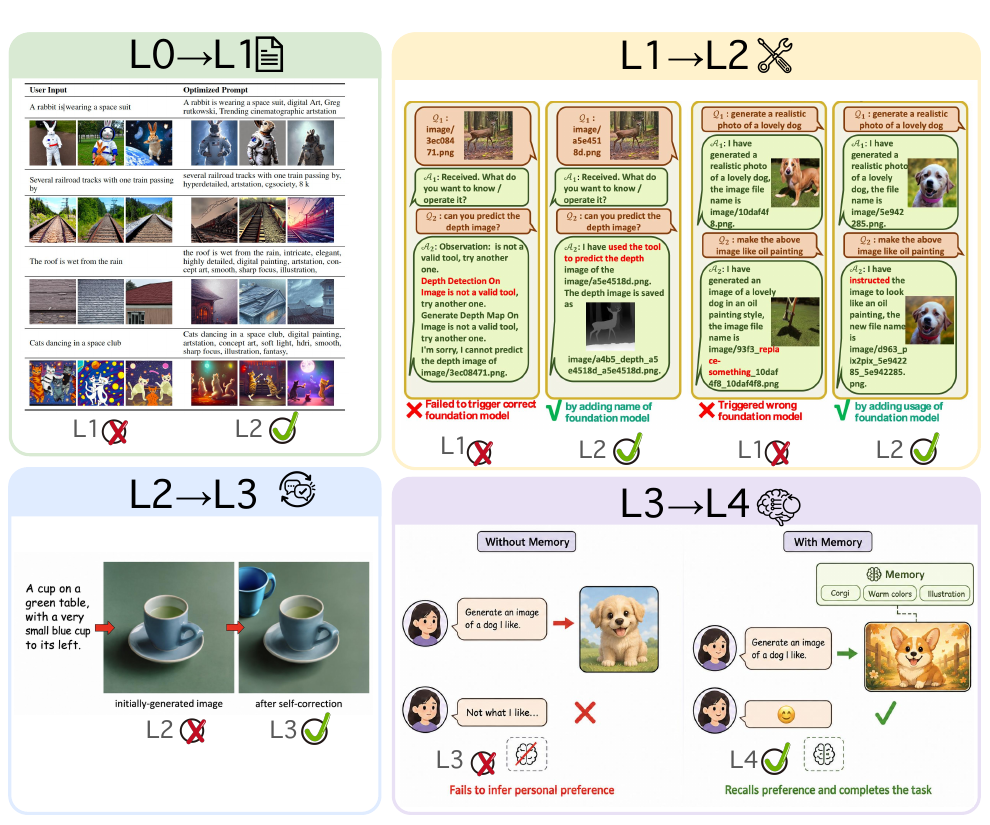}
\caption{Illustrative comparisons across adjacent controller levels. Each panel adds one decision capability to the preceding level: condition construction in L1 Conditioning Control, operation selection in L2 Execution Control, outcome-dependent revision in L3 Outcome-Adaptive Control, or cross-task experience reuse in L4 Experience-Adaptive Control. The examples are conceptual and do not constitute quantitative comparisons.}
\label{fig:level-comparisons}
\end{figure*}

\subsection{Conceptual Foundations: Agents and Visual Generation}
This separation begins with the two concepts that the field often conflates. Agentic visual generation combines a controller with the state-construction capability of a visual generator. Separating these functions is important because a model that accepts rich conditions does not automatically control a generation trajectory, and a controller that only interprets images is outside the generation scope of this work.

\subsubsection{Agents}
We first characterize the controller side of this separation. An agent is a goal-directed system that selects actions from observations and maintains enough state to adapt later decisions. Its controller can be a language model, a multimodal language model, a learned policy, a search procedure, or a team of specialized roles. The defining property is not the controller architecture. It is the presence of intermediate decisions that influence task execution. In visual generation, these decisions include prompt revision, layout construction, model routing, reference selection, editing, verification, memory update, and stopping.

Agentic behavior can be implemented by multiple coordinated models or by a single model~\cite{tu2023implicit,tu2022multiple}. A multi-model system may combine a controller with specialized generators, editors, or evaluators. A single-model system, such as a unified multimodal model, may perform several of these functions through one learned policy. In both cases, agenticity requires an action space, an observation process, and a mechanism through which current state or feedback changes a later generation decision. Model count does not determine agenticity.

\subsubsection{Visual Generation}
The generator side supplies the complementary state-construction capability. Visual generation covers the creation, editing~\cite{tu2024motioneditor,tu2024motionfollower}, and refinement of images~\cite{yang2026arcflow}, videos~\cite{tu2026flashportrait,tu2026baton,tu2025stableavatar,tu2025stableanimator,tu2025stableanimator++}, visual stories, slides, user interfaces, 3D scenes~\cite{leng2026preference}, and interactive environments. For slides and user interfaces, the action representation may be PowerPoint objects, XML, HTML, CSS, or executable code, but the synthesized content is still judged through its rendered visual structure and behavior. Latent Diffusion~\cite{ldm}, Video Diffusion Models~\cite{video_diffusion}, and DreamFusion~\cite{dreamfusion} learn mappings from conditions to pixels, temporal sequences, or radiance fields. Design2Code~\cite{design2code} instead predicts executable interface representations. Generation controllers place these mappings within a larger decision process. Their generated content is not only a terminal output. It may also be an observation, a memory item, a candidate for comparison, or a persistent world state.

\subsection{Definition of Agentic Visual Generation}
With the controller and generator roles separated, the next question is how they must interact for a complete system to enter the scope of this work. We define an agentic visual generation system as a system that contains a visual generator or editor and a decision process that controls generation over one or more steps. Let $g$ denote the user goal, $s_t$ the current multimodal state, $a_t$ an action, and $o_{t+1}$ the resulting observation. The controller follows $a_t \sim \pi(a_t\mid g,s_t)$, while the environment transition $s_{t+1}=F(s_t,a_t,o_{t+1})$ may include a newly generated image, video segment, scene, critique, or retrieved reference. The objective is not only to maximize visual quality. It may balance instruction satisfaction, consistency, controllability, latency, monetary cost, and human effort.

This formulation highlights three properties. First, the system creates part of its own future observation space. Second, the action space can mix symbolic actions, tool calls, and visual generation. Third, the quality of a trajectory depends on both the final output and the decisions used to obtain it.

\subsection{Basic Components}
The definition specifies controller decision-making scope abstractly, so we next identify the components that realize it. Every system in our scope contains both a \emph{controller} and a \emph{primary visual generator or editor}. The generator creates or updates the synthesized visual content, while the controller maps the goal, state, and observations to generation-level actions. In the predominant architecture, an LLM, VLM, or MLLM serves as this controller, while visual generators and renderers are exposed as tools or executors. The following paragraphs distinguish distributed roles and tools, single-model controllers, and the optional memory or learning mechanisms that can extend either design. A generator does not become an agent merely by producing complex visual content.

\textbf{Collaborating roles and tools.} The controller may collaborate with specialized agents or invoke external tools. Collaborating agents can interpret user intent, construct plans, assign subtasks, critique intermediate results, preserve consistency, or manage long-horizon workflows. External tools can include retrievers, controlled generators, editors, detectors, segmenters, renderers, simulators, PowerPoint object models, browser runtimes, verifiers, and reward models. Their outputs provide conditions, executable actions, evidence, or feedback that the controller uses to decide how the synthesized content should be created or revised.

\textbf{Single-model controllers.} When these functions are not distributed across roles and tools, agentic behavior can instead be implemented by one learned model. Unified multimodal models are the main example in the current literature. UI2Code$^{\mathrm{N}}$~\cite{ui2coden}, for instance, generates synthesized content, inspects its rendered state, and decides whether to refine it. We treat a single model as a controller only when it demonstrates a control decision over conditions, execution, or later actions. Joint understanding and generation, one-shot visual-token prediction, or reward-based generator post-training alone remain insufficient. Within the analyzed corpus, most systems still combine a language-based or multimodal language-based controller with a separate visual generator.

\textbf{Memory and learning.} Regardless of whether control is distributed or implemented by one model, memory and learning can further retain accepted synthesized outputs, user preferences, world states, tool experience, and successful strategies. These mechanisms are optional for open-loop and within-episode agents, but become defining capabilities when information from completed tasks changes later control. This persistence criterion also applies to a future generator-as-controller system, which would need to select tools or models, manage state, interpret outcomes, and alter its own trajectory rather than only update pixels or visual tokens.

\subsection{Task Axis}
Components describe what a system contains, but comparison also requires separating what is generated from how generation is controlled. The task axis includes image generation and editing, video generation and editing, slide and user-interface generation, 3D asset and scene construction, world-grounded synthesis, and interactive simulation. These tasks differ in temporal horizon, state persistence, action granularity, and the cost of evaluating intermediate results. Task type does not determine controller capability. An image system can exhibit a longer adaptive trajectory than a video system, while a multi-shot video pipeline can remain open loop.

\subsection{Mechanism Axis}
Because task type does not reveal controller decision-making scope, a second axis records the mechanism used to realize it. The mechanism axis includes intent grounding, explicit or latent planning, tool routing, retrieval, multi-agent collaboration, verification, memory, and reinforcement learning. These mechanisms describe how a controller is implemented, but they do not define its level. Tool use specifies an action space, multi-agent design specifies a topology, and reinforcement learning specifies an optimization procedure. Any of them can appear at several capability levels.

\section{A Hierarchy of Controller Decision-Making Scope}
\label{sec:taxonomy}

The task and mechanism axes describe a system, but neither orders agenticity. Our organizing principle is instead causal: \emph{agenticity in visual generation is determined by the deepest point in a generation trajectory at which a controller can causally change a future generation decision}. ``Deepest'' refers to temporal reach along the trajectory. A condition precedes execution, an execution decision determines which visual operation occurs, an observed outcome can redirect a later action within the same task, and persistent experience can affect an action after the current task has ended.

\subsection{Controller-Capability Axis}
We classify each complete system by the latest future generation decision that its controller can change. Table~\ref{tab:taxonomy} is the canonical definition of the five labels. Lower capabilities remain visible as a capability path, while the primary label records only the maximum demonstrated reach. For example, L1+L2+L3 records specification construction, operation invocation, and outcome-dependent revision, while assigning the system to L3 Outcome-Adaptive Control. The following paragraphs first establish L0 Fixed Support and its boundary role, then separate decision-making scope from architecture, and finally state the decision procedure.

\textbf{L0 Fixed Support.} L0 is not a peer agent level. It records fixed supporting components and marks the inclusion boundary. A generator, editor, retriever, evaluator, reward model, benchmark, or fixed pipeline can be essential to an agent without selecting generation-level actions itself. Separating this boundary prevents complexity and support quality from being mistaken for controller decision-making scope.

\textbf{Decision-making scope rather than architecture.} With L0 Fixed Support established, the next distinction separates capability from implementation because common architectural labels collapse causally different systems. Several planners may still terminate in one declarative specification, whereas a compact router can directly determine which generator or editor executes. Likewise, a dynamically assembled workflow can remain open loop: its route may vary by request without changing after a generated result is observed. The hierarchy therefore records the controller's decision-making scope, not implementation complexity, model size, output quality, tool or role count, or training procedure.

\textbf{Decision procedure.} These distinctions lead to an assignment rule. We apply the tests from highest to lowest. Cross-task persistence establishes L4 Experience-Adaptive Control; an outcome-to-action causal link establishes L3 Outcome-Adaptive Control; selection and invocation of a visual operation establishes L2 Execution Control; and specification construction for a predetermined executor establishes L1 Conditioning Control. If none holds, the method is assigned to L0 Fixed Support. Ambiguous cases receive the lower level until the paper demonstrates the missing causal link. This rule makes the taxonomy reproducible and prevents terminology in a title, such as ``multi-agent,'' ``self-reflective,'' or ``self-evolving,'' from serving as evidence by itself.

\subsection{Fine-Grained Taxonomy and Roadmap}
The controller-capability level identifies the latest decision that a system can change, but it does not by itself organize systems that act at the same level. We therefore use one explicit second-stage axis per level. L0 is organized by support function, L1 by the primary generator-facing specification, L2 by the primary executable object, L3 by the decisive feedback source, and L4 by the persistent experience carrier. A \textbf{subcategory} captures a stable distinction within that axis rather than an implementation architecture or a label created for one paper. Each system receives one primary placement within its level; secondary capabilities remain part of the prose comparison. Figure~\ref{fig:fine-taxonomy-roadmap} gives the complete category-to-subcategory roadmap.

% Four-tier roadmap implemented with the same Forest grammar as Figs. 2--3
% of arXiv:2606.12191: level -> subsection -> branch -> papers.
\providecommand{\roadmaplink}[2]{#2}
\providecommand{\roadmapmethods}[1]{#1}
\providecommand{\tablefont}{\sffamily}
\providecommand{\roadmapfont}{\rmfamily}

\definecolor{roadmapnavy}{HTML}{14446A}
\definecolor{roadmapred}{HTML}{BD114A}
\definecolor{roadmapsectionblue}{HTML}{5E7AC4}
\definecolor{lzerofill}{HTML}{FFF7D5}
\definecolor{lonefill}{HTML}{F5D5D3}
\definecolor{ltwofill}{HTML}{DDF0DF}
\definecolor{lthreefill}{HTML}{DCF0F8}
\definecolor{lfourfill}{HTML}{E9E1F5}

% Keep each section marker directly after the corresponding title and make
% the marker itself a link to that section in the paper.
\newcommand{\roadmaplevel}[3]{#2 {\color{roadmapsectionblue}\roadmaplink{#1}{(#3)}}}
\newcommand{\roadmapsection}[3]{%
  \parbox[c]{11.5em}{\centering #2 {\color{roadmapsectionblue}\roadmaplink{#1}{(#3)}}\par}%
}

\forestset{
  roadmap tree/.style={
    forked edges,
    for tree={
      grow=east,
      reversed=true,
      anchor=base west,
      parent anchor=east,
      child anchor=west,
      base=left,
      font=\roadmapfont\normalsize,
      rectangle,
      draw=roadmapnavy,
      rounded corners,
      align=left,
      minimum width=4em,
      minimum height=1.5em,
      inner xsep=6pt,
      inner ysep=5pt,
      line width=1.1pt,
      edge+={darkgray,line width=1pt},
      edge path={
        \noexpand\path[\forestoption{edge}]
        (!u.parent anchor)--+(6pt,0)|-(.child anchor)
        \forestoption{edge label};
      },
      s sep=2pt,
    },
  },
  roadmap root/.style={phantom,draw=none,inner sep=0pt,for children={no edge}},
  roadmap level/.style={
    rotate=90,
    child anchor=north,
    parent anchor=south,
    anchor=center,
    align=center,
    inner xsep=8pt,
    inner ysep=5pt,
  },
  roadmap subsection/.style={
    anchor=center,
    parent anchor=east,
    child anchor=west,
    align=center,
    inner xsep=7pt,
    inner ysep=5pt,
  },
  roadmap branch/.style={
    text width=14em,
    anchor=center,
    parent anchor=east,
    child anchor=west,
    align=center,
    inner xsep=8pt,
    inner ysep=4.5pt,
  },
  roadmap papers/.style={
    text width=46.5em,
    anchor=center,
    parent anchor=east,
    child anchor=west,
    align=left,
    fill=#1,
    inner xsep=9pt,
    inner ysep=4.5pt,
    edge path={
      \noexpand\path[\forestoption{edge}]
      (!u.parent anchor)--(.child anchor)
      \forestoption{edge label};
    },
  },
}

\newcommand{\roadmappartone}{%
\begingroup
\let\tablefont\roadmapfont
\begin{forest} roadmap tree
[{},roadmap root
  [{\roadmaplevel{sec:l0}{L0: Fixed Support}{\S IV}},roadmap level,xshift=-1.4em,line width=1.4pt
    [{\roadmapsection{sec:l0-a}{Generation and Retrieval\\Components}{\S IV.A}},roadmap subsection
      [{(i) Controlled generation\\and editing},roadmap branch
        [{\roadmapmethods{ldm,video_diffusion,dreamfusion,controlnet,t2i_adapter,dreambooth,ip_adapter,instruct_pix2pix,a230917102,a231206739,magic3d,dreamgaussian}},roadmap papers=lzerofill]]
      [{(ii) Fixed retrieval},roadmap branch
        [{\roadmapmethods{re_imagen}},roadmap papers=lzerofill]]
    ]
    [{\roadmapsection{sec:l0-b}{Data Construction\\and Training\\Infrastructure}{\S IV.B}},roadmap subsection
      [{(i) Data construction},roadmap branch
        [{\roadmapmethods{src260320644,src260603168,src260631537,a251209081,src250604676}},roadmap papers=lzerofill]]
      [{(ii) Fixed generator training},roadmap branch
        [% Full reviewed list also includes src250602015, src260804964, and src260627376.
         {\roadmapmethodsEtc{dpok,aeslides,frontcoder,src250500703,src250524875}},roadmap papers=lzerofill]]
    ]
    [{\roadmapsection{sec:l0-c}{Evaluators and\\Benchmarks}{\S IV.C}},roadmap subsection
      [{(i) Evaluators and\\reward models},roadmap branch
        [% Full reviewed list also includes a250407046.
         {\roadmapmethodsEtc{imagereward,pickapic,a251119458}},roadmap papers=lzerofill]]
      [{(ii) Benchmarks},roadmap branch
        [{\roadmapmethods{t2i_compbench,geneval,vbench,evalcrafter,src260527705,src260530090,src260805485,src260213318,src260601057,src260810408}},roadmap papers=lzerofill]]
    ]
  ]
  [{\roadmaplevel{sec:l1}{L1: Conditioning Control}{\S V}},roadmap level,xshift=-1.4em,line width=1.4pt
    [{\roadmapsection{sec:l1-a}{Textual Prompt\\Specifications}{\S V.A}},roadmap subsection
      [{(i) Prompt transformation and\\constraint preservation},roadmap branch
        [{\roadmapmethods{a221209611,a241108127,a260600204,a240304997,a260527374,a240210882,a251223568,a230917102}},roadmap papers=lonefill]]
    ]
    [{\roadmapsection{sec:l1-b}{Spatial and Geometric\\Specifications}{\S V.B}},roadmap subsection
      [{(i) Layout and region\\specifications},roadmap branch
        [% Full reviewed list also includes a241106558.
         {\roadmapmethodsEtc{llm_grounded,layoutgpt,got,a250719939}},roadmap papers=lonefill]]
      [{(ii) Scene and pose\\specifications},roadmap branch
        [{\roadmapmethods{a231010640,src260629395,src260804622}},roadmap papers=lonefill]]
    ]
    [{\roadmapsection{sec:l1-c}{Retrieved Evidence\\Specifications}{\S V.C}},roadmap subsection
      [{(i) Retrieval policy},roadmap branch
        [{\roadmapmethods{world_to_image,a260328767}},roadmap papers=lonefill]]
      [{(ii) Evidence-to-condition\\grounding},roadmap branch
        [% Full reviewed list also includes src251022521 and src260620764.
         {\roadmapmethodsEtc{a250521956,a250200848,src250215972}},roadmap papers=lonefill]]
    ]
    [{\roadmapsection{sec:l1-d}{Temporal and Camera\\Specifications}{\S V.D}},roadmap subsection
      [{(i) Subject, action, and\\temporal specifications},roadmap branch
        [% Full reviewed list also includes src260407721 and src260303646.
         {\roadmapmethodsEtc{a260518748,src260528056,src260628016,src260722241,a241202259,a250307314,src260617536,src250718634,src260423579}},roadmap papers=lonefill]]
      [{(ii) Camera and viewpoint\\specifications},roadmap branch
        [% Full reviewed list also includes a260409195.
         {\roadmapmethodsEtc{a260311421,src260726910,src260407966}},roadmap papers=lonefill]]
    ]
    [{\roadmapsection{sec:l1-e}{Structured Content\\Specifications}{\S V.E}},roadmap subsection
      [{(i) Compositional and narrative\\specifications},roadmap branch
        [{\roadmapmethods{src260110332,src260307148,src260413452,src260522448,a250502648,a251010633,src260528173,a250305242}},roadmap papers=lonefill]]
      [{(ii) Document and interface\\specifications},roadmap branch
        [{\roadmapmethods{slidetailor,screencoder,src250817188}},roadmap papers=lonefill]]
    ]
  ]
]
\end{forest}%
\endgroup
}

\newcommand{\roadmapparttwo}{%
\begingroup
\let\tablefont\roadmapfont
\begin{forest} roadmap tree
[{},roadmap root
  [{\roadmaplevel{sec:l2}{L2: Execution Control}{\S VI}},roadmap level,xshift=-1.4em
    [{\roadmapsection{sec:l2-a}{Model and Tool\\Operations}{\S VI.A}},roadmap subsection
      [{(i) Operation and model routing},roadmap branch
        [% Full reviewed list also includes a260201756.
         {\roadmapmethodsEtc{visual_chatgpt,a250913642,src260713125,searchgen}},roadmap papers=ltwofill]]
      [{(ii) Multi-step workflow\\construction},roadmap branch
        [{\roadmapmethods{src250605010,src250609790,a260530248}},roadmap papers=ltwofill]]
    ]
    [{\roadmapsection{sec:l2-b}{Image and Structured-\\Graphic Operations}{\S VI.B}},roadmap subsection
      [{(i) Program-level image\\operations},roadmap branch
        [{\roadmapmethods{a230515328,a250604158,src260308059}},roadmap papers=ltwofill]]
      [{(ii) Element-level structured\\operations},roadmap branch
        [{\roadmapmethods{src251027452,src260104589,src260616103}},roadmap papers=ltwofill]]
    ]
    [{\roadmapsection{sec:l2-c}{Video and Audiovisual\\Operations}{\S VI.C}},roadmap subsection
      [{(i) Clip-level generation and\\editing operations},roadmap branch
        [{\roadmapmethods{a251114100,src260623327}},roadmap papers=ltwofill]]
      [{(ii) Shot and timeline operations},roadmap branch
        [% Full reviewed list also includes a260607649.
         {\roadmapmethodsEtc{src260117737,a241104925,a240811788,a260410456}},roadmap papers=ltwofill]]
      [{(iii) Cross-modal audiovisual\\operations},roadmap branch
        [{\roadmapmethods{a250310719,a240313248,src260103250}},roadmap papers=ltwofill]]
    ]
    [{\roadmapsection{sec:l2-d}{Document and Interface\\Operations}{\S VI.D}},roadmap subsection
      [{(i) Document and interface\\object operations},roadmap branch
        [{\roadmapmethods{src260511363,src260602320}},roadmap papers=ltwofill]]
    ]
    [{\roadmapsection{sec:l2-e}{3D, CAD, and World\\Operations}{\S VI.E}},roadmap subsection
      [{(i) Geometry and asset\\operations},roadmap branch
        [{\roadmapmethods{src260800891,src260219542}},roadmap papers=ltwofill]]
      [{(ii) Scene, engine, and world\\operations},roadmap branch
        [{\roadmapmethods{src260307106,src260114602,a260329620}},roadmap papers=ltwofill]]
    ]
  ]
]
\end{forest}%
\endgroup
}

\newcommand{\roadmappartthree}{%
\begingroup
\let\tablefont\roadmapfont
\begin{forest} roadmap tree
[{},roadmap root
  [{\roadmaplevel{sec:l3}{L3: Outcome-Adaptive Control}{\S VII}},roadmap level,xshift=5.4em
    [{\roadmapsection{sec:l3-a}{Perceptual Outcome\\Feedback}{\S VII.A}},roadmap subsection
      [{(i) Image and structured-\\content renders},roadmap branch
        [{\roadmapmethodsEtc{rpg,self_correcting,src260623221,genpilot,a250816644,a260605031,src260806751,a260413491,a250623138,src260416958,a250414868,src250515779}},roadmap papers=lthreefill]]
      [{(ii) Video renders},roadmap branch
        [{\roadmapmethodsEtc{a241204440,a250203207,a250818781,a251022431,a251222536,a250808487,a260424842,src260522144,a240809787,a251015831,src260517423,a260416541,src260220664,src260312310,src260405489}},roadmap papers=lthreefill]]
      [{(iii) 3D and multiview renders},roadmap branch
        [{\roadmapmethods{src260312238,src260111109,src260608402,scenethesis,src260805248,a240810453,src260319708,src260613368,src260800629}},roadmap papers=lthreefill]]
      [{(iv) Document and interface\\renders},roadmap branch
        [{\roadmapmethods{pptagent,src260203866,src260104794,src260222839,ui2coden,designcoder,visrefiner,visionguided,src260314724,frontalk}},roadmap papers=lthreefill]]
    ]
    [{\roadmapsection{sec:l3-b}{Structured and Execution\\Feedback}{\S VII.B}},roadmap subsection
      [{(i) Program and workflow state},roadmap branch
        [% Full reviewed list also includes src260719947 and webvia.
         {\roadmapmethodsEtc{src260311048,a250400010,src260103741,src260804071,src260728073,src260602915,talkslides,src260702590,autoslides,src250521660,slideagent,pptarena,src260803298,src260316839}},roadmap papers=lthreefill]]
      [{(ii) Timeline and source state},roadmap branch
        [% Full reviewed list also includes a260329664 and a251212196.
         {\roadmapmethodsEtc{a260526525,a260211790,src260314790,lave,src260802694,a260607636,src260208368,src260404875,src260405076,a250910761}},roadmap papers=lthreefill]]
      [{(iii) Scene and engine state},roadmap branch
        [{\roadmapmethods{src260410383,src260609738,src260614168,src260701766,src260311554,agentic3d,src260723491,src260515187,src260519587,a250112909,src260425318,src260812314,src260606002,src260720889}},roadmap papers=lthreefill]]
    ]
    [{\roadmapsection{sec:l3-c}{Physical and Constraint\\Feedback}{\S VII.C}},roadmap subsection
      [{(i) Simulated dynamics},roadmap branch
        [% Full reviewed list also includes src260423580 and src260721522.
         {\roadmapmethodsEtc{newton,a241010076,a251204221,a260727380}},roadmap papers=lthreefill]]
      [{(ii) Geometric and world\\constraints},roadmap branch
        [% Full reviewed list also includes src260109150 and src260711594.
         {\roadmapmethodsEtc{src260214968,src260210116,src260720866,src260209153,src260806161}},roadmap papers=lthreefill]]
    ]
    [{\roadmapsection{sec:l3-d}{Human Review Feedback}{\S VII.D}},roadmap subsection
      [{(i) In-episode review and\\approval},roadmap branch
        [{\roadmapmethods{a230409337,src260703731,src260701588,src260105016}},roadmap papers=lthreefill]]
    ]
  ]
]
\end{forest}%
\endgroup
}

% L3 is intentionally one Forest tree so A--D all connect to one L3 root.
\newcommand{\roadmappartfour}{}

\newcommand{\roadmappartfive}{%
\begingroup
\let\tablefont\roadmapfont
\begin{forest} roadmap tree
[{},roadmap root
  [{\roadmaplevel{sec:l4}{L4: Experience-Adaptive Control}{\S VIII}},roadmap level,xshift=-1.4em,before drawing tree={y+=2.55em}
    [{\roadmapsection{sec:l4-a}{Capability and Tool\\Profiles}{\S VIII.A}},roadmap subsection
      [{(i) Empirical capability records},roadmap branch
        [{\roadmapmethods{diffusionagent,octot2i,src260122571,src260816721}},roadmap papers=lfourfill]]
    ]
    [{\roadmapsection{sec:l4-b}{Episodic and User\\Memory}{\S VIII.B}},roadmap subsection
      [{(i) Retrieved episodes and\\preferences},roadmap branch
        [{\roadmapmethods{a260603243,a260302816,a251108521,src260617162,src260501477}},roadmap papers=lfourfill]]
    ]
    [{\roadmapsection{sec:l4-c}{Reusable Procedures\\and Skills}{\S VIII.C}},roadmap subsection
      [{(i) Abstracted action procedures},roadmap branch
        [% Full reviewed list also includes scenecraft and src260509423.
         {\roadmapmethodsEtc{genevolve,src260409568,src260630296,src260628971}},roadmap papers=lfourfill]]
    ]
    [{\roadmapsection{sec:l4-d}{Executable Workflows\\and Harnesses}{\S VIII.D}},roadmap subsection
      [{(i) Executable workflow\\revisions},roadmap branch
        [% Full reviewed list also includes a260608091 and avaencoder.
         {\roadmapmethodsEtc{src260701709,src260329590,src260813560}},roadmap papers=lfourfill]]
    ]
    [{\roadmapsection{sec:l4-e}{Policy and Model\\Updates}{\S VIII.E}},roadmap subsection
      [{(i) Persistent behavioral updates},roadmap branch
        [% Full reviewed list also includes src260712042.
         {\roadmapmethodsEtc{a260202051,src251123002,spiral,src260716352}},roadmap papers=lfourfill]]
    ]
  ]
]
\end{forest}%
\endgroup
}

% Compose the four logical level groups into two consistently scaled pages.
% Keeping the levels separate internally preserves horizontal terminal edges,
% while the shared width makes the columns align across each continued page.
\newcommand{\roadmappageone}{%
  \begin{tabular}{@{}c@{}}\roadmappartone\\[2pt]\roadmapparttwo\end{tabular}%
}
\newcommand{\roadmappagetwo}{%
  \begin{tabular}{@{}c@{}}\roadmappartthree\\[2pt]\roadmappartfive\end{tabular}%
}

\begin{figure*}[!t]
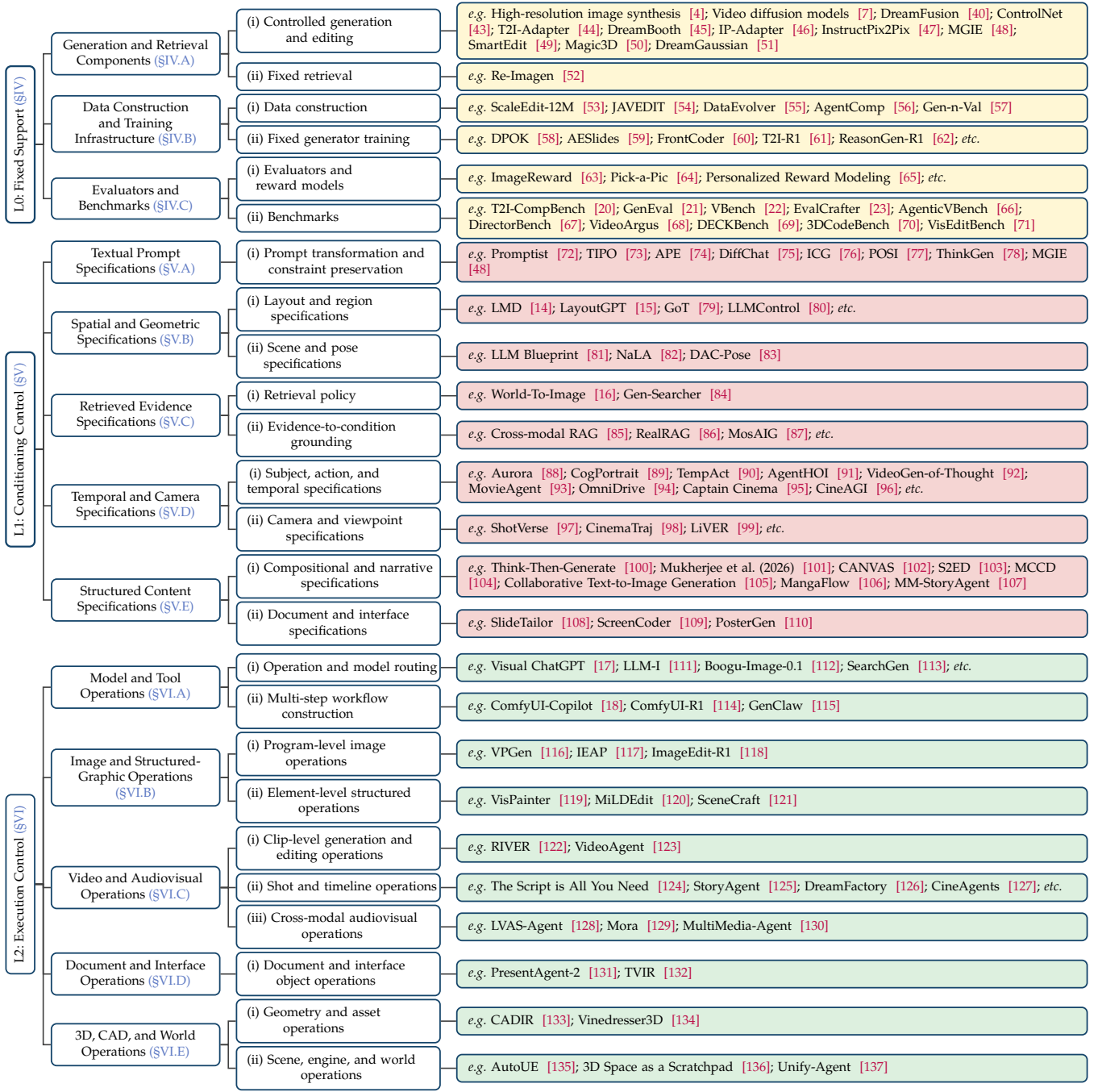

\centering
\resizebox{0.98\textwidth}{!}{\roadmappageone}
\caption{Four-tier taxonomy and roadmap of controller decision-making scope. Part I covers L0--L2.}
\label{fig:fine-taxonomy-roadmap}
\end{figure*}

\begingroup
\renewcommand{\theHfigure}{\arabic{figure}-continued}
\begin{figure*}[!t]
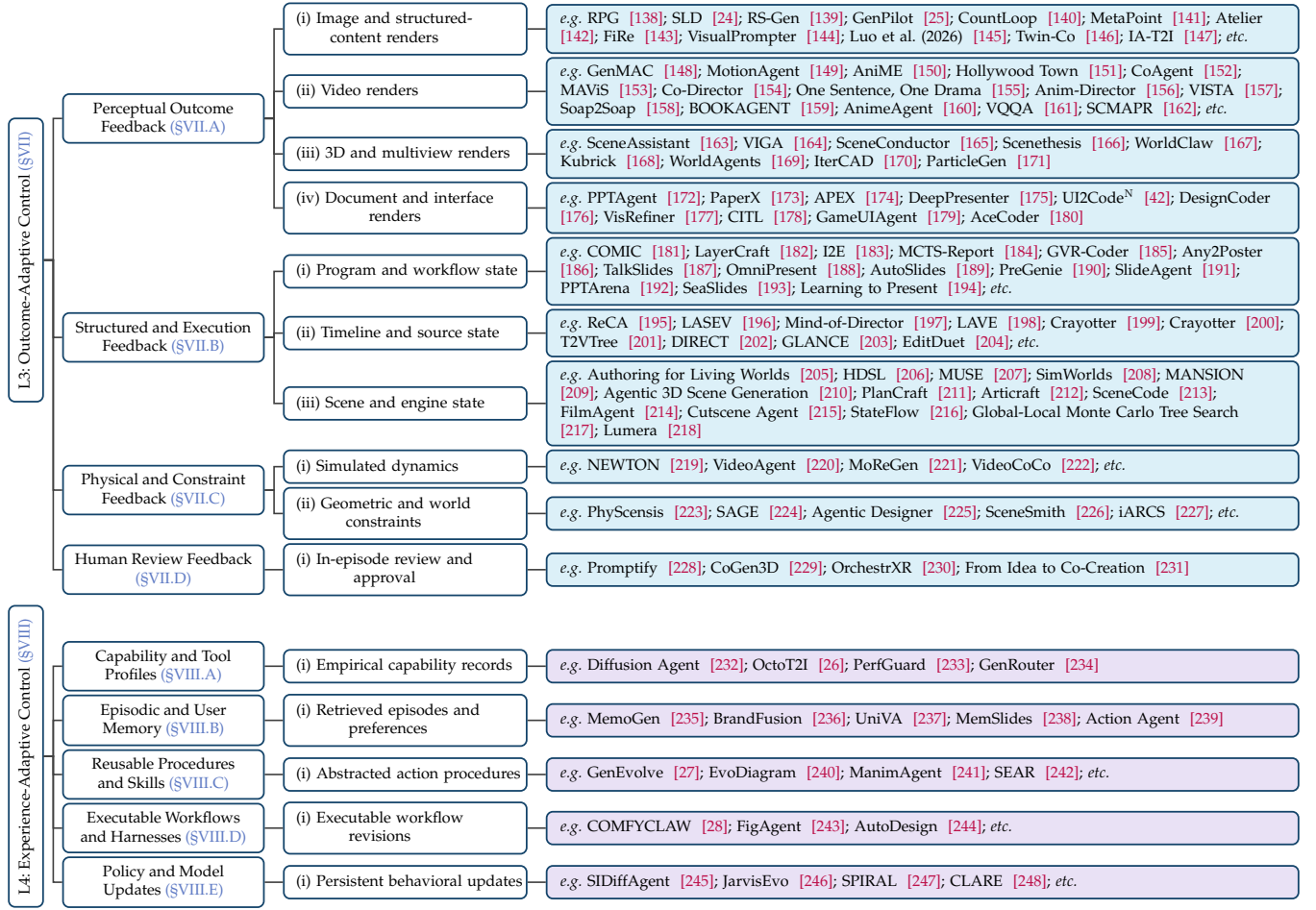

\ContinuedFloat
\centering
\resizebox{0.98\textwidth}{!}{\roadmappagetwo}
\caption{Four-tier taxonomy and roadmap of controller decision-making scope (continued). Part II covers L3--L4.}
\label{fig:fine-taxonomy-roadmap-continued}
\end{figure*}
\endgroup

\subsection{Why a Hierarchy Rather Than Independent Capability Dimensions?}
Planning, tool use, feedback, and memory are useful annotations, but treating them as independent categories does not answer which system can intervene later in the causal chain. Let a system have a capability vector
\begin{equation}
\mathbf{c}(S)=(c_1,c_2,c_3,c_4), \qquad c_i\in\{0,1\},
\end{equation}
where the four entries denote demonstrated control over specification, execution, within-task outcome adaptation, and cross-task experience reuse. We retain this vector as the capability path and define the primary level as
\begin{equation}
L(S)=\max\{i:c_i=1\}.
\end{equation}

Two properties justify this projection: it orders systems by maximum causal reach and applies the same test across visual modalities.

\textbf{Maximum causal reach.} The hierarchy is therefore a projection of a richer design space onto maximum causal reach. Each successive level extends the controller's causal reach to a later decision in the generation trajectory, while output quality and implementation complexity remain separate evaluation dimensions. It makes one comparison legible: how late can evidence still change a future generation decision?

\textbf{Modality independence.} Beyond ordering causal reach, the same projection remains stable across modalities. A storyboard planner and a spatial-layout planner can both be L1 because their decision-making scope terminates at a specification. An image router and a 3D asset selector can both be L2 because they invoke different visual operations. The representation changes, but the causal test does not.

\subsection{Boundary Cases and Conservative Assignment}
Table~\ref{tab:boundary-cases} applies the decision test to cases that are often mislabeled. The primary level follows the demonstrated inference-time decision-making scope, not the language used in a paper title. When evidence for a causal link is absent, we assign the lower level.

Two distinctions are especially important. First, within-task memory is ordinary trajectory state and cannot establish L4. Second, reinforcement learning can optimize a generator, router, repair policy, or memory policy. Its level follows the learned policy's inference-time action space and causal reach, not the optimizer.

\subsection{Quantitative Landscape of the Analyzed Systems}

The hierarchy becomes more informative when applied consistently at corpus scale. We deduplicated systems from seven task collections and annotated first release date, primary level, capability path, modality, mechanism, controller type, visual executor, feedback boundary, evaluation type, cross-task persistence, and public resources. The complete CSV and JSON records, generation script, and summary statistics are released with the project repository.\footnote{Structured corpus: \url{https://github.com/YinmingHuang/Awesome-agentic-visual-generation-model}} Counts below are generated from the same README records rather than transcribed into separate spreadsheets.

\begin{table*}[!t]
\caption{Controller-Capability Taxonomy of Agentic Visual Generation}
\label{tab:taxonomy}
\centering
\setlength{\tabcolsep}{2pt}
\renewcommand{\arraystretch}{1.28}
\begin{tabular}{T{0.06\textwidth} T{0.22\textwidth} T{0.22\textwidth} T{0.24\textwidth} T{0.18\textwidth}}
\toprule
\textbf{Level} & \textbf{Scope} & \textbf{Test question} & \textbf{Controlled variables} & \textbf{Status} \\
\midrule
\rowcolor{gray!8}\textbf{L0} & Fixed Support & Is the inference path predetermined? & Generator, retriever, evaluator, benchmark & Inclusion boundary \\
\textbf{L1} & Conditioning Control & What specification reaches a fixed executor? & Prompt, layout, reference, knowledge, storyboard & Controller level \\
\rowcolor{gray!8}\textbf{L2} & Execution Control & Which visual operation is invoked? & Tool/model, mode, program, call order & Controller level \\
\textbf{L3} & Outcome-Adaptive Control & What action follows an observed result? & Repair, reroute, regenerate, rollback, stop & Controller level \\
\rowcolor{gray!8}\textbf{L4} & Experience-Adaptive Control & What changes on the next task? & Memory, skills, profiles, policy, workflow library & Controller level \\
\bottomrule
\end{tabular}
\end{table*}

\begin{table*}[!t]
\caption{Boundary Cases Under the Maximum-Causal-Reach Rule}
\label{tab:boundary-cases}
\centering
\small
\setlength{\tabcolsep}{2pt}
\renewcommand{\arraystretch}{1.28}
\begin{tabular}{T{0.34\textwidth}T{0.10\textwidth}T{0.46\textwidth}}
\toprule
Behavior & Level & Decisive test \\
\midrule
\rowcolor{gray!15}Prompt rewrite + fixed generator & L1 & Scope ends at the declarative specification \\
Generator/editor choice & L2 & Controller selects the operation that executes \\
\rowcolor{gray!15}
Critique + repair/regeneration & L3 & Outcome changes a later action in the same task \\
Current-task state only & $\leq$L3 & State does not cross the episode boundary \\
\rowcolor{gray!15}
Reusable skill from a completed task & L4 & Experience changes a later task's control \\
RL-trained fixed generator & L0/L1 & Level follows inference-time decisions, not optimization \\
\rowcolor{gray!15}
Multi-planner storyboard + fixed executor & L1 & Role count does not extend causal reach \\
Candidate scoring without follow-up action & L0 & Evaluation provides evidence only \\
\bottomrule
\end{tabular}
\end{table*}

\subsubsection{Capability Evolution Over Time}
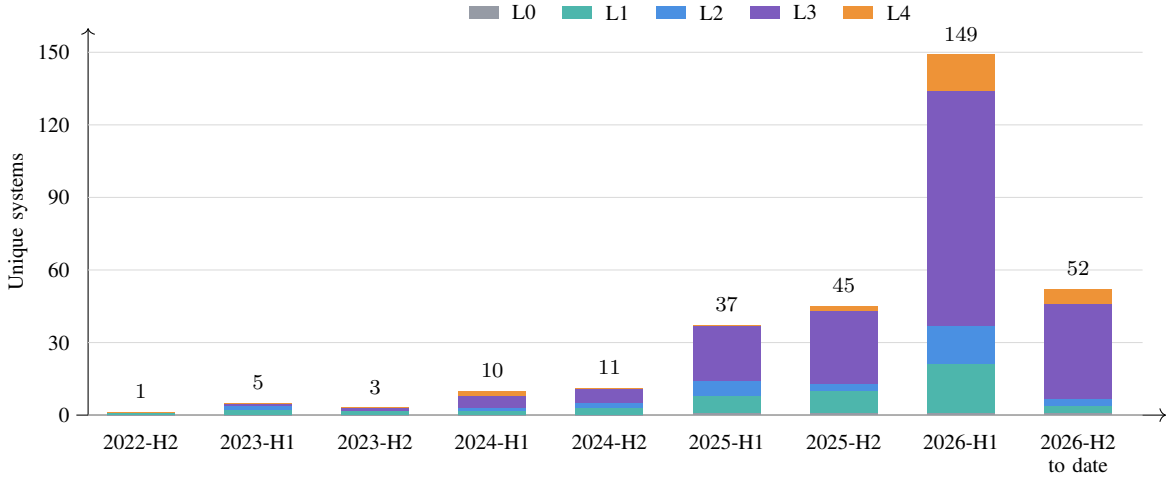
\begin{figure*}[!t]
\centering
\begin{tikzpicture}[x=1.55cm,y=0.032cm]
\draw[->] (0.45,0)--(9.75,0); \draw[->] (0.55,0)--(0.55,160);
\foreach \y in {0,30,60,90,120,150}{\draw[black!14](0.55,\y)--(9.55,\y);\node[anchor=east,font=\footnotesize]at(0.47,\y){\y};}
\def\stackbar#1#2#3#4#5#6#7{%
\pgfmathsetmacro{\a}{#2}\pgfmathsetmacro{\b}{#2+#3}\pgfmathsetmacro{\c}{#2+#3+#4}\pgfmathsetmacro{\d}{#2+#3+#4+#5}\pgfmathsetmacro{\e}{#2+#3+#4+#5+#6}%
\fill[Lzero](#1-0.29,0)rectangle(#1+0.29,\a);\fill[Lone](#1-0.29,\a)rectangle(#1+0.29,\b);\fill[Ltwo](#1-0.29,\b)rectangle(#1+0.29,\c);\fill[Lthree](#1-0.29,\c)rectangle(#1+0.29,\d);\fill[Lfour](#1-0.29,\d)rectangle(#1+0.29,\e);%
\node[anchor=south,font=\footnotesize]at(#1,\e+2){\pgfmathprintnumber{\e}};\node[anchor=north,align=center,font=\footnotesize]at(#1,-4){#7};}
\stackbar{1}{0}{1}{0}{0}{0}{2022-H2}
\stackbar{2}{0}{2}{2}{1}{0}{2023-H1}
\stackbar{3}{0}{2}{0}{1}{0}{2023-H2}
\stackbar{4}{0}{2}{1}{5}{2}{2024-H1}
\stackbar{5}{0}{3}{2}{6}{0}{2024-H2}
\stackbar{6}{1}{7}{6}{23}{0}{2025-H1}
\stackbar{7}{1}{9}{3}{30}{2}{2025-H2}
\stackbar{8}{1}{20}{16}{97}{15}{2026-H1}
\stackbar{9}{1}{3}{3}{39}{6}{2026-H2\\to date}
\node[rotate=90,font=\footnotesize]at(-0.05,80){Unique systems};
\foreach \x/\name/\clr in {3.8/L0/Lzero,4.6/L1/Lone,5.4/L2/Ltwo,6.2/L3/Lthree,7.0/L4/Lfour}{\fill[\clr](\x,164)rectangle(\x+0.25,169);\node[anchor=west,font=\footnotesize]at(\x+0.29,166.5){\name};}
\end{tikzpicture}
\caption{Half-year capability evolution in the structured corpus, grouped by first public release. The 2026-H2 bar contains 52 records released through August 24, 2026. Unlike a total-only trend, the stacked bars expose which kind of controller decision drives growth.}
\label{fig:capability-evolution}
\end{figure*}

Figure~\ref{fig:capability-evolution} replaces a single total-count curve with level composition. The corpus begins with sparse L1 Conditioning Control, while the steep growth after 2025 is dominated by L3 Outcome-Adaptive Control. Only four reviewed records are assigned to L0 Fixed Support, because the corpus was assembled around controller-bearing systems and retains L0 cases only when they resolve an important classification ambiguity. Recent work increasingly treats rendered outputs, execution results, and verifier diagnoses as state for subsequent decisions. In contrast, L4 Experience-Adaptive Control remains a small fraction of the literature, suggesting that persistent experience reuse is substantially less mature than within-task correction.

\subsubsection{Task and Controller Organization}
Temporal growth does not imply uniform adoption across tasks. Figure~\ref{fig:landscape-heatmaps}(a) shows that L3 is prevalent across all modalities, although its operational meaning varies by task. Editing and user-interface tasks expose rendered states that can be inspected and revised, whereas video, 3D, and world tasks require state consistency across time or viewpoints. Image generation has the largest absolute count and also contains most L1 controllers based on prompt construction, layout planning, and retrieval.

\textbf{Controller organization.} Figure~\ref{fig:landscape-heatmaps}(b) uses one architectural axis instead of mixing capabilities, topology, and training methods. Each system receives exactly one primary organization label: a single language or multimodal controller, a multi-role controller, or a unified multimodal policy. Single-controller systems dominate every level. Multi-role organizations occur mainly from L1 through L3, while unified policies remain less common but appear at every controller level. Reinforcement learning is omitted from this comparison because it is an optimization procedure rather than a controller organization and is analyzed separately in Section~\ref{sec:training}.

\begin{figure*}[!t]
\centering
\begin{tikzpicture}[font=\footnotesize]
\colorlet{heatviolet}{violet!58!white}
\colorlet{heatblue}{blue!52!white}
\node[font=\bfseries] at (4.0,3.9) {(a) Modality $\times$ primary level};
\foreach \x/\lab in {0/L0,1/L1,2/L2,3/L3,4/L4}{\node at (2.2+\x,3.45){\lab};}
\foreach \y/\lab in {0/Image,1/Editing,2/Video,3/Slide,4/UI,5/3D,6/World}{\node[anchor=east]at(1.65,3-0.48*\y){\lab};}
\def\mcell#1#2#3#4{\fill[heatviolet!#4](1.75+#2,2.78-0.48*#1)rectangle(2.55+#2,3.18-0.48*#1);\node at(2.15+#2,2.98-0.48*#1){#3};}
\mcell{0}{0}{2}{2}\mcell{0}{1}{30}{28}\mcell{0}{2}{19}{18}\mcell{0}{3}{107}{100}\mcell{0}{4}{15}{14}
\mcell{1}{0}{1}{2}\mcell{1}{1}{4}{8}\mcell{1}{2}{13}{25}\mcell{1}{3}{51}{100}\mcell{1}{4}{4}{8}
\mcell{2}{0}{1}{2}\mcell{2}{1}{16}{38}\mcell{2}{2}{10}{24}\mcell{2}{3}{42}{100}\mcell{2}{4}{6}{14}
\mcell{3}{0}{1}{9}\mcell{3}{1}{1}{9}\mcell{3}{2}{1}{9}\mcell{3}{3}{11}{100}\mcell{3}{4}{1}{9}
\mcell{4}{0}{0}{0}\mcell{4}{1}{1}{14}\mcell{4}{2}{0}{0}\mcell{4}{3}{7}{100}\mcell{4}{4}{0}{0}
\mcell{5}{0}{0}{0}\mcell{5}{1}{2}{6}\mcell{5}{2}{4}{12}\mcell{5}{3}{33}{100}\mcell{5}{4}{2}{6}
\mcell{6}{0}{1}{8}\mcell{6}{1}{3}{23}\mcell{6}{2}{3}{23}\mcell{6}{3}{13}{100}\mcell{6}{4}{2}{15}

\node[font=\bfseries] at (12.65,3.9) {(b) Controller organization $\times$ primary level};
\foreach \x/\lab in {0/L0,1/L1,2/L2,3/L3,4/L4}{\node at (10.85+\x,3.45){\lab};}
\foreach \y/\lab in {0/Single controller,1/Multi-role controller,2/Unified multimodal policy}{\node[anchor=east]at(10.30,2.85-1.12*\y){\lab};}
\def\ccell#1#2#3#4{\fill[heatblue!#4](10.40+#2,2.58-1.12*#1)rectangle(11.20+#2,3.12-1.12*#1);\node at(10.80+#2,2.85-1.12*#1){#3};}
\ccell{0}{0}{4}{2}\ccell{0}{1}{42}{26}\ccell{0}{2}{25}{15}\ccell{0}{3}{163}{100}\ccell{0}{4}{23}{14}
\ccell{1}{0}{0}{0}\ccell{1}{1}{6}{21}\ccell{1}{2}{6}{21}\ccell{1}{3}{28}{100}\ccell{1}{4}{0}{0}
\ccell{2}{0}{0}{0}\ccell{2}{1}{1}{9}\ccell{2}{2}{2}{18}\ccell{2}{3}{11}{100}\ccell{2}{4}{2}{18}
\end{tikzpicture}
\caption{Corpus cross-sections. Panel (a) is multi-label because one system may span generation and editing or multiple output types. Panel (b) assigns each system to exactly one primary controller organization. Cell shading is normalized within each row.}
\label{fig:landscape-heatmaps}
\end{figure*}

\textbf{Implications.} These observations support two conclusions that guide the rest of the paper. First, outcome-conditioned revision is currently the most common form of agentic behavior. Its evaluation must therefore distinguish genuine outcome-driven repair from repeated sampling followed by selection, where apparent improvement may come from the selector rather than from adaptive revision. Second, the scarcity of cross-task persistence makes the transition from L3 to L4 an open research challenge rather than a routine extension. The following sections therefore examine not only what each level can do, but also what evidence is required to establish its additional causal reach.

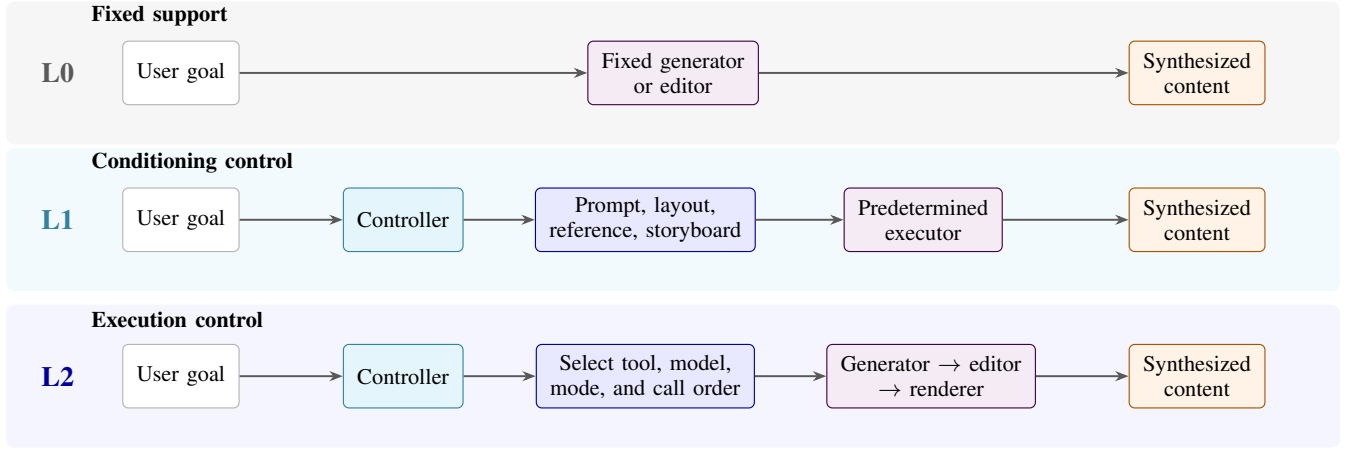
\begin{figure*}[!t]
\centering
\resizebox{0.98\textwidth}{!}{\begin{tikzpicture}[
  font=\footnotesize,
  flow/.style={-{Stealth[length=1.7mm]},line width=0.7pt,draw=black!65},
  box/.style={draw=black!30,rounded corners=2pt,minimum height=8mm,align=center,fill=white,inner xsep=5pt,inner ysep=3pt},
  controller/.style={box,draw=cyan!55!black,fill=cyan!9},
  decision/.style={box,draw=blue!60!black,fill=blue!9},
  executor/.style={box,draw=violet!50!black,fill=violet!8},
  output/.style={box,draw=orange!65!black,fill=orange!10},
  note/.style={font=\scriptsize,align=center,text=black!70}
]
\fill[gray!7,rounded corners=3pt] (0,4.00) rectangle (16.8,5.80);
\fill[cyan!4,rounded corners=3pt] (0,2.15) rectangle (16.8,3.95);
\fill[blue!4,rounded corners=3pt] (0,0.18) rectangle (16.8,1.98);

\node[font=\bfseries\normalsize,text=gray!70!black] at (0.65,4.90) {L0};
\node[font=\bfseries\normalsize,text=cyan!55!black] at (0.65,3.05) {L1};
\node[font=\bfseries\normalsize,text=blue!60!black] at (0.65,1.08) {L2};

\node[box] (l0input) at (2.20,4.90) {User goal};
\node[executor] (l0exec) at (8.40,4.90) {Fixed generator\\or editor};
\node[output] (l0out) at (15.00,4.90) {Synthesized\\content};
\draw[flow] (l0input) -- (l0exec);
\draw[flow] (l0exec) -- (l0out);

\node[box] (l1goal) at (2.20,3.05) {User goal};
\node[controller] (l1ctrl) at (5.00,3.05) {Controller};
\node[decision] (l1spec) at (8.05,3.05) {Prompt, layout,\\reference, storyboard};
\node[executor] (l1exec) at (11.55,3.05) {Predetermined\\executor};
\node[output] (l1out) at (15.00,3.05) {Synthesized\\content};
\draw[flow] (l1goal) -- (l1ctrl);
\draw[flow] (l1ctrl) -- (l1spec);
\draw[flow] (l1spec) -- (l1exec);
\draw[flow] (l1exec) -- (l1out);

\node[box] (l2goal) at (2.20,1.08) {User goal};
\node[controller] (l2ctrl) at (5.00,1.08) {Controller};
\node[decision] (l2route) at (8.05,1.08) {Select tool, model,\\mode, and call order};
\node[executor] (l2ops) at (11.65,1.08) {Generator $\rightarrow$ editor\\$\rightarrow$ renderer};
\node[output] (l2out) at (15.00,1.08) {Synthesized\\content};
\draw[flow] (l2goal) -- (l2ctrl);
\draw[flow] (l2ctrl) -- (l2route);
\draw[flow] (l2route) -- (l2ops);
\draw[flow] (l2ops) -- (l2out);

\node[font=\bfseries\footnotesize,anchor=west] at (0.95,5.62) {Fixed support};
\node[font=\bfseries\footnotesize,anchor=west] at (0.95,3.77) {Conditioning control};
\node[font=\bfseries\footnotesize,anchor=west] at (0.95,1.80) {Execution control};
\end{tikzpicture}}
\caption{From L0 Fixed Support to L2 Execution Control. L0 components such as ControlNet~\cite{controlnet} execute a supplied condition along a predetermined path. L1 Conditioning Control systems such as LayoutGPT~\cite{layoutgpt} construct a generator-facing specification. L2 Execution Control systems such as Visual ChatGPT~\cite{visual_chatgpt} and ComfyUI-Copilot~\cite{src250605010} select and invoke actual visual operations, but remain open-loop when the resulting synthesized content does not redirect the chosen route.}
\label{fig:l0-l2-control-flow}
\end{figure*}

\section{L0: Fixed Support}
\label{sec:l0}

L0 Fixed Support supplies components to later controllers and marks the inclusion boundary rather than a peer agent category. We organize these components by their support function: generation or retrieval, data or training, and evaluation. As Figure~\ref{fig:l0-l2-control-flow} shows, a supplied condition $c$ follows the predetermined inference path
\begin{equation}
y = G(c),
\end{equation}
through a fixed generator or editor $G$. Table~\ref{tab:l0-fine-taxonomy} summarizes the three parallel support functions. The final boundary test is a decision rule rather than a fourth component category.

\begin{table*}[!t]
\caption{Fine-Grained Organization of L0 Fixed Support}
\label{tab:l0-fine-taxonomy}
\centering
\small
\setlength{\tabcolsep}{2pt}
\renewcommand{\arraystretch}{1.28}
\begin{tabular}{T{0.14\textwidth}T{0.17\textwidth}T{0.22\textwidth}T{0.42\textwidth}}
\toprule
\textbf{Category} & \textbf{Subcategory} & \textbf{Controlled decision} & \textbf{Representative systems} \\
\midrule
\rowcolor{gray!15}\cellcolor{white}
& Controlled generation and editing & Execute a supplied condition & Latent Diffusion~\cite{ldm}, Video Diffusion~\cite{video_diffusion}, ControlNet~\cite{controlnet}, InstructPix2Pix~\cite{instruct_pix2pix} \\
\multirow{-2}{=}{\textbf{Generation/\allowbreak retrieval}} & Fixed retrieval & Follow a predetermined retrieval path & Re-Imagen~\cite{re_imagen} \\
\midrule
\rowcolor{gray!15}\cellcolor{white}
& Data construction & Construct or filter records & AgentComp~\cite{a251209081}, Gen-n-Val~\cite{src250604676}, ScaleEdit-12M~\cite{src260320644} \\
\multirow{-2}{=}{\textbf{Data/\allowbreak training}} & Fixed generator training & Update parameters under fixed inference & DPOK~\cite{dpok}, AeSlides~\cite{aeslides}, OSPO~\cite{src250602015} \\
\midrule
\rowcolor{gray!15}\cellcolor{white}
& Evaluators and reward models & Score quality or validity & ImageReward~\cite{imagereward}, GenEval~\cite{geneval}, VBench~\cite{vbench} \\
\multirow{-2}{=}{\textbf{Evaluation}} & Benchmarks & Diagnose capability or failure & AgenticVBench~\cite{src260527705}, DECKBench~\cite{src260213318}, 3DCodeBench~\cite{src260601057} \\
\bottomrule
\end{tabular}
\end{table*}

\subsection{Generation and Retrieval Components}\label{sec:l0-a}
L0 begins with fixed operations that later controllers may invoke. Latent Diffusion Models~\cite{ldm}, Video Diffusion Models~\cite{video_diffusion}, and DreamFusion~\cite{dreamfusion} map supplied conditions to image, video, and 3D representations. They establish what can be executed, but not which operation should be selected in a task. We separate these primitives into controlled generation and editing, which transform a supplied condition, and fixed retrieval, which supplies evidence through a fixed path.

\textbf{Controlled generation and editing.} ControlNet~\cite{controlnet} and T2I-Adapter~\cite{t2i_adapter} execute spatial conditions, while DreamBooth~\cite{dreambooth} and IP-Adapter~\cite{ip_adapter} support subject or reference conditioning. InstructPix2Pix~\cite{instruct_pix2pix}, MGIE~\cite{a230917102}, and SmartEdit~\cite{a231206739} similarly expose instruction-based editing operations. Recent controllable generators broaden this fixed-operation interface: CreatiLayout~\cite{creatilayout} controls layout-to-image synthesis, CreatiDesign~\cite{creatidesign} unifies image, layout, and text conditions for graphic design, MagicMotion~\cite{magicmotion} controls image-to-video trajectories, and Seg2Any~\cite{seg2any} turns open-set segmentation masks into shape- and semantic-controlled images. Generator controllability is not the controller's decision-making scope: these methods apply a supplied condition but do not independently decide whether another operation should be invoked or whether a failed result should be repaired. The same boundary holds for DreamFusion, Magic3D~\cite{magic3d}, and DreamGaussian~\cite{dreamgaussian}, whose internal optimization creates 3D content under a fixed objective.

\textbf{Fixed retrieval.} Beyond direct generation and editing, Re-Imagen~\cite{re_imagen} supplies retrieved image-text examples through a predetermined retrieval-and-generation path. Retrieval can improve the condition presented to a generator, but it remains an L0 component when neither the retrieval result nor the generated output changes a later generation action. A controller enters L1 or above only when it decides how retrieved evidence changes the specification or subsequent route.

\subsection{Data Construction and Training Infrastructure}\label{sec:l0-b}
Before deployment, data pipelines construct supervision and optimization procedures update model parameters. We examine this offline sequence in two stages: data construction creates or filters supervision, and fixed generator training uses that supervision or a reward to update parameters. Either stage can contain tool-using or multi-role processes, but its complexity does not determine the decision-making scope of the resulting inference system.

\textbf{Data construction.} ScaleEdit-12M~\cite{src260320644} uses multiple roles to construct image-editing data, JAVEDIT~\cite{src260603168} curates joint audio-visual editing supervision, and DataEvolver~\cite{src260631537} evolves text-rich image data. AgentComp~\cite{a251209081} builds compositional preference records with tool-using language models, while Gen-n-Val~\cite{src250604676} optimizes prompts and filters synthetic instances. These systems may use agentic processes during curation, but the resulting dataset or fixed generator does not inherit their generation-level decisions.

\textbf{Fixed generator training.} Once supervision or feedback has been constructed, DPOK~\cite{dpok} optimizes diffusion parameters against image rewards, AeSlides~\cite{aeslides} trains a slide generator with verifiable layout rewards, and FrontCoder~\cite{frontcoder} combines pre-training, supervised fine-tuning, and reinforcement learning for frontend generation. T2I-R1~\cite{src250500703}, ReasonGen-R1~\cite{src250524875}, and OSPO~\cite{src250602015} likewise improve image generation through reasoning supervision or preference optimization. WorldCycle~\cite{src260804964} and Ask, Solve, Generate~\cite{src260627376} add self-verifiable or self-consistency rewards. All remain L0 when their deployed inference path does not choose among generation-level actions.

\subsection{Evaluators and Benchmarks}\label{sec:l0-c}
Evaluation infrastructure measures synthesized outputs or diagnoses systems without deciding what the tested generator should do next. We distinguish evaluators and reward models, which assign judgments to synthesized outputs, from benchmarks, which organize tasks and expose capability failures. Either can support training or a later controller, but neither establishes an L1--L4 decision-making scope by itself.

\textbf{Evaluators and reward models.} ImageReward~\cite{imagereward} scores text-image alignment and preference, Pick-a-Pic~\cite{pickapic} supplies human preference data, and PIGReward~\cite{a251119458} derives personalized evaluation dimensions. CIGEval~\cite{a250407046} organizes conditional-image assessment through agentic evaluators. These components become part of an L3 loop only when another mechanism maps their judgment to a later revision, rerouting, or stopping action.

\textbf{Benchmarks.} Whereas evaluators produce scores or judgments, benchmarks define the cases over which those judgments are interpreted. T2I-CompBench~\cite{t2i_compbench} and GenEval~\cite{geneval} test compositional image requirements, while VBench~\cite{vbench} and EvalCrafter~\cite{evalcrafter} evaluate video quality. AgenticVBench~\cite{src260527705}, DirectorBench~\cite{src260530090}, and VideoArgus~\cite{src260805485} diagnose video-generation and editing systems. DECKBench~\cite{src260213318}, 3DCodeBench~\cite{src260601057}, and VisEditBench~\cite{src260810408} extend evaluation to slides, procedural 3D modeling, and visualization-code editing. WeEdit~\cite{weedit} specifically benchmarks text-centric image editing through instruction adherence, text clarity, and background preservation. Their measurements characterize capabilities and failures, but do not select the system's next generation action.

\subsection{Discussion: Boundary Test for L0 Fixed Support}\label{sec:l0-d}
The final test applies to the complete deployed decision process rather than to an individual component. We first examine fixed pipelines and aggregation, then apply a counterfactual rule that asks whether the deployed system could choose another generation action. A multi-stage pipeline remains L0 when every transition is prescribed, even if it contains retrieval, rewards, several roles, or repeated sampling.

\textbf{Fixed pipelines and aggregation.} Slide Translation~\cite{slidetranslation} generates several layout candidates and selects one, but the selected score does not trigger another layout operation. ShareVerse~\cite{a260302697} combines collaborative attention and spatiotemporal retrieval inside a distributed video generator without exposing a policy that chooses later generation actions. Fixed candidate ranking and internal state can improve synthesized content without creating a generation controller.

\textbf{Counterfactual decision rule.} To distinguish those fixed pipelines from controller-bearing systems, ask whether the deployed system can choose among materially different generation actions given the same goal and state. If no such choice exists, it remains L0. If it decides only what condition to supply to a predetermined executor, it enters L1. Selecting and invoking an operation establishes L2, using an observed outcome to change a later current-task action establishes L3, and persistent cross-task updates establish L4. This rule separates the supporting infrastructure described above from the controller levels that follow.

\section{L1: Conditioning Control}
\label{sec:l1}

L1 Conditioning Control begins where L0 Fixed Support ends. L0 can provide a fixed generator, retriever, evaluator, or training component, but its deployed path does not decide how the task specification should change. This fixed path leaves ambiguity in the prompt, spatial arrangement, retrieved evidence, or temporal structure to the generator itself. L1 adds a pre-execution specification policy that translates the goal into an explicit condition before invoking that fixed executor. This reduces ambiguity and makes constraints inspectable, but it still commits before observing the generated result. When the system must choose and invoke a different operation after that commitment, the problem moves to L2 Execution Control.

L1 methods share the interface $c=\pi(g)$ followed by $y=G(c)$, but differ in the primary generator-facing specification they commit to an executor. We distinguish textual prompts, spatial and geometric structure, retrieved evidence, temporal and camera controls, and structured content descriptions. Reasoning, multimodal input, and role decomposition are ways to construct these specifications rather than peer categories. Figure~\ref{fig:l0-l2-control-flow} locates the commitment before execution, and Table~\ref{tab:l1-fine-taxonomy} summarizes the five specification types.

\begin{table*}[!t]
\caption{Categories and Subcategories of L1 Conditioning Control}
\label{tab:l1-fine-taxonomy}
\centering
\small
\setlength{\tabcolsep}{2pt}
\renewcommand{\arraystretch}{1.28}
\begin{tabular}{T{0.14\textwidth}T{0.17\textwidth}T{0.22\textwidth}T{0.42\textwidth}}
\toprule
\textbf{Category} & \textbf{Subcategory} & \textbf{Controlled decision} & \textbf{Representative systems} \\
\midrule
\rowcolor{gray!15}\cellcolor{white}
\textbf{Text prompt} & Prompt transformation and constraint preservation & Produce model-ready text while retaining intent and policy constraints & Promptist~\cite{a221209611}, TIPO~\cite{a241108127}, DiffChat~\cite{a240304997}, POSI~\cite{a240210882} \\
\midrule
& Layout and region specifications & Commit boxes, regions, or spatial relations & LMD~\cite{llm_grounded}, LayoutGPT~\cite{layoutgpt}, LLMControl~\cite{a250719939} \\
\rowcolor{gray!15}\cellcolor{white}
\multirow{-2}{=}{\textbf{Spatial/\allowbreak geometric}} & Scene and pose specifications & Commit scene structure, assets, or poses & LLM Blueprint~\cite{a231010640}, NaLA~\cite{src260629395}, DAC-Pose~\cite{src260804622} \\
\midrule
& Retrieval policy & Decide whether, where, and what to retrieve & World-To-Image~\cite{world_to_image}, Gen-Searcher~\cite{a260328767} \\
\rowcolor{gray!15}\cellcolor{white}
\multirow{-2}{=}{\textbf{Retrieved\allowbreak evidence}} & Evidence-to-condition grounding & Associate evidence with the intended entity or attribute & Cross-modal RAG~\cite{a250521956}, RealRAG~\cite{a250200848}, MosAIG~\cite{src250215972} \\
\midrule
& Subject, action, and temporal specifications & Commit motion, identity, and action progression & Aurora~\cite{a260518748}, TempAct~\cite{src260628016}, AgentHOI~\cite{src260722241} \\
\rowcolor{gray!15}\cellcolor{white}
\multirow{-2}{=}{\textbf{Temporal/\allowbreak camera}} & Camera and viewpoint specifications & Commit camera paths and shot continuity & ShotVerse~\cite{a260311421}, CinemaTraj~\cite{src260726910} \\
\midrule
& Compositional and narrative specifications & Commit semantic, story, panel, or storyboard structure & Think-Then-Generate~\cite{src260110332}, CANVAS~\cite{src260413452}, S2ED~\cite{src260522448} \\
\rowcolor{gray!15}\cellcolor{white}
\multirow{-2}{=}{\textbf{Structured\allowbreak content}} & Document and interface specifications & Commit slide, poster, or interface structure & SlideTailor~\cite{slidetailor}, ScreenCoder~\cite{screencoder}, PosterGen~\cite{src250817188} \\
\bottomrule
\end{tabular}
\end{table*}

\subsection{Textual Prompt Specifications}\label{sec:l1-a}
A textual specification transforms the user's request into model-facing language while committing before any generated output is observed. Generator compatibility, user preference, and safety are competing objectives of this transformation, not separate types of controller action.

\textbf{Prompt transformation and constraint preservation.} Promptist~\cite{a221209611} establishes learned prompt adaptation by optimizing model-facing text for aesthetic quality while preserving semantic intent. TIPO~\cite{a241108127} instead uses lightweight text presampling to expand prompts toward the language distribution of text-to-image data, and APE~\cite{a260600204} trains compact prompt enhancers for generation and editing. Instruction-oriented systems change the objective rather than the action type. DiffChat~\cite{a240304997} modifies prompts from explicit user instructions, ICG~\cite{a260527374} injects personalized preference context, and POSI~\cite{a240210882} rewrites unsafe prompts while retaining requested semantics. ThinkGen and MGIE further use multimodal reasoning to construct model-ready generation or editing instructions~\cite{a251223568,a230917102}. All belong to one branch because their controlled object is the textual condition; rewards and multimodal inputs determine how that condition is constructed.

\subsection{Spatial and Geometric Specifications}\label{sec:l1-b}
Spatial specifications externalize geometry that a text encoder may represent weakly. We distinguish explicit layouts and regions from scene or pose descriptions, while treating reasoning and role organization as construction mechanisms.

\textbf{Layout and region specifications.} LMD~\cite{llm_grounded} converts a complex request into object descriptions and bounding boxes that guide a frozen diffusion model. LayoutGPT~\cite{layoutgpt} generalizes coordinate prediction to 2D and 3D arrangements through in-context examples, while GoT and LLMControl provide related spatial planning interfaces~\cite{got,a250719939}. Regional-Aware text-to-image Generation (RAG)~\cite{a241106558} binds attributes to localized regions so that composition constraints remain explicit. These methods commit the arrangement before rendering, so an omitted object or invalid region remains a specification error rather than an outcome-driven repair.

\textbf{Scene and pose specifications.} LLM Blueprint~\cite{a231010640} represents an intended scene through a structured description. NaLA~\cite{src260629395} combines language with native 3D asset features to construct an arrangement, and DAC-Pose~\cite{src260804622} specializes the condition to pose-guided human composition. Although these systems use different modalities and planners, their primary commitment is geometric: the executor receives a scene, asset, or pose specification that fixes spatial relations before generation.

\subsection{Retrieved Evidence Specifications}\label{sec:l1-c}
Retrieved evidence supplies facts or appearances absent from the fixed generator. The two decisions form a clear sequence: first choose whether and what to retrieve, then ground the selected evidence in the generator-facing condition.

\textbf{Retrieval policy.} World-To-Image~\cite{world_to_image} probes whether a concept failure reflects missing knowledge and retrieves definitions or reference images only when the generator lacks adequate coverage. Gen-Searcher~\cite{a260328767} learns a multi-hop text and image search policy with textual and visual rewards. Both control the evidence acquisition step before rendering, rather than repairing a generated output afterward.

\textbf{Evidence-to-condition grounding.} Cross-modal RAG~\cite{a250521956} decomposes a request into evidence roles, while RealRAG~\cite{a250200848} aligns retrieved content with realistic synthesis. MosAIG~\cite{src250215972} shows why culturally specific cues must remain associated with their intended concepts. ORIG~\cite{src251022521} combines textual facts and visual references, whereas WMGen-v1~\cite{src260620764} studies how a single image can condition image and world construction. These methods are classified by the retrieved evidence they commit, not by whether retrieval uses text, images, or multimodal reasoning.

\subsection{Temporal and Camera Specifications}\label{sec:l1-d}
Temporal specifications describe how content evolves, while camera specifications describe how that content is viewed. Both are committed before the predetermined video or scene executor produces an outcome.

\textbf{Subject, action, and temporal specifications.} Aurora~\cite{a260518748} converts an underspecified editing request into text, references, and spatial grounding for one diffusion transformer. CogPortrait, TempAct, and AgentHOI construct portrait motion, action progression, or ordered human-object contacts~\cite{src260528056,src260628016,src260722241}. VideoGen-of-Thought~\cite{a241202259} adds shot structure and identity constraints. MovieAgent and OmniDrive use multiple roles to construct scripts, scene descriptions, and shared temporal world conditions~\cite{a250307314,src260617536}. Captain Cinema, CineAGI, Sima 1.0, and InfinityStory similarly commit keyframes, character constraints, documentary structure, or shot transitions before generation~\cite{src250718634,src260423579,src260407721,src260303646}. Their role count is secondary to the temporal specification that reaches the executor.

\textbf{Camera and viewpoint specifications.} ShotVerse~\cite{a260311421} maps language into globally aligned camera trajectories across shots. CinemaTraj~\cite{src260726910} grounds cinematic paths in a reconstructed 3D scene and optimizes permitted parameters against collision and occlusion constraints. LiVER and Camera Artist similarly commit illumination, viewpoint, or cinematic camera language before synthesis~\cite{src260407966,a260409195}. These methods separate subject evolution from scene observation.

\subsection{Structured Content Specifications}\label{sec:l1-e}
Some generators consume an explicit semantic or document structure rather than only text, geometry, or retrieved evidence. We separate compositional and narrative structures from schemas tied to editable documents and interfaces.

\textbf{Compositional and narrative specifications.} Think-Then-Generate and an offline-RL styling planner construct semantic or style decisions before image synthesis~\cite{src260110332,src260307148}. CANVAS and S2ED represent story progression through storyboard or executable descriptions~\cite{src260413452,src260522448}, while MCCD and collaborative text-to-image generation coordinate roles to produce a compositional condition~\cite{a250502648,a251010633}. MangaFlow and MM-StoryAgent add story-section, panel-layout, reference, or modality-specific conditions before generation~\cite{src260528173,a250305242}. These systems are grouped by the structured content description they produce, not by reasoning or multi-role architecture.

\textbf{Document and interface specifications.} SlideTailor~\cite{slidetailor} commits slide content and layout preferences to a fixed realization path. ScreenCoder~\cite{screencoder} constructs a grounded interface hierarchy and code specification, while PosterGen~\cite{src250817188} turns a paper into content, layout, and style specifications for a poster. In each case the structure is modality-specific and established before rendered feedback could redirect execution.

\subsection{Discussion: Commitments, Failure Modes, and Level Boundaries}\label{sec:l1-f}
Taken together, L1 methods share a central tradeoff: richer conditions reduce ambiguity and expose constraints before expensive sampling, but they also commit the system more strongly to decisions made before generation. We discuss this tradeoff through two connected questions: which failures arise at the condition interface and how portable each representation is across generators.

\textbf{Failure modes.} The analyzed methods reveal three distinct sources of conditioning error. Prompt optimizers can distort intent while making language more compatible with a generator. Spatial planners can omit an entity or encode an impossible arrangement. Retrieval systems can introduce irrelevant evidence that the generator follows faithfully. These failures require different diagnostics. Semantic preservation should be measured for Promptist-like rewriting, geometric validity for LayoutGPT-like planning, and evidence relevance for World-To-Image-like retrieval. A single final-image preference score cannot identify which interface failed.

\textbf{Portability.} Beyond diagnosing condition errors, representation choice determines how much control survives generator replacement. Natural-language prompts are broadly portable but weakly binding. Boxes, trajectories, and scene descriptions expose stronger constraints, yet they depend on a generator that understands the chosen control format. LMD and LLMControl illustrate this tradeoff for spatial control, while AgentHOI and CinemaTraj expose it for temporal and camera conditions. A useful L1 comparison should therefore report both within-generator quality and transfer across generator families.

Training changes how a specification is constructed, but not where that specification enters the workflow. Promptist, APE, Gen-Searcher, and learned motion planners optimize different condition policies, yet all must commit before observing the generated result. This limitation motivates execution control, where the controller can choose and invoke visual operations, and outcome-adaptive control, where observed results can redirect later actions.

\section{L2: Execution Control}
\label{sec:l2}

L2 Execution Control changes the action interface from a specification to an executable choice $a_t\sim\pi(a_t\mid g,s_t)$. We classify systems by the primary executable object that the controller selects, invokes, or composes before observing its outcome: model or tool operations, image or structured-graphic operations, video or audiovisual operations, document or interface operations, and 3D, CAD, or world operations. Workflow construction and role coordination are implementation strategies within these five action spaces. Figure~\ref{fig:l0-l2-control-flow} contrasts L2 Execution Control with L0 Fixed Support and L1 Conditioning Control, while Table~\ref{tab:l2-fine-taxonomy} summarizes the executable objects.

\begin{table*}[!t]
\caption{Categories and Subcategories of L2 Execution Control}
\label{tab:l2-fine-taxonomy}
\centering
\small
\setlength{\tabcolsep}{2pt}
\renewcommand{\arraystretch}{1.28}
\begin{tabular}{T{0.11\textwidth}T{0.18\textwidth}T{0.22\textwidth}T{0.44\textwidth}}
\toprule
\textbf{Category} & \textbf{Subcategory} & \textbf{Controlled decision} & \textbf{Representative systems} \\
\midrule
\rowcolor{gray!15}\cellcolor{white}
& Operation and model routing & Select a generator, editor, search tool, or mode & Visual ChatGPT~\cite{visual_chatgpt}, LLM-I~\cite{a250913642}, SearchGen~\cite{searchgen} \\
\multirow{-2}{=}{\textbf{Model/tool}} & Multi-step workflow construction & Compose executable tool, graph, or code operations & ComfyUI-Copilot~\cite{src250605010}, ComfyUI-R1~\cite{src250609790}, GenClaw~\cite{a260530248} \\
\midrule
\rowcolor{gray!15}\cellcolor{white}
& Program-level image operations & Select ordered generation or editing calls & VPGen~\cite{a230515328}, IEAP~\cite{a250604158}, ImageEdit-R1~\cite{src260308059} \\
\multirow{-2}{=}{\textbf{Image/\allowbreak graphic}} & Element-level structured operations & Operate on layers, vectors, objects, or graphs & VisPainter~\cite{src251027452}, MiLDEdit~\cite{src260104589}, SceneCraft~\cite{src260616103} \\
\midrule
& Clip-level generation and editing operations & Select operations over a clip or media state & VideoAgent~\cite{src260623327}, RIVER~\cite{a251114100} \\
\rowcolor{gray!15}\cellcolor{white}
& Shot and timeline operations & Compose scene, shot, or timeline calls & The Script is All You Need~\cite{src260117737}, DreamFactory~\cite{a240811788}, ViMax~\cite{a260607649} \\
\multirow{-3}{=}{\textbf{Video/\\audiovisual}} & Cross-modal audiovisual operations & Coordinate video, audio, and related media tools & LVAS-Agent~\cite{a250310719}, Mora~\cite{a240313248}, MultiMedia-Agent~\cite{src260103250} \\
\midrule
\rowcolor{gray!15}\cellcolor{white}
\textbf{Document/\allowbreak interface} & Document and interface object operations & Invoke operations over document or interface state & PresentAgent-2~\cite{src260511363}, TVIR~\cite{src260602320} \\
\midrule
& Geometry and asset operations & Execute typed geometry or appearance edits & CADIR~\cite{src260800891}, Vinedresser3D~\cite{src260219542} \\
\rowcolor{gray!15}\cellcolor{white}
\multirow{-2}{=}{\textbf{3D/CAD/\allowbreak world}} & Scene, engine, and world operations & Dispatch assets, views, and engine commands & AutoUE~\cite{src260307106}, 3D Space as a Scratchpad~\cite{src260114602}, Unify-Agent~\cite{a260329620} \\
\bottomrule
\end{tabular}
\end{table*}

\subsection{Model and Tool Operations}\label{sec:l2-a}
This action space is organized by whether the controller selects one operation or composes several operations into an executable route. The selected tools can serve any visual modality, but the primary controlled object is the tool invocation itself.

\textbf{Operation and model routing.} Visual ChatGPT~\cite{visual_chatgpt} exposes visual foundation models as callable operations and lets a language controller select a model and construct its arguments. LLM-I~\cite{a250913642} broadens the route to search, code, generation, and editing, while Boogu-Image-0.1~\cite{src260713125} learns operation selection under an action budget. SearchGen~\cite{searchgen} decides whether missing generator knowledge justifies text or image retrieval, and Mind-Brush~\cite{a260201756} selects between generation and editing according to user intent. These systems differ in available tools, but all control which executable operation acts next.

\textbf{Multi-step workflow construction.} ComfyUI-Copilot~\cite{src250605010} constructs executable workflow graphs whose nodes expose data-flow and compatibility constraints. ComfyUI-R1~\cite{src250609790} trains reasoning models to generate such node-based workflows, while GenClaw~\cite{a260530248} uses code-driven canvas operations as its action language. Unlike single-step routing, these systems commit an ordered graph or program before generated outcomes can redirect it.

\subsection{Image and Structured-Graphic Operations}\label{sec:l2-b}
Image generation and editing make the action representation visible at the output level. In addition to selecting a generator, the controller must specify what an operation changes and which structures remain editable. We therefore distinguish executable edit programs from persistent editable structure.

\textbf{Program-level image operations.} Programmatic editing turns a request into ordered operations before they modify pixels. VPGen~\cite{a230515328} establishes an executable representation for generation and evaluation, while IEAP~\cite{a250604158} specializes it to ordered diffusion operations. ImageEdit-R1~\cite{src260308059} learns decomposition and sequencing roles that produce an executable edit program. These systems control operation order rather than only supplying one global prompt.

\textbf{Element-level structured operations.} VisPainter~\cite{src251027452} converts raster intent into vector-oriented operations, while MiLDEdit~\cite{src260104589} exposes layer-level actions over individual design elements. SceneCraft~\cite{src260616103} uses a graph whose nodes dispatch generators and editors over explicit content state. Persistent editability is a consequence of operating on elements, layers, vectors, or graph nodes, rather than a separate action-space category.

\subsection{Video and Audiovisual Operations}\label{sec:l2-c}
Video action spaces differ by the temporal and modal extent of the executable object. We separate operations over one clip, operations that organize several shots or timeline units, and operations that coordinate video with audio or other media.

\textbf{Clip-level generation and editing operations.} RIVER~\cite{a251114100} converts an edit request into an ordered sequence of executable operations. VideoAgent connects video understanding to generation and editing tools over the current media state~\cite{src260623327}. These controllers choose how a clip is transformed, but their route does not depend on inspecting the resulting clip.

\textbf{Shot and timeline operations.} The Script is All You Need~\cite{src260117737} lets a director orchestrate generation calls across scenes, while StoryAgent and DreamFactory construct and execute storyboard or multi-scene workflows~\cite{a241104925,a240811788}. CineAgents~\cite{a260410456} grounds a blueprint in narrative memory before one-pass tool assembly. ViMax~\cite{a260607649} orders screenwriting, shot planning, character styling, and generation operations, with a fixed candidate-selection stage that does not trigger repair. The shared controlled object is an executable sequence over shots or scenes.

\textbf{Cross-modal audiovisual operations.} LVAS-Agent~\cite{a250310719} coordinates planning and audio-generation operations over an existing temporal structure. Mora and MultiMedia-Agent expose reusable generation and editing operations across media types~\cite{a240313248,src260103250}. These systems are classified here because synchronization across media organizes the route, not because they use a multi-role architecture.

\subsection{Document and Interface Operations}\label{sec:l2-d}
This category covers executable actions over document or interface objects. Systems that only construct a slide layout, poster schema, or interface hierarchy remain in L1, while systems whose rendered or browser outcomes trigger revision belong in L3.

\textbf{Document and interface object operations.} PresentAgent-2~\cite{src260511363} invokes research, media, and presentation operations in a selected workflow. TVIR~\cite{src260602320} similarly coordinates retrieval and visual operations to construct an interleaved report. Their controller acts on structured content objects before observing a rendered result, which separates them from L1 specification construction and L3 browser-driven repair.

\subsection{3D, CAD, and World Operations}\label{sec:l2-e}
Three-dimensional action spaces expose either geometry and assets or a broader scene and engine state. Both categories require executable operations, but neither uses an observed outcome to redirect the current route.

\textbf{Geometry and asset operations.} CADIR~\cite{src260800891} provides a cross-backend intermediate representation whose typed features and dependencies support localized CAD operations. Vinedresser3D~\cite{src260219542} decomposes a text-guided 3D edit into staged geometry and appearance actions. Both organize execution around geometry or asset state before rendering.

World construction makes the L2 to L3 boundary especially visible because executable constraints can become observations. A controller is L2 when it chooses environment-building operations but follows the chosen route without inspecting their results. It becomes L3 when collision, reachability, compilation, rendering, or engine state triggers a different construction action.

\textbf{Scene, engine, and world operations.} AutoUE~\cite{src260307106} allocates executable tasks inside Unreal Engine, while 3D Space as a Scratchpad~\cite{src260114602} externalizes spatial reasoning through editable scene operations. Unify-Agent~\cite{a260329620} combines search-grounded context with world-building actions when the available generator lacks a requested entity. The primary executable object is the scene or engine workspace rather than an individual geometric feature.

\subsection{Discussion: Cross-Domain Execution and the Open-Loop Boundary}\label{sec:l2-f}
L2 systems show that a long plan, many roles, and executable state do not by themselves establish feedback. The level changes only when an observation returned by execution affects the next generation-level action.

Across L2, action abstraction is the main architectural choice. Flat tool interfaces, as in Visual ChatGPT, make heterogeneous models immediately accessible but leave compatibility implicit in language descriptions. Workflow graphs, as in ComfyUI-Copilot, expose data types and dependencies. Code and domain-specific languages, as in GenClaw and CADIR, provide stronger execution guarantees but require the controller to satisfy a formal syntax. Structured content models occupy a middle ground because PowerPoint objects, layers, browser components, and scene graphs are executable while remaining aligned with visual concepts. These representations can be reused across image, video, document, interface, and 3D tasks, but that cross-domain reuse is a property of the action space rather than a separate L2 category.

Role decomposition and tool routing solve different problems. A router decides which visual operation should act, whereas a multi-role workflow partitions expertise and state. SearchGen focuses on whether external evidence is necessary for the selected generator. Combining role decomposition with routing can improve specialization, but additional role communication also increases latency and coordination cost.

The open-loop boundary has practical consequences for evaluation. L2 studies should verify plan validity, tool arguments, execution success, and state preservation even when the terminal output looks plausible. A workflow can obtain a strong image from an accidental route, or it can produce a weak image despite selecting the correct capabilities. Reporting both trajectory correctness and output quality separates controller failure from generator failure. It also provides the baseline needed to demonstrate L3: the same tools and initial plan should improve when rendered feedback is allowed to redirect execution.

\section{L3: Outcome-Adaptive Control}
\label{sec:l3}

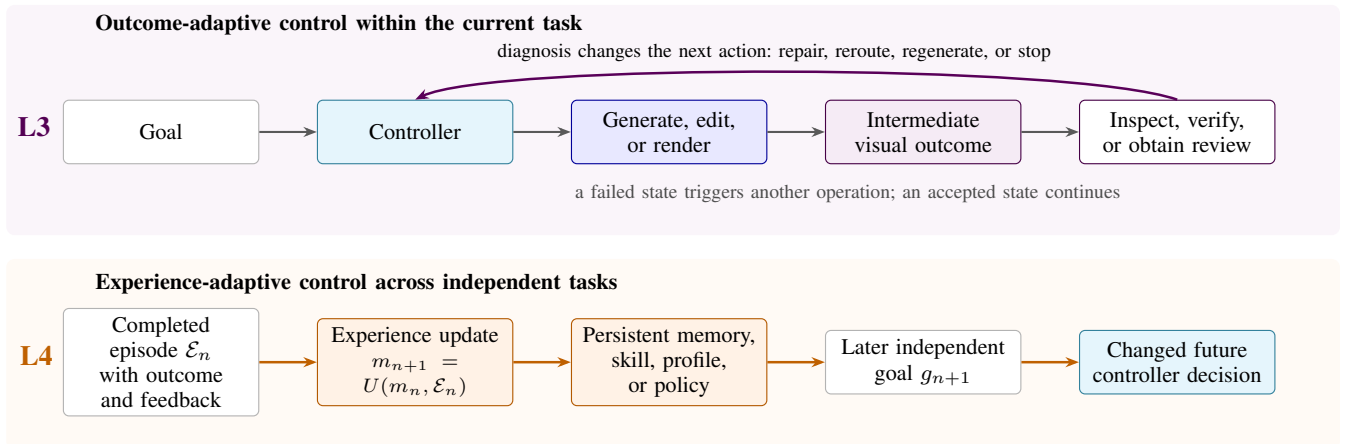
\begin{figure*}[!t]
\centering
\resizebox{0.98\textwidth}{!}{\begin{tikzpicture}[
  font=\footnotesize,
  flow/.style={-{Stealth[length=1.7mm]},line width=0.75pt,draw=black!65},
  feedback/.style={-{Stealth[length=1.7mm]},line width=0.9pt,draw=violet!70!black},
  transfer/.style={-{Stealth[length=1.7mm]},line width=0.9pt,draw=orange!75!black},
  box/.style={draw=black!30,rounded corners=2pt,minimum height=8mm,text width=2.25cm,align=center,fill=white,inner xsep=3pt,inner ysep=3pt},
  controller/.style={box,draw=cyan!55!black,fill=cyan!9},
  operation/.style={box,draw=blue!60!black,fill=blue!9},
  output/.style={box,draw=violet!55!black,fill=violet!8},
  memory/.style={box,draw=orange!70!black,fill=orange!10},
  note/.style={font=\scriptsize,align=center,text=black!70}
]
\fill[violet!4,rounded corners=3pt] (0,2.75) rectangle (16.8,5.65);
\fill[orange!4,rounded corners=3pt] (0,0.05) rectangle (16.8,2.45);

\node[font=\bfseries\normalsize,text=violet!65!black] at (0.35,4.12) {L3};
\node[font=\bfseries\footnotesize,anchor=west] at (1.00,5.40) {Outcome-adaptive control within the current task};
\node[box] (l3goal) at (1.95,4.05) {Goal};
\node[controller] (l3ctrl) at (5.15,4.05) {Controller};
\node[operation] (l3op) at (8.35,4.05) {Generate, edit,\\or render};
\node[output] (l3out) at (11.55,4.05) {Intermediate\\visual outcome};
\node[box,draw=violet!55!black] (l3obs) at (14.75,4.05) {Inspect, verify,\\or obtain review};
\draw[flow] (l3goal) -- (l3ctrl);
\draw[flow] (l3ctrl) -- (l3op);
\draw[flow] (l3op) -- (l3out);
\draw[flow] (l3out) -- (l3obs);
\draw[feedback] (l3obs.north) .. controls (13.40,4.92) and (6.10,4.92) ..
  node[above,pos=0.51,font=\scriptsize,align=center]{diagnosis changes the next action: repair, reroute, regenerate, or stop}
  (l3ctrl.north);
\node[note] at (10.60,3.28) {a failed state triggers another operation; an accepted state continues};

\node[font=\bfseries\normalsize,text=orange!75!black] at (0.38,1.23) {L4};
\node[font=\bfseries\footnotesize,anchor=west] at (1.00,2.14) {Experience-adaptive control across independent tasks};
\node[box] (episode) at (1.95,1.15) {Completed episode $\mathcal{E}_n$\\with outcome and feedback};
\node[memory] (update) at (5.15,1.15) {Experience update\\{\scriptsize $m_{n+1}=U(m_n,\mathcal{E}_n)$}};
\node[memory] (store) at (8.35,1.15) {Persistent memory,\\skill, profile, or policy};
\draw[transfer] (episode) -- (update);
\draw[transfer] (update) -- (store);
\node[box] (nextgoal) at (11.55,1.15) {Later independent\\goal $g_{n+1}$};
\node[controller] (nextctrl) at (14.75,1.15) {Changed future\\controller decision};
\draw[transfer] (store) -- (nextgoal);
\draw[transfer] (nextgoal) -- (nextctrl);
\end{tikzpicture}}
\caption{The distinction between L3 Outcome-Adaptive Control and L4 Experience-Adaptive Control. L3 systems such as GenArtist~\cite{genartist} and PPTAgent~\cite{pptagent} use a generated or rendered outcome to choose a later action in the same task. L4 systems such as GenEvolve~\cite{genevolve} and COMFYCLAW~\cite{src260701709} transform completed-task evidence into persistent procedures or workflow skills that change control on a later independent task.}
\label{fig:l3-l4-adaptation-flow}
\end{figure*}

L3 adds the outcome-dependent state transition visualized in Figure~\ref{fig:l3-l4-adaptation-flow}. After $a_t$ produces $o_{t+1}$, the controller updates $s_{t+1}=F(s_t,a_t,o_{t+1})$ before selecting the next action. We classify systems by the decisive observation that triggers the next current-task action: a perceptual outcome, structured or execution state, physical or hard-constraint result, or in-episode human review. Search, iterative repair, rerouting, stopping, and state preservation describe how a controller responds to feedback and are therefore secondary strategies. Table~\ref{tab:l3-fine-taxonomy} summarizes the four feedback sources.

\begin{table*}[!t]
\caption{Categories and Subcategories of L3 Outcome-Adaptive Control}
\label{tab:l3-fine-taxonomy}
\centering
\small
\setlength{\tabcolsep}{2pt}
\renewcommand{\arraystretch}{1.28}
\begin{tabular}{T{0.11\textwidth}T{0.18\textwidth}T{0.22\textwidth}T{0.44\textwidth}}
\toprule
\textbf{Category} & \textbf{Subcategory} & \textbf{Controlled decision} & \textbf{Representative systems} \\
\midrule
\rowcolor{gray!15}\cellcolor{white}
& Image and structured-content renders & Revise from visible semantic, compositional, or layout defects & RPG~\cite{rpg}, GenArtist~\cite{genartist}, PaperBanana~\cite{src260123265} \\
& Video renders & Revise motion, composition, or camera from generated frames & GenMAC~\cite{a241204440}, MotionAgent~\cite{a250203207}, Co-Director~\cite{a260424842} \\
\rowcolor{gray!15}\cellcolor{white}
& 3D and multiview renders & Revise geometry or placement from rendered views & SceneAssistant~\cite{src260312238}, VIGA~\cite{src260111109}, WorldClaw~\cite{src260805248} \\
\multirow{-4}{=}{\textbf{Perceptual\allowbreak outcome}} & Document and interface renders & Revise document or code from rendered appearance & PPTAgent~\cite{pptagent}, UI2Code$^{\mathrm{N}}$~\cite{ui2coden}, VisRefiner~\cite{visrefiner} \\
\midrule
& Program and workflow state & Revise from execution results or structured content state & ComfySearch~\cite{src260104060}, GVR-Coder~\cite{src260728073} \\
\rowcolor{gray!15}\cellcolor{white}
& Timeline and source state & Revise later edits from retained temporal state & ReCA~\cite{a260526525}, LAVE~\cite{lave}, Crayotter~\cite{src260802694} \\
\multirow{-3}{=}{\textbf{Structured/\allowbreak execution}} & Scene and engine state & Revise from scene, program, or runtime observations & Authoring for Living Worlds~\cite{src260410383}, HDSL~\cite{src260609738}, MUSE~\cite{src260614168} \\
\midrule
& Simulated dynamics & Revise motion or behavior from simulation results & NEWTON~\cite{newton}, MoReGen~\cite{a251204221}, VideoCoCo~\cite{a260727380} \\
\rowcolor{gray!15}\cellcolor{white}
\multirow{-2}{=}{\textbf{Physical/\allowbreak constraint}} & Geometric and world constraints & Revise from collision, stability, reachability, or rule checks & PhyScensis~\cite{src260214968}, World Craft~\cite{src260109150}, MAGIC~\cite{src260711594} \\
\midrule
\textbf{Human review} & In-episode review and approval & Revise, continue, or stop from explicit human judgment & Promptify~\cite{a230409337}, CoGen3D~\cite{src260703731}, OrchestrXR~\cite{src260701588} \\
\bottomrule
\end{tabular}
\end{table*}

\subsection{Perceptual Outcome Feedback}\label{sec:l3-a}
Perceptual feedback is a rendered image, video segment, multiview projection, slide, or interface screenshot whose visible discrepancy changes a later action. The controller may revise a condition, search another candidate, choose a local edit, reroute a tool, or stop. Those responses do not create separate feedback categories.

\textbf{Image and structured-content renders.} RPG and SLD localize compositional failures and revise regional conditions or latent operations~\cite{rpg,self_correcting}. RS-Gen and GenPilot connect rendered diagnosis to regeneration~\cite{src260623221,genpilot}, while CountLoop, MetaPoint, Atelier, FiRe, VisualPrompter, narrative-product critics, and IA-T2I target counting, point placement, style, or reference-use failures~\cite{a250816644,a260605031,src260806751,a260413491,a250623138,src260416958,a250414868,src250515779}. Search-based controllers allocate additional prompt, latent, or operation exploration from the current output~\cite{a240317804,a250616853,a250922761,a251111483,a260517969,a260118543,a250722076,a250912446,a250601370,a260206166,a250209411,src250810494,a240212741,a260312829,a250910704,src250420054,src260318627,src260300483,src260420730,src260724353,src260701883,src260811635,compagent,src260524453}.

GenArtist and T2I-Copilot map visible defects to tools or localized edits~\cite{genartist,t2i_copilot}; related repair systems specialize regeneration, retouching, task decomposition, and stopping~\cite{a260507457,a260425636,src250405306,src260530611,whattoeditnext,a260102046,a260222809,src250705259,src250817435,src251121087,src260217558,src260316967,src260403156,src260415917,src260515181,src260608016,src260619073,a251111780,src260806075}. State-preserving editors retain accepted canvas elements while interpreting the next visual diagnosis~\cite{a260209084,src250806916,src260329602,a260312155,src260723588,src260705465,src260708497,src260700920,src260723920,src260800548,src260327817,src260701102,src260104390,src260802218,src260523527}. Unified and latent policies implement the same render-to-action relation inside one learned controller~\cite{image_cot,a260212279,toolartist,a260626907,a260202437,a251205112,a250805606,src260306032,src260302681,a260328088,a260514709,a260504040,src250906945,src260613679,a260404746,a260516961,a250301298}. Scientific-figure and executable-workflow systems apply rendered diagnosis to diagrams, visual explanations, or workflow outputs~\cite{src260123265,src260203828,src260312597,src260709839,src250317671,src250517908,src260104060,src260715845}.

\textbf{Video renders.} GenMAC~\cite{a241204440} routes a failed compositional requirement to redesign, while MotionAgent~\cite{a250203207} changes trajectory and camera controls after inspecting generated motion. AniME, Hollywood Town, CoAgent, and MAViS apply perceptual correction at different temporal scales~\cite{a250818781,a251022431,a251222536,a250808487}. Automated reviewer systems similarly convert generated frames or trajectories into later generation, story, or camera actions~\cite{a260424842,src260522144,a240809787,a251015831,src260517423,a260416541,src260220664,src260312310,src260405489,src260516748,src260618591,src260616184,src260613861,src260719038,src260716355,src260812290}. A reviewer role belongs here when it reads the rendered outcome; its architectural name does not create a distinct feedback source.

\textbf{3D and multiview renders.} SceneAssistant~\cite{src260312238} revises assets and layouts from scene views, while VIGA~\cite{src260111109} interleaves code generation, rendering, inspection, and repair. SceneConductor, Scenethesis, WorldClaw, and Kubrick correct geometry, placement, lighting, or scripts from image evidence~\cite{src260608402,scenethesis,src260805248,a240810453}. WorldAgents, IterCAD, and ParticleGen add multiview verification, visually grounded CAD correction, or rendered particle-effect feedback~\cite{src260319708,src260613368,src260800629}. Their shared limitation is observability because one attractive view cannot establish validity behind the camera or under interaction.

\textbf{Document and interface renders.} PPTAgent~\cite{pptagent} uses intermediate slide renders to guide incremental editing, while PaperX, APEX, and DeepPresenter revise slide or poster content from visual diagnosis~\cite{src260203866,src260104794,src260222839}. UI2Code$^{\mathrm{N}}$, DesignCoder, VisRefiner, CITL, GameUIAgent, and AceCoder map browser screenshots to later code changes~\cite{ui2coden,designcoder,visrefiner,visionguided,src260314724,frontalk}. These systems are classified by rendered appearance even when the repair itself operates on structured objects.

\subsection{Structured and Execution Feedback}\label{sec:l3-b}
Structured feedback exposes program, document, timeline, scene, or runtime state. The decisive observation is an execution result or state transition rather than appearance alone.

\textbf{Program and workflow state.} COMIC, LayerCraft, I2E, MCTS-Report, GVR-Coder, and Any2Poster use critic state, layers, object representations, or executable validation to select a later repair~\cite{src260311048,a250400010,src260103741,src260804071,src260728073,src260602915}. Talk to Your Slides uses execution errors to trigger another structured edit~\cite{talkslides}. OmniPresent, Auto-Slides, PreGenie, Automatic Slide Updating, PPTArena, SeaSlides, Learning to Present, ETPDesigner, and WebVIA likewise connect document or browser state to later operations~\cite{src260702590,autoslides,src250521660,slideagent,pptarena,src260803298,src260316839,src260719947,webvia}. Graph search, code repair, and localized editing are strategies for acting on this state.

\textbf{Timeline and source state.} ReCA~\cite{a260526525} extracts visual, narrative, and transition state after each multi-shot unit. LASEV and Mind-of-Director use compiled or explicit temporal state to alter later operations~\cite{a260211790,src260314790}. LAVE grounds requests in source clips and timeline operations, while Crayotter optimizes traceable long-horizon edit trajectories~\cite{lave,src260802694,a260607636}. T2VTree, DIRECT, GLANCE, EditDuet, CutClaw, and AutoMV similarly revise source selection, ordering, synchronization, or transitions from evolving edit state~\cite{src260208368,src260404875,src260405076,a250910761,a260329664,a251212196}.

\textbf{Scene and engine state.} Authoring for Living Worlds~\cite{src260410383} checks each event against accumulated simulator state, HDSL~\cite{src260609738} verifies and repairs a hierarchical scene program, and MUSE~\cite{src260614168} updates requirement and preservation state after each edit. SimWorlds and MANSION use runtime or queried scene state to advance, retry, or provision another object~\cite{src260701766,src260311554}. Agentic 3D Scene Generation and PlanCraft retain spatial contracts~\cite{agentic3d,src260723491}, while Articraft, SceneCode, FilmAgent, Cutscene Agent, StateFlow, Global-Local Monte Carlo Tree Search, and Lumera feed tests, outcomes, scripts, or world state to the controller~\cite{src260515187,src260519587,a250112909,src260425318,src260812314,src260606002,src260720889}.

\subsection{Physical and Constraint Feedback}\label{sec:l3-c}
Physical feedback is distinguished from general execution state because a simulation, geometric relation, reachability test, or hard rule supplies the accept or repair condition. A renderer or engine may implement the check, but the violated constraint must change the next action.

\textbf{Simulated dynamics.} NEWTON~\cite{newton} combines scientific computation with verifier-triggered replanning for dynamics-aware video. Environment feedback exposes collision, occlusion, or rule failures during generation~\cite{a241010076}. MoReGen, VideoCoCo, PhysCodeBench, and GS-Agent close related loops around simulated motion, executable physical behavior, or physics-engine state~\cite{a251204221,a260727380,src260423580,src260721522}. These methods differ from perceptual correction because plausible appearance does not establish correct dynamics.

\textbf{Geometric and world constraints.} PhyScensis uses physics-solver feedback, and SAGE gates simulation-ready scenes with physical-stability checks~\cite{src260214968,src260210116}. Agentic Designer and SceneSmith correct collision, support, and stability failures during layout construction~\cite{src260720866,src260209153}. iARCS adapts scene generation from constraint rewards~\cite{src260806161}, while World Craft and MAGIC use collision, connectivity, navigability, or transition checks to repair world structure~\cite{src260109150,src260711594}.

\subsection{Human Review Feedback}\label{sec:l3-d}
Human review is a separate feedback source only when explicit judgment inside the episode changes a later generation action. A terminal preference study remains evaluation.

\textbf{In-episode review and approval.} Promptify~\cite{a230409337} converts preferences over generated images into changes to the current prompt. CoGen3D lets a user revise or approve a concept image before image-to-3D generation, while OrchestrXR supports review and patching across study design, scene construction, and interaction logic~\cite{src260703731,src260701588}. From Idea to Co-Creation~\cite{src260105016} also lets a human supervisor redirect critic-guided revision. The criterion is causal intervention by a person, not the presence of a role named reviewer.

\subsection{Discussion: Diagnosis, Preservation, and Stopping}\label{sec:l3-e}
L3 improves tasks by spending computation conditionally, but every additional action creates another failure opportunity. This subsection compares the properties that cut across prompt revision, tool-based editing, video control, and executable visual outputs: how diagnosis is connected to action, how accepted state is preserved, which feedback sources expose different failures, and where within-task adaptation ends.

\textbf{Diagnosis-action coupling.} The strongest distinction within L3 is the coupling between diagnosis and action. Prompt refinement systems translate a visual defect back into language. Tool-based editors can instead choose a localized operation, while executable-content systems map the defect to code, XML, or scene state. The latter two provide a clearer path for credit assignment because the affected object remains identifiable. They also require more precise critics. A mistaken global prompt may degrade the whole sample, but an incorrect code edit can silently break a previously valid component.

\textbf{Preservation and stopping.} Once a diagnosis has selected an action, preservation and stopping become as important as correction. Agent Banana records accepted image content, ReCA maintains visual and narrative state across video segments, and browser-based systems test whether earlier interface requirements regress. These mechanisms address the same problem at different scales: a successful local repair is not progress if it destroys more valuable state elsewhere. Studies should report non-target change and regression after every accepted action, not only the quality of the final output.

\textbf{Feedback sources.} Whether that action is reliable depends in turn on which feedback source produced the diagnosis. Multimodal critics can detect semantic mismatch and some aesthetic defects, executable validators can expose compilation or structural errors, and simulators can test geometry or physics. No one source covers all three. NEWTON combines scientific computation with visual verification because appearance alone cannot establish physical correctness. PPTAgent and UI2Code$^{\mathrm{N}}$ combine structured execution with rendered evidence because valid files can still look wrong. A robust L3 controller should retain these signals separately and select repairs according to the violated requirement rather than collapse them prematurely into one scalar score.

\textbf{Episode boundary.} Finally, even reliable within-task feedback does not provide cross-task adaptation. GenPilot can remember failed actions during one request, and a long-video system can preserve state for hundreds of steps, yet neither history helps a later independent goal if it is discarded when the task ends. The next section examines systems that retain and reuse such experience.

\section{L4: Experience-Adaptive Control}
\label{sec:l4}

L4 carries an update across the episode boundary. After episode $\mathcal{E}_n$, retained state changes as $m_{n+1}=U(m_n,\mathcal{E}_n)$ and influences later goals. We classify systems by the carrier in which experience persists: capability and tool profiles, episodic or user memory, reusable procedures and skills, executable workflows and harnesses, or policy and model updates. Image, video, document, interface, and 3D applications are examples within these carrier types rather than peer categories. Figure~\ref{fig:l3-l4-adaptation-flow} separates this persistent update from an L3 repair loop, and Table~\ref{tab:l4-fine-taxonomy} summarizes the five carriers.

\begin{table*}[!t]
\caption{Categories and Subcategories of L4 Experience-Adaptive Control}
\label{tab:l4-fine-taxonomy}
\centering
\small
\setlength{\tabcolsep}{2pt}
\renewcommand{\arraystretch}{1.28}
\begin{tabular}{T{0.11\textwidth}T{0.18\textwidth}T{0.22\textwidth}T{0.44\textwidth}}
\toprule
\textbf{Category} & \textbf{Subcategory} & \textbf{Controlled decision} & \textbf{Representative systems} \\
\midrule
\rowcolor{gray!15}\cellcolor{white}
\textbf{Capability and tool profiles} & Empirical capability records & Update future model or tool routing & OctoT2I~\cite{octot2i}, DiffusionAgent~\cite{diffusionagent}, PerfGuard~\cite{src260122571}, GenRouter~\cite{src260816721} \\
\midrule
\textbf{Episodic and user memory} & Retrieved episodes and preferences & Condition later control on prior outcomes or users & MemoGen~\cite{a260603243}, BrandFusion~\cite{a260302816}, UniVA~\cite{a251108521}, MemSlides~\cite{src260617162} \\
\midrule
\rowcolor{gray!15}\cellcolor{white}
\textbf{Procedures and skills} & Abstracted action procedures & Reuse a distilled strategy on a later task & GenEvolve~\cite{genevolve}, EvoDiagram~\cite{src260409568}, SceneCraft~\cite{scenecraft} \\
\midrule
\textbf{Workflows and harnesses} & Executable workflow revisions & Reuse a verified graph, program, or harness & COMFYCLAW~\cite{src260701709}, VideoWeaver~\cite{a260608091}, AutoDesign~\cite{src260813560} \\
\midrule
\rowcolor{gray!15}\cellcolor{white}
\multirow{2}{=}{\textbf{Policy and\allowbreak model\allowbreak updates}} & Persistent behavioral updates & Change future control through learned parameters or policy & SIDiffAgent~\cite{a260202051}, JarvisEvo~\cite{src251123002}, SPIRAL~\cite{spiral} \\
& Recursive self-improvement & Change the improvement mechanism through accepted self-updates & G\"odel Agent~\cite{godelagent}, Darwin G\"odel Machine~\cite{darwingodel} \\
\bottomrule
\end{tabular}
\end{table*}

\subsection{Capability and Tool Profiles}\label{sec:l4-a}
Capability profiles retain empirical evidence about which generator, editor, or tool should serve a later request. Their persistent carrier is a routing record rather than a prior episode, executable procedure, or parameter update.

\textbf{Empirical capability records.} DiffusionAgent~\cite{diffusionagent} stores expert descriptions that guide later diffusion-model selection. OctoT2I~\cite{octot2i} makes those profiles empirical through a propose, solve, evaluate, and learn loop that updates future routing from measured quality and efficiency. PerfGuard~\cite{src260122571} updates tool preferences from ranked execution evidence, while GenRouter~\cite{src260816721} changes later workflow routing from accumulated outcomes. These profiles should be evaluated for calibration when tools, prices, or request distributions change.

\subsection{Episodic and User Memory}\label{sec:l4-b}
Episodic memory retrieves records of prior requests, outcomes, or user preferences as context for a later independent task. Unlike a capability profile, the stored unit describes experience with a goal or user rather than a general tool estimate.

\textbf{Retrieved episodes and preferences.} MemoGen~\cite{a260603243} stores relation-level successes and failures and retrieves relevant episodes for a new request. BrandFusion~\cite{a260302816} collects user feedback from completed branded videos and reuses positive or negative integration strategies. UniVA~\cite{a251108521} retains trajectory and user-preference memory for later video workflows, while MemSlides~\cite{src260617162} stores hierarchical slide-generation experience for later decks. Action Agent~\cite{src260501477} links retained navigation experience with subsequent video-generation decisions. These systems qualify only when retrieved experience changes control on a later task.

\subsection{Reusable Procedures and Skills}\label{sec:l4-c}
Procedural memory abstracts an action pattern from completed trajectories. The retained unit is model-readable operational knowledge that must be interpreted before execution.

\textbf{Abstracted action procedures.} GenEvolve~\cite{genevolve} distills successful and failed image-generation trajectories into procedures for new open-ended goals. EvoDiagram and ManimAgent accumulate reusable diagram or visual-explanation skills~\cite{src260409568,src260630296}, while SEAR~\cite{src260628971} retains plans for later degradations. SceneCraft~\cite{scenecraft} stores Blender programs and spatial procedures, and SimWorld Studio~\cite{src260509423} adds reusable engine tools and skills after verifier-guided world construction. These systems reuse an abstracted strategy rather than replay one completed output.

\subsection{Executable Workflows and Harnesses}\label{sec:l4-d}
Executable persistence stores a graph, program, middleware component, or control harness that can organize a later workflow. This carrier can be validated and versioned because its execution semantics exceed those of a declarative skill.

\textbf{Executable workflow revisions.} COMFYCLAW~\cite{src260701709} evolves typed ComfyUI graphs and promotes a workflow only after staged verification. FigAgent~\cite{src260329590} evolves reusable drawing middleware, while AutoDesign~\cite{src260813560} updates the harness that coordinates later design trajectories. VideoWeaver~\cite{a260608091} evaluates and revises long-video workflow skills before merging them across task categories. AVA-Encoder~\cite{avaencoder} evolves a program for representing film-level state across later video tasks. In each case the retained executable object, rather than the application modality, defines the category.

\subsection{Policy and Model Updates}\label{sec:l4-e}
The final carrier writes completed experience into a learned policy or model so later control changes without retrieving a separate memory item or workflow.

\textbf{Persistent behavioral updates.} SIDiffAgent~\cite{a260202051} changes later generation behavior from completed diffusion trajectories, while JarvisEvo~\cite{src251123002} co-evolves editing and evaluation components. SPIRAL~\cite{spiral} uses critic trajectories to improve behavior on later action-conditioned video goals. CLARE~\cite{src260716352} updates its clarification policy from completed 3D interactions, and SymbOmni~\cite{src260712042} incorporates accumulated symbolic concepts into later model behavior. Parameter and policy updates can generalize beyond explicit retrieval, but they are harder to inspect and roll back.

\textbf{Recursive self-improvement.} A stronger L4 case is recursive self-improvement (RSI), in which a retained update changes not only task behavior but also the mechanism that proposes or evaluates later updates. G\"odel Agent makes agent logic editable under a high-level objective, whereas the Darwin G\"odel Machine iteratively modifies agent code and accepts variants through empirical benchmarks~\cite{godelagent,darwingodel}. These systems address general-purpose agents rather than visual generation, so they are architectural precedents rather than members of the visual-generation corpus. A visual-generation controller would qualify as RSI only when an accepted change to its controller or improvement policy measurably improves its ability to generate and validate still later changes on independent visual tasks. Repeated prompt revision, self-reflection, or training within a fixed improvement loop does not meet this condition because the improvement mechanism itself remains unchanged. RSI is therefore not an additional level in our hierarchy. It is the most self-referential form of L4 because its causal reach still crosses episode boundaries.

\subsection{Discussion: Transfer, Failure, and Rollback}\label{sec:l4-f}
Experience can preserve mistakes as easily as useful behavior. L4 evaluation must use chronological held-out tasks and report forward transfer, negative transfer, stale-experience sensitivity, provenance, and rollback. A self-evolution label without cross-task behavioral evidence is insufficient.

RSI imposes a stricter evaluation requirement. Later generations must become better at producing validated updates, rather than only improving task performance under one fixed update procedure. Evaluation should therefore report task-level gains separately from gains in the improvement process and should use external validation to detect self-confirming regressions.

The five carriers differ in reversibility. A profile entry or episode can be deleted, a procedure can be revised, and an executable workflow can be versioned. A harmful parameter update may instead require a checkpoint or a new training run. Comparing these carriers requires measuring both later-task benefit and the cost of detecting and undoing a bad update.

\section{Training and Reinforcement Learning for Generation Controllers}
\label{sec:training}

The capability taxonomy describes what a controller can change at inference time, whereas training determines how these capabilities are learned. This distinction matters because neither supervised fine-tuning nor reinforcement learning inherently determines the controller level. A fixed generator optimized with an image reward may remain in L0 Fixed Support, while a supervised router can make L2 Execution Control decisions and an untrained repair loop can make L3 Outcome-Adaptive Control decisions. We organize controller learning into four causal stages. Trajectory data record decisions and their consequences, supervised fine-tuning initializes behavior from demonstrations, feedback provides credit signals, and policy optimization improves the resulting policy.

\subsection{Data and Trajectory Supervision}
Generation controllers act through sequences rather than isolated condition-output pairs. Thus, their training data must connect a request and controller state to an action, the resulting observation, resource use, and termination. We distinguish nominal action trajectories, failure and repair trajectories, and records that preserve long-horizon or cross-task context. This subsection concerns record content and construction, not how model parameters are updated from it.

\textbf{Nominal action trajectories.} A nominal controller trajectory records the ordered states, decisions, actions, and observations produced during successful generation. Useful datasets preserve plans, tool calls, intermediate results, and termination rather than retaining only final synthesized outputs. Visual ChatGPT~\cite{visual_chatgpt} and T2I-Copilot~\cite{t2i_copilot} make tool and role decisions observable, while GenArtist~\cite{genartist} adds decomposition and verification. Temporal alignment is essential because training must recover which observation followed each action rather than merely imitate the final appearance distribution.

\textbf{Failure and repair trajectories.} Successful demonstrations provide weak evidence about when to revise, stop, or abandon a route. Failure and repair data retain the state before an error, the responsible action, the observed consequence, and a feasible alternative. Image-POSER~\cite{a251111780} separates routing, editing, and stopping decisions so that poor operation selection is not confused with poor execution. Crayotter~\cite{src260802694} applies the same principle to long video-editing sequences, where a missing event may originate in retrieval, ranking, or timeline assembly. Matched counterfactuals and rejected search branches provide clearer credit than preference pairs in which several operations change simultaneously.

\textbf{Long-horizon and cross-task records.} Beyond local success and failure pairs, temporal and persistent controllers require data that preserve state across longer horizons and task boundaries. SPIRAL~\cite{spiral} aligns plans and critic feedback with action-conditioned video segments, while VideoWeaver~\cite{a260608091} records complete workflow experiences for subsequent skill revision. GenEvolve~\cite{genevolve} connects completed tool orchestration with reusable procedures, and OctoT2I~\cite{octot2i} accumulates evidence about generator capabilities and costs. Such records should also preserve provenance, tool versions, timestamps, and rollback information. Without this context, a route learned under an earlier generator or evaluator may be reused after the assumptions underlying that route have become invalid.

\subsection{Supervised Fine-Tuning}
Supervised fine-tuning (SFT) updates a generator or controller to imitate labeled outputs and action trajectories. Unlike data construction, SFT specifies a parameter-learning step. Unlike reward-based optimization, it learns from demonstrated targets rather than assigning scalar credit to sampled alternatives. We distinguish SFT of a generator or condition policy from SFT of an executable controller trajectory.

\textbf{Generator and condition SFT.} Supervised targets can initialize behavior within a predetermined generation interface. Promptist~\cite{a221209611} learns model-facing prompt rewrites from manually engineered examples before reward optimization. ReasonGen-R1~\cite{src250524875} fine-tunes an autoregressive image generator on explicit reasoning traces before GRPO, while FrontCoder~\cite{frontcoder} uses supervised fine-tuning as one stage of frontend generation training. These methods learn useful conditions or generator behavior, but SFT alone does not broaden the deployed action space.

\textbf{Controller trajectory SFT.} Whereas generator and condition SFT targets a predetermined interface, demonstrations with interleaved reasoning, operations, and observations can initialize an executable controller. GenAgent~\cite{a260118543} learns tool invocation and reflection from multimodal trajectories before agentic reinforcement learning. ToolArtist~\cite{toolartist} converts teacher trajectories with search and image generation into a unified multimodal format, then uses them to train reasoning, tool use, and native generation within one policy. SFT therefore provides a stable starting policy for multi-step control, while later reward optimization determines which sampled trajectories should be preferred.

\subsection{Reward Models and Feedback}
Trajectory data specify what occurred, but they do not determine which decisions deserve credit. We separate terminal outcome and preference rewards from process and verifiable rewards. Content-specific checks instantiate these two reward locations rather than forming a third peer category.

\textbf{Terminal outcome and preference rewards.} Terminal rewards judge the completed output and are most useful when intermediate decisions are difficult to label. ImageReward~\cite{imagereward} and Pick-a-Pic~\cite{pickapic} learn broad human preferences over generated images, while PIGReward~\cite{a251119458} adapts evaluation dimensions to an individual user. Such rewards capture holistic appeal and intent better than one mechanical metric, but they cannot identify which tool call or revision caused the improvement. They should therefore supervise terminal success while process feedback handles credit within the trajectory.

\textbf{Process and verifiable rewards.} Because terminal rewards cannot identify which decision caused success, process feedback evaluates whether intermediate actions are valid and useful. It can test tool execution, constraint satisfaction, diagnosis-edit consistency, preservation of accepted content, and whether another action is worth its cost. GUV~\cite{a251013804} supplies localized multimodal judgments, while GenEval~\cite{geneval} and T2I-CompBench~\cite{t2i_compbench} expose entity and relation failures that can be converted into targeted signals. AlphaGRPO~\cite{a260512495} decomposes complex requests into verifiable components for denser reinforcement learning, although independent component success does not guarantee a coherent result. A practical objective therefore combines goal and process evidence with action cost,
\begin{equation}
r_t = \alpha r_t^{\mathrm{goal}} + \beta r_t^{\mathrm{process}} - \lambda c(a_t),
\end{equation}
where $r_t^{\mathrm{goal}}$ measures progress toward the requested output, $r_t^{\mathrm{process}}$ evaluates the validity and usefulness of the intermediate decision, and $c(a_t)$ is the resource cost of action $a_t$. The nonnegative coefficients $\alpha$ and $\beta$ weight goal-level and process-level evidence, while $\lambda$ controls the penalty on computation, tool use, or interaction cost. The components should be reported separately so that an apparent gain cannot be attributed only to the chosen weighting.

The evidence used for process verification depends on the content type. Image editing must protect accepted regions~\cite{a260209084,a260507457}, and video requires action-completion and temporal-state checks~\cite{spiral,a241204440}. Slides and interfaces add compilation, editability, rendering, and interaction~\cite{src260803298,ui2coden}, while 3D scenes require geometry and simulation evidence that may be hidden from one view~\cite{src260312238}. Across modalities, soft appearance rewards should operate only among synthesized outputs that satisfy task-specific hard constraints.

\subsection{Reinforcement Learning and Policy Optimization}
Once trajectories and feedback are defined, optimization determines which part of the generation process changes. The state can contain the request, current output, action history, verifier outputs, persistent experience, and remaining budget. The action can modify a generator, construct a condition, select a tool, revise an observed output, update retained state, or terminate. A cost-aware controller can be written as
\begin{equation}
J(\theta)=\mathbb{E}_{\tau\sim\pi_\theta}\left[\sum_{t=0}^{T}\gamma^t r_t\right],
\end{equation}
where $\theta$ denotes the policy parameters, $\pi_\theta$ is the controller policy, and $\tau$ is a trajectory induced by that policy and the task environment. The horizon $T$ is the final decision step, $t$ indexes decisions within the trajectory, $\gamma\in[0,1]$ is the discount factor, and $r_t$ is the cost-aware reward defined in (4). The meaning of this objective depends on which policy is optimized and when interaction is available.

The following discussion answers three different questions and does not present four mutually exclusive algorithm families. The first two branches identify the optimization target, the third identifies when interaction is available, and the fourth identifies whether the update survives the current task.

\textbf{Optimization target: generators and conditions.} The narrowest target changes visual content generation within a prescribed interface. DPOK~\cite{dpok} fine-tunes diffusion parameters against image rewards, while AeSlides~\cite{aeslides} and FrontCoder~\cite{frontcoder} use verifiable feedback for slide and interface generation. Condition policies instead treat language or structure as the action. Promptist~\cite{a221209611} and DiffChat~\cite{a240304997} optimize model-facing instructions against semantic and preference signals, and PASTA~\cite{a241210419} adapts this decision across several user turns. These methods can improve visual quality without learning a broader workflow because the executor and invocation pattern remain fixed.

\textbf{Optimization target: workflow controllers.} Broader policies choose among generation operations and can condition later decisions on visual state. Policy Optimized T2I Pipeline Design~\cite{a250521478} learns combinations of generators and processing blocks, while Image-POSER~\cite{a251111780} and GenAgent~\cite{a260118543} add routing, editing, reflection, and stopping. SearchGen~\cite{searchgen} jointly recalibrates retrieval and generation. Image CoT~\cite{image_cot}, UniGen~\cite{a250514682}, and LLM-I~\cite{a250913642} place several controller functions in one model or policy, with further realizations spanning latent, browser, and interface actions~\cite{a250922761,a260516961,ui2coden,visrefiner}. Architecture does not determine the level; the learned policy must govern a consequential generation action.

\textbf{Interaction regime.} After specifying what the policy controls, learning differs by when new interaction is available. Offline methods train from fixed data and inherit its coverage and tool boundaries. Promptist~\cite{a221209611} combines supervised initialization with reward optimization, while TIPO~\cite{a241108127} learns low-cost prompt expansion from text distributions. Online reinforcement learning can discover routes through controlled interaction, as NEWTON~\cite{newton} demonstrates inside a video generation and verification loop. Test-time methods adapt only to the current output: MILR~\cite{a250922761} optimizes multimodal representations, and Generation Navigator~\cite{a260517969} chooses among stopping, refinement, and regeneration. One policy can combine several regimes, so they are reported as training conditions rather than exclusive method categories.

\textbf{Persistence horizon.} Some updates end with the current request, while others alter later independent-task control. OctoT2I~\cite{octot2i} updates a capability profile for future routing. GenEvolve~\cite{genevolve} distills completed experience into procedures, COMFYCLAW~\cite{src260701709} promotes verified workflows, and SIDiffAgent~\cite{a260202051} retains successful and failed diffusion experience. VideoWeaver~\cite{a260608091}, VISTA~\cite{a251015831}, and SPIRAL~\cite{spiral} extend persistent adaptation to video behavior. These carriers differ in reversibility, but all require measurable change on a later independent task. Held-out transfer, negative transfer, stale experience, and rollback are evaluated in Section~\ref{sec:evaluation}.

The four stages expose different failure points. Incomplete trajectories hide the action that caused an outcome, supervised fine-tuning can imitate narrow or inconsistent demonstrations, weak rewards assign credit to the wrong behavior, and unconstrained optimization exploits whichever signal is easiest to increase. Reporting all four stages is therefore necessary to distinguish an improved controller from a stronger generator, a larger search budget, or a policy that overfits its evaluator.

\section{Evaluation and Benchmarking}
\label{sec:evaluation}

\begin{figure*}[!t]
\centering
\includegraphics[width=\textwidth]{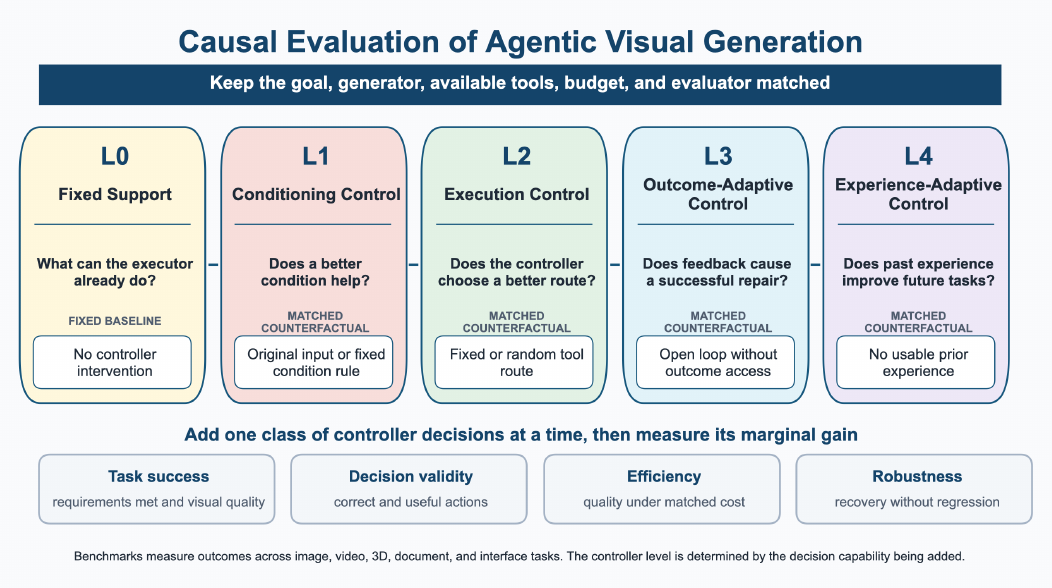}
\caption{Causal evaluation of agentic visual generation. The generator, available tools, budget, and evaluator remain matched while one class of controller decisions is added at each level. The comparison attributes marginal improvement to specification construction in L1 Conditioning Control, operation selection in L2 Execution Control, outcome-dependent repair in L3 Outcome-Adaptive Control, or cross-task experience reuse in L4 Experience-Adaptive Control. Content-specific benchmarks provide measurement signals but do not determine the controller level.}
\label{fig:evaluation-causal-framework}
\end{figure*}

Training specifies how a controller changes, but evaluation must determine whether better decisions caused the improvement. A stronger generator, a larger sampling budget, or an evaluator shared with the controller can otherwise create the appearance of agentic progress. Our second major contribution is a \emph{level-conditioned evaluation framework that isolates the causal value of broader controller decision-making scope by matching generators, tools, budgets, and evaluators across L1--L4}. L0 establishes the fixed-executor baseline. Each later level then changes one class of decisions and compares it with a matched counterfactual that removes that capability. Figure~\ref{fig:evaluation-causal-framework} presents this logic, and Table~\ref{tab:level-evaluation} specifies the corresponding controls and evidence.

\begin{table*}[!t]
\caption{Level-Conditioned Evaluation Protocol}
\label{tab:level-evaluation}
\centering
\small
\setlength{\tabcolsep}{2pt}
\renewcommand{\arraystretch}{1.28}
\begin{tabular}{T{0.24\textwidth}T{0.21\textwidth}T{0.21\textwidth}T{0.27\textwidth}}
\toprule
\textbf{Level-specific question} & \textbf{Matched factors} & \textbf{Counterfactual} & \textbf{Evidence of added value} \\
\midrule
\rowcolor{gray!8}L0 Fixed Support: What can the fixed executor already do? & Input condition, random seeds, sampling parameters, candidate budget & Fixed executor without controller intervention & Requirement success, synthesized-content quality, human preference, cost \\
L1 Conditioning Control: Does specification control improve the same executor? & Generator, seeds, sampling parameters, candidate budget & Original input or fixed specification rule & Intent preservation, specification validity, generator compliance \\
\rowcolor{gray!8}L2 Execution Control: Does the controller select a better route? & Available models and tools, action budget, evaluator & Fixed, random, frequency-based, or oracle route & Routing regret, invalid actions, success and quality under matched cost \\
L3 Outcome-Adaptive Control: Does outcome feedback cause a successful repair? & Initial plan, tools, generator, evaluator, total budget & Open-loop execution without intermediate outcome access & Diagnosis precision, repair success, preservation, calibrated stopping \\
\rowcolor{gray!8}L4 Experience-Adaptive Control: Does completed experience improve future tasks? & Future-task distribution, tools, generator, budget & Removed, shuffled, stale, or irrelevant experience & Forward transfer, negative transfer, forgetting, retrieval and rollback success \\
Cross-level: What is the marginal value of added scope? & Generator, evaluator, goal set, calls and monetary cost & Add one decision class at a time from L1 to L4 & Quality--cost frontier and marginal gain at each transition \\
\bottomrule
\end{tabular}
\end{table*}

\subsection{L0 Fixed Support: What Can the Fixed Executor Already Do?}
L0 Fixed Support supplies the baseline against which every claim of controller benefit must be measured. We first test synthesized-content validity and task completion, then quality and human preference, and finally how these signals should be reported without hiding hard failures through aggregation. The executor, input condition, seeds, sampling parameters, and candidate budget remain fixed throughout this baseline.

\textbf{Synthesized-content validity and task completion.} Evaluation should first test requirements that cannot be compensated for by visual appeal. These include requested entities and relations, exact text or counts, preservation of content during editing, file validity, executability, and action completion. The concrete checks depend on the synthesized content, but they serve the same role: an image that omits an object, a video that misses an action, a scene that violates physical state, and an interface that fails to run are unsuccessful even when selected views appear convincing. Such requirements should be reported as separate pass rates rather than folded into a quality average.

\textbf{Quality and human preference.} Among synthesized outputs that satisfy their hard requirements, automatic evaluators can measure fidelity, alignment, perceptual quality, and consistency. ImageReward~\cite{imagereward} and Pick-a-Pic~\cite{pickapic} add learned evidence of overall human preference, while direct human evaluation remains necessary for creative intent and usability. Studies should disclose the evaluator population, instructions, comparison count, randomization, agreement, and confidence intervals. Preference and absolute quality scores remain complementary to requirement-specific checks because an attractive output can still violate the task.

\textbf{Reporting and aggregation.} A useful L0 report presents a vector containing hard-constraint success, synthesized-content quality, human preference, executability, and resource cost. Aggregation into one number should be avoided unless the weighting is disclosed, and hard failures should never be averaged away by a high aesthetic score. The resulting frozen report becomes the common output baseline for the L1--L4 interventions below.

\subsection{Controls Shared by L1--L4 Evaluations}
The L0 baseline establishes whether the final output succeeds. Controller evaluation additionally asks whether the trajectory is valid, efficient, robust, and causally responsible for that success. DiffusionAgent~\cite{diffusionagent}, GenArtist~\cite{genartist}, OctoT2I~\cite{octot2i}, and GenEvolve~\cite{genevolve} expose different parts of this trajectory evidence. The four dimensions below are shared reporting requirements rather than another taxonomy, and each supports the level-specific counterfactuals that follow.

\textbf{Decision validity.} Decision validity covers constraint coverage, dependency consistency, and correspondence between planned and executed states. It also includes invalid-call rate, argument correctness, evidence relevance, redundant actions, and whether a diagnosis predicts a repair that improves the targeted region. For persistent controllers, this dimension tests whether memory is relevant rather than stale or unrelated.

\textbf{Efficiency and trajectory utility.} Efficiency records model calls, generated candidates, editing operations, tokens, latency, accelerator time, and monetary cost. Let $\tau=(s_0,a_0,o_1,\ldots,s_T)$ denote a trajectory, $Q(y_T)$ terminal quality, and $C(\tau)$ total resource cost. Comparisons should report the Pareto frontier of $Q$ and $C$ and success under fixed budgets. This prevents best-of-many sampling from being compared directly with a single-call baseline without accounting for additional search.

\textbf{Robustness and recovery.} This dimension measures whether the controller detects a known failure and returns to a valid route under the same remaining budget. The injected failure should target the state available at the evaluated level. L1 can receive an invalid layout or irrelevant reference, L2 an unavailable tool, L3 a corrupted intermediate outcome or critic error, and L4 stale or unrelated experience. Reports should identify the injection point, the expected affected decision, recovery success, and any regression in previously satisfied requirements. Human controllability can be measured through edit effort, intervention count, undo success, and persistence of accepted modifications.

\textbf{Faithful attribution.} Attribution should be evaluated through interventions rather than the fluency of a rationale. Planned entities should correspond to rendered regions, reported actions should match tool calls, and a diagnosis should predict which repair succeeds. Counterfactual tests can remove a cited reference, alter one planned constraint, or replace a retrieved skill and check whether the claimed part of controller behavior changes.

\subsection{L1 Conditioning Control: Does the Specification Improve a Fixed Generator?}
L1 evaluation isolates the value of the specification while preventing execution differences from entering the comparison. The controller should be credited only when its prompt, layout, reference, or other condition improves the same predetermined executor under the same sampling budget.

The generator, random seeds, sampling parameters, and candidate budget should be held fixed while the condition policy is varied. Promptist~\cite{a221209611} should therefore be compared with the unmodified request and with supervised prompt rewriting under identical sampling conditions. LayoutGPT~\cite{layoutgpt} requires an additional comparison against layouts produced without in-context spatial reasoning. Reports should separate intent preservation, specification validity, generator compliance, and terminal preference. If a rewritten prompt improves aesthetics but deletes a requested object, the condition policy has traded away goal coverage rather than solved the request. Cross-generator evaluation is also necessary because a specification optimized for one generator may not transfer to another.

\subsection{L2 Execution Control: Does the Controller Select a Better Route?}
L2 evaluation holds the available operations fixed and varies only who decides which operation is invoked. The relevant counterfactual is not a weaker tool set. It is a fixed, random, frequency-based, or oracle route over the same generators, editors, search tools, and renderers.

Visual ChatGPT~\cite{visual_chatgpt} gains capabilities from the models it can invoke, so a fair ablation compares its controller with a fixed routing rule over the same tool set. Mind-Brush~\cite{a260201756} should be compared with a system that retains the same search, reasoning, generation, and editing components but fixes generation-versus-editing mode in advance. The evaluation should record routing regret, invalid calls, redundant calls, parameter errors, execution failures, cost, and the fraction of goals for which the selected route outperforms the best single-operation baseline. An oracle router supplies the attainable upper bound. Random and frequency-based routers reveal how much improvement comes from the decision policy rather than raw tool strength.

\subsection{L3 Outcome-Adaptive Control: Does Feedback Cause a Successful Repair?}
L3 evaluation must establish the full outcome-to-action link. We first compare matched open and closed loops, then audit selection bias that can mimic successful repair. A higher final score after several iterations is insufficient because repeated sampling, best-of-many selection, or an evaluator shared with the controller can produce the same pattern without a correct diagnosis or repair.

\textbf{Matched open and closed loops.} L3 requires paired open-loop and closed-loop executions. The open-loop variant receives the same initial goal and budget but cannot observe intermediate renders. The closed-loop variant may diagnose and revise, while the generated candidates and evaluator scores are logged at every step. For SCOPE~\cite{scope}, one can remove persistent commitments or replace conditional skill selection with a fixed sequence to determine whether structured state is responsible for successful repair. For NEWTON~\cite{newton}, freezing the initial physical specification tests whether verifier-triggered replanning corrects dynamics beyond the first plan. The main metric should not be the quality of the selected final output alone. Studies should report diagnosis precision, repair success conditioned on a correct diagnosis, regression on previously satisfied constraints, and calibrated stopping.

\textbf{Selection bias.} Selection bias is especially severe in iterative systems. If a method generates eight candidates and reports the best one, it should not be compared with a one-sample baseline. The evaluator that selects a candidate may also be the metric used for publication, producing circular gains. A stronger protocol separates the controller critic, the stopping critic, and the held-out evaluator. Human judgments or independent task verifiers should audit cases in which the internal score rises but a hard requirement becomes false. Reward-model agreement should also be measured before and after optimization because a policy can actively search for blind spots that were absent in static evaluation data. SLD~\cite{self_correcting} is an informative test case because its value depends on whether feedback localizes a compositional failure rather than merely encouraging another sample. A complete L3 report should also measure non-target preservation, improvement per accepted revision, and recovery after an injected critic or tool failure.

\subsection{L4 Experience-Adaptive Control: Does Experience Improve Future Tasks?}
L4 evaluation crosses the episode boundary and must therefore preserve temporal order. The protocol should separate an experience-acquisition phase from a future-task evaluation phase. Future goals must remain disjoint from stored trajectories while still permitting transfer of a learned capability, preference, or procedure. Every retained item should carry provenance and a timestamp, and each later decision should identify which memory item, capability profile, or reusable skill influenced it. This design prevents duplicated tasks or leaked future information from being mistaken for adaptation.

OctoT2I~\cite{octot2i} can be evaluated through routing regret before and after capability-profile updates. GenEvolve~\cite{genevolve} requires held-out goals that can benefit from learned procedures without reproducing the stored trajectories. The decisive counterfactuals remove experience, shuffle it across unrelated tasks, replace it with stale experience, or inject an irrelevant precedent. Reports should distinguish forward transfer from negative transfer and forgetting. They should also measure retrieval accuracy, stale-memory recovery, rollback success, and the cost of maintaining the experience store.

\subsection{Cross-Level Matched Evaluation}
The four experiments above isolate one class of controller decisions at a time. Cross-level comparison preserves the budget, adds decision-making scope incrementally, and ties each gain to a logged intervention.

\textbf{Budget preservation.} Ablations must preserve the action budget. Removing reflection often makes a system cheaper, while removing memory can increase search. Comparing variants at unrestricted compute would conflate architecture with expenditure. Each variant should therefore be evaluated under both matched-call and matched-cost settings. The first reveals decision efficiency under equal opportunities; the second reflects practical resource use when tools have different prices and latencies. Quality-cost curves should include confidence intervals over goals and stochastic runs, because a controller may improve average quality by spending disproportionately on a small subset of difficult cases.

\textbf{Incremental decision-making scope.} Cross-level comparisons should test one added class of decisions at a time. Starting from an L1 condition policy, an experiment can add L2 routing, then L3 outcome feedback, and finally L4 experience reuse while preserving the same generators and evaluators. The marginal gain at each transition estimates the value of the newly enabled decision. This design is stronger than comparing unrelated named systems whose tools, backbones, and budgets differ. It also exposes the distinct error modes introduced at each transition, such as critic amplification during feedback or stale constraints during experience reuse. The taxonomy is therefore an evaluation scaffold that attributes gains and failures to the newly enabled decision.

\textbf{Logged interventions.} Finally, causal claims should be tied to logged interventions. If a controller states that an object is missing, the evaluation can force the corresponding local repair and compare it with an unrelated edit. If a retrieved reference is claimed to guide style, removing or replacing it should change the relevant visual attributes. If a stored skill is credited for a later success, executing the same goal without that skill should reduce success under the same budget. Such counterfactual tests convert plans, critiques, and memories from plausible narratives into falsifiable components of the generation process.

\subsection{Task Environments and Modality-Specific Evaluators}
Level-conditioned ablations determine what must be compared, while task benchmarks supply goals and modality-specific evaluators. We first map modalities to suitable evaluators, then introduce controlled perturbations that test recovery. A benchmark does not determine the controller level by itself. The same task can test an L1 specification policy, an L2 router, an L3 repair loop, or an L4 skill library when paired with the appropriate counterfactual.

\textbf{Modality-specific evaluators.} Table~\ref{tab:modality-evaluators} summarizes how existing benchmarks instantiate the output side of this protocol. Their scores remain supporting evidence. For example, a higher GenEval score does not reveal whether the gain came from an L1 specification, L2 routing, or L3 repair unless the corresponding control variables are matched.

\begin{table*}[!t]
\caption{Modality-Specific Evaluators Within the Level-Conditioned Protocol}
\label{tab:modality-evaluators}
\centering
\small
\setlength{\tabcolsep}{2pt}
\renewcommand{\arraystretch}{1.28}
\begin{tabular}{T{0.12\textwidth}T{0.30\textwidth}T{0.25\textwidth}T{0.27\textwidth}}
\toprule
Output type & Benchmarks / evaluators & Signals & Protocol use \\
\midrule
\rowcolor{gray!8}Image & T2I-CompBench~\cite{t2i_compbench}, GenEval~\cite{geneval}, Draw ALL Your Imagine~\cite{a250524787}, MME-Unify~\cite{a250403641}, AtelierEval~\cite{a260522645}, AgentGen-Bench~\cite{searchgen} & Entities, counts, relations, alignment & Match generator for L1; tools, feedback, or experience for L2--L4 \\
Video & VBench~\cite{vbench}, EvalCrafter~\cite{evalcrafter}, AIGVE-MACS~\cite{a250701255}, UniVA-Bench~\cite{a251108521}, ActVideoGen-Bench~\cite{spiral}, CineBench~\cite{a260410456} & Appearance, motion, temporal state, actions & Match clip budget; separate quality from repair \\
\rowcolor{gray!8}3D/world & SceneCraft~\cite{scenecraft}, Agentic 3D Scene Generation~\cite{agentic3d}, SPIRAL~\cite{spiral} & Program validity, geometry, physics, transitions & Freeze engine, assets, views, simulator \\
Slides & PPTBench~\cite{pptbench}, PresentBench~\cite{presentbench}, PPTArena~\cite{pptarena}, PPT-Eval~\cite{ppteval}, DynaSlide~\cite{slideagent}, TSBench~\cite{talkslides} & Edits, fidelity, design, editability & Freeze deck, object model, operations, render budget \\
\rowcolor{gray!8}Interfaces & Design2Code~\cite{design2code}, FronTalk~\cite{frontalk}, Vision2Web~\cite{vision2web} & Fidelity, runnable code, interaction, regression & Freeze browser, code environment, tests, revisions \\
\bottomrule
\end{tabular}
\end{table*}

\textbf{Controlled perturbations.} Benchmark suites should instantiate the shared robustness protocol with modality-specific state changes. An image benchmark may alter a region constraint, while a video benchmark may corrupt a temporal reference. A document benchmark can introduce an invalid file state, and a 3D benchmark can violate geometry or reachability. Each perturbation should expose the same underlying causal test: whether the observation changes the relevant controller decision and restores the violated requirement under a matched budget. The environment should freeze tool versions and retain generated outputs so that recovery cannot be attributed to an unreported executor change.

Together, these protocols separate synthesized content success from controller attribution. Content-specific metrics determine whether the output satisfies the task. Matched level counterfactuals determine whether the newly enabled decision caused the gain. Cost and robustness measurements determine whether that gain survives practical constraints. Modality-specific benchmarks supply the observations needed for these tests, but the intervention and its causal reach determine the evaluated controller level.

\section{Challenges and Future Directions}
\label{sec:challenges}

The preceding protocols do more than rank systems. They reveal what is missing at each increase in causal reach. We therefore organize open problems by the transition they block rather than by a disconnected list of mechanisms.

\subsection{L1 to L2: From Declarative to Executable Control}
The first transition requires a specification policy to assume responsibility for an operation. After establishing this execution problem, we examine two consequences: safety and provenance requirements, and the need to intervene selectively under a resource budget. DiffusionAgent~\cite{diffusionagent} addresses compatible pipeline construction, while GenArtist~\cite{genartist} exposes coordination challenges across generators and editors.

\textbf{Safety and provenance.} Executable control also expands the safety and provenance surface. Retrieval, model routing, 3D asset reuse, and code execution can introduce licenses, private content, or policy violations before any final output is visible. World-To-Image~\cite{world_to_image} illustrates how retrieved world knowledge enters a generated image, and SceneCraft~\cite{scenecraft} composes external assets through executable programs. Controllers should therefore attach machine-readable provenance to references, models, licenses, parameters, and edits. Permission-aware routing should reject an invalid operation before execution rather than rely only on filtering the export.

\textbf{Budgeted intervention.} The practical objective is not maximal tool use but useful intervention under a budget. A router should estimate expected quality gain, latency, monetary cost, privacy risk, and tool availability. Matched-tool evaluation is essential here because an apparent L2 gain may otherwise come entirely from access to a stronger generator.

\subsection{L2 to L3: From Execution to Reliable Feedback Control}
Once a system can act, the next transition requires observations that justify a different later action. We organize this challenge around multisource feedback, credit assignment from a diagnosis to a repair, and human feedback that must preserve accepted state. GenEval~\cite{geneval} makes image binding errors measurable, while VBench~\cite{vbench} separates several temporal properties of generated video. The difficult cases are precisely those in which a controller most needs feedback, so a fluent critique cannot be treated as reliable evidence.

\textbf{Multisource feedback.} Future feedback should combine localized multimodal diagnosis, specialized detectors, executable checks, simulator state, uncertainty, and human escalation. The diagnosis must also map to an actionable region, object, frame interval, or code component. Otherwise repeated generation can raise a selector score without repairing the stated defect. Independent held-out evaluators are important when one model family both creates and judges the synthesized content.

\textbf{Credit assignment.} Reliable observation still leaves credit assignment unresolved. A failure may originate in the prompt, retrieval, route, mask, generator, verifier, or stopping rule. GenPilot~\cite{genpilot} demonstrates feedback-guided refinement, but a rigorous study must compare its chosen repair with counterfactual actions at the same state. Matched trajectory pairs, intervention tests, and process rewards can determine whether the diagnosis caused the improvement. These tests should preserve call and cost budgets because extra candidates are not free causal evidence.

\textbf{Human feedback.} Human co-creation belongs to the same transition because user feedback~\cite{huang2026va} is an outcome that should change later actions without erasing accepted work. LAVE~\cite{lave} exposes editable timeline state for video, while UI2Code$^{\mathrm{N}}$~\cite{ui2coden} repeatedly compares code with a rendered interface. Future interfaces should expose plans, masks, timelines, layouts, candidates, uncertainty, and rollback. Evaluation should measure correction effort, preservation of prior decisions, and trust calibration in addition to final preference.

\subsection{L3 to L4: From Episodic State to Reusable Experience}
Within-task state becomes L4 only when a completed trajectory changes control on a later independent task. We first consider chronological robustness to stale experience, then use long-horizon consistency to clarify why extensive current-task state is still not L4. OctoT2I~\cite{octot2i}, GenEvolve~\cite{genevolve}, and COMFYCLAW~\cite{src260701709} illustrate capability-profile, procedure, and workflow-skill updates across tasks.

\textbf{Chronological robustness.} Persistent state must be evaluated chronologically. Tool endpoints, prices, user preferences, and safety policies can change, making earlier experience stale. A robust memory policy needs timestamps, provenance, conflict detection, selective forgetting, and rollback. Held-out transfer, retrieval ablations, memory shuffling, and negative-transfer rates are stronger evidence than success on near-duplicate requests. Privacy also becomes a control property because a stored image, prompt, or preference can influence many later tasks.

\textbf{Long-horizon consistency.} Long-horizon consistency provides a demanding test of this distinction. VisAgent~\cite{visagent} preserves narrative and character state within visual stories, Agentic 3D Scene Generation~\cite{agentic3d} maintains structured scene state, and SPIRAL~\cite{spiral} propagates action-conditioned world state across video segments. These mechanisms may support L3 even when the state is extensive. Establishing L4 requires showing that a distilled rule or memory changes control after the original project has ended.

\subsection{Beyond L4: Generator as Controller}
The L1--L4 corpus remains predominantly controller-centric. An LLM or VLM interprets the goal and selects actions, while the visual generator executes external conditions. This division creates an interface bottleneck because each result must be rendered, encoded again, and translated into a new instruction. We examine a possible generator-as-controller regime through shared generative and control state, shorter feedback loops, persistent world state, and the control evidence that remains necessary even in a unified model.

\textbf{Shared generative and control state.} A generator that also controls the trajectory could act on the same state from which it constructs the synthesized content. Instead of asking a language controller to describe a defect and then converting that description back into a prompt, the policy could associate the defect with the relevant visual tokens, latent region, frame interval, scene object, or generation step. This shared state offers a finer action space for localized revision and makes it easier to preserve accepted content across iterations. GoT~\cite{got} and Image CoT~\cite{image_cot} move in this direction by interleaving reasoning with visual generation, although their interfaces do not yet establish the complete generator-as-controller regime described here.

\textbf{Shorter feedback loops.} The same integration could shorten the feedback loop. Current LLM/VLM controllers repeatedly serialize visual state into language, call an external generator, and inspect another rendered result. A unified visual policy could allocate additional computation only to unresolved regions or transitions, reuse intermediate generation state, and stop without another cross-model exchange. The benefit is not architectural elegance by itself. It is the possibility of lower interaction cost, more precise repair, and clearer credit assignment between an observed failure and the generative decision that produced it. UI2Code$^{\mathrm{N}}$~\cite{ui2coden} demonstrates the value of making rendered feedback directly determine a later content revision, while SPIRAL~\cite{spiral} shows how action-conditioned generation and persistent state can be coupled over a longer horizon.

\textbf{Persistent world state.} Generator-side control is also attractive for persistent worlds. An external language controller usually stores a symbolic summary of what happened, while the generator separately models appearance and dynamics. If control and generation share a state representation, an action, its visual consequence, and the resulting world update can remain linked across steps. Experience from completed tasks could then improve both which action is selected and how its visual consequence is generated. Visual Generation in the New Era~\cite{visual_generation_new_era} identifies world modeling as a major stage of generator evolution. The additional requirement here is that this representation must support decisions over observation, tool use, revision, stopping, and experience reuse.

\textbf{Control remains necessary.} Using a single model does not automatically create agenticity. A unified model that produces synthesized content through a fixed inference path remains in L0 Fixed Support, regardless of whether it contains language and visual tokens. A generator-as-controller system must demonstrate that its evolving visual state causes different later actions and that retained experience changes later tasks. It must also address risks that are easier to audit when functions are distributed across several models, including opaque action traces, self-confirming evaluation, unsafe tool use, and harmful persistent updates. An intermediate design is therefore hybrid: the visual policy controls fine-grained generation and revision, while an external controller enforces user intent, permissions, provenance, and rollback. Evaluation should compare this design with an LLM/VLM-controlled counterpart under the same generator capacity, tools, and budget, rather than crediting a single-model implementation alone.

\section{Conclusion}
\label{sec:conclusion}
The world-modeling direction makes the paper's central distinction especially clear. Agentic visual generation is not simply an extension of image generation to video or 3D. It changes the unit of computation from one conditional sample to a goal-directed trajectory. Controllers ground intent, construct plans, coordinate generators and tools, create persistent visual states, judge intermediate outcomes, and adapt future actions through feedback, memory, or learning.

This work organizes agentic visual generation by the maximum causal reach of the decisions a controller can make, while a finer category and subcategory taxonomy distinguishes the technical routes taken at each level. This organization separates capability from modality, tool count, multi-role topology, training procedure, and output quality. The structured corpus shows that recent growth is concentrated in within-trajectory feedback, while persistent cross-task experience remains comparatively rare. Progress therefore depends on executable action interfaces, reliable visual verification, causal credit assignment, transparent resource accounting, selective long-term memory, editable co-creation state, and auditable provenance. Moving control into the generator could remove the lossy round trip in which visual state is repeatedly translated into language and back into generator conditions. It could also support finer revisions, lower interaction cost, and more direct credit assignment by allowing the policy to act on the state that produced the synthesized output. A potential next regime is generator-as-controller, in which generation, observation, tool selection, trajectory revision, and experience reuse are coordinated by a unified visual policy.

\nocite{*}
\bibliographystyle{IEEEtran}
\bibliography{agentic_visual_generation}

\end{document}